\documentclass[10pt,twocolumn,letterpaper]{article}
\pdfoutput=1

\usepackage{cvpr}
\usepackage{times}
\usepackage{epsfig}
\usepackage{graphicx}
\usepackage{amsmath}
\usepackage{amssymb}
\usepackage{subcaption}
\usepackage[table,xcdraw]{xcolor}
\usepackage[titletoc,toc,title]{appendix}

\DeclareMathOperator*{\argmin}{arg\,min}

\usepackage[pagebackref=true,breaklinks=true,letterpaper=true,colorlinks,bookmarks=false]{hyperref}

\cvprfinalcopy 

\def\cvprPaperID{3046} 

\ifcvprfinal\fi
\begin{document}

\title{Authoring image decompositions with generative models}

\author{Jason Rock, Theerasit Issaranon, Aditya Deshpande, and David Forsyth\\
University of Illinois at Urbana Champaign\\
{\tt\small \{jjrock2, issaran1, ardesph2, daf\}@illinois.edu}
}

\maketitle

\begin{abstract}
We show how to extend traditional intrinsic image decompositions to incorporate further layers above albedo and shading. It is hard to obtain data to learn a multi-layer decomposition.  Instead, we can learn to decompose an image into layers that are ``like this'' by authoring generative models for each layer using proxy examples that capture the {\textit Platonic ideal} (Mondrian images for albedo; rendered 3D primitives for shading; material swatches for shading detail). Our method then generates image layers, one from each model, that explain the image. Our approach rests on innovation in generative models for images.  We introduce a Convolutional Variational Auto Encoder (conv-VAE), a novel VAE architecture that can reconstruct high fidelity images. The approach is general, and does not require that layers admit a physical interpretation.
\end{abstract}

\section{Disclaimer}
This version of the paper has relatively low quality images, and some phenomena are
masked due to jpeg artifacts.  For high quality images,
please download the version available at \url{http://web.engr.illinois.edu/~jjrock2/}.

\section{Introduction}

We wish to decompose images into their component parts.
Traditionally we think of objects as smooth surfaces with albedo maps on them.  This
leads to the plausible assumption that shading effects are smooth, and locally a function
of surface normals and lighting~\cite{Ravi:sg:2001}.  However, as discussed in~\cite{Liao:2013up}, real objects are not like this.
Objects have shading detail; small bumps, pits, grooves, scratches etc. on the surface.  These mesoscopic effects
are formally due to shape, but are not captured by current intrinsic image methods because they create effects
which are qualitatively different from smooth shading.   It is compelling to consider methods that
are capable of decomposing objects into smooth shading, shading detail, and albedo.

\begin{figure}[!t]
\begin{center}
\includegraphics[trim=0 0 0 0, width=\linewidth]{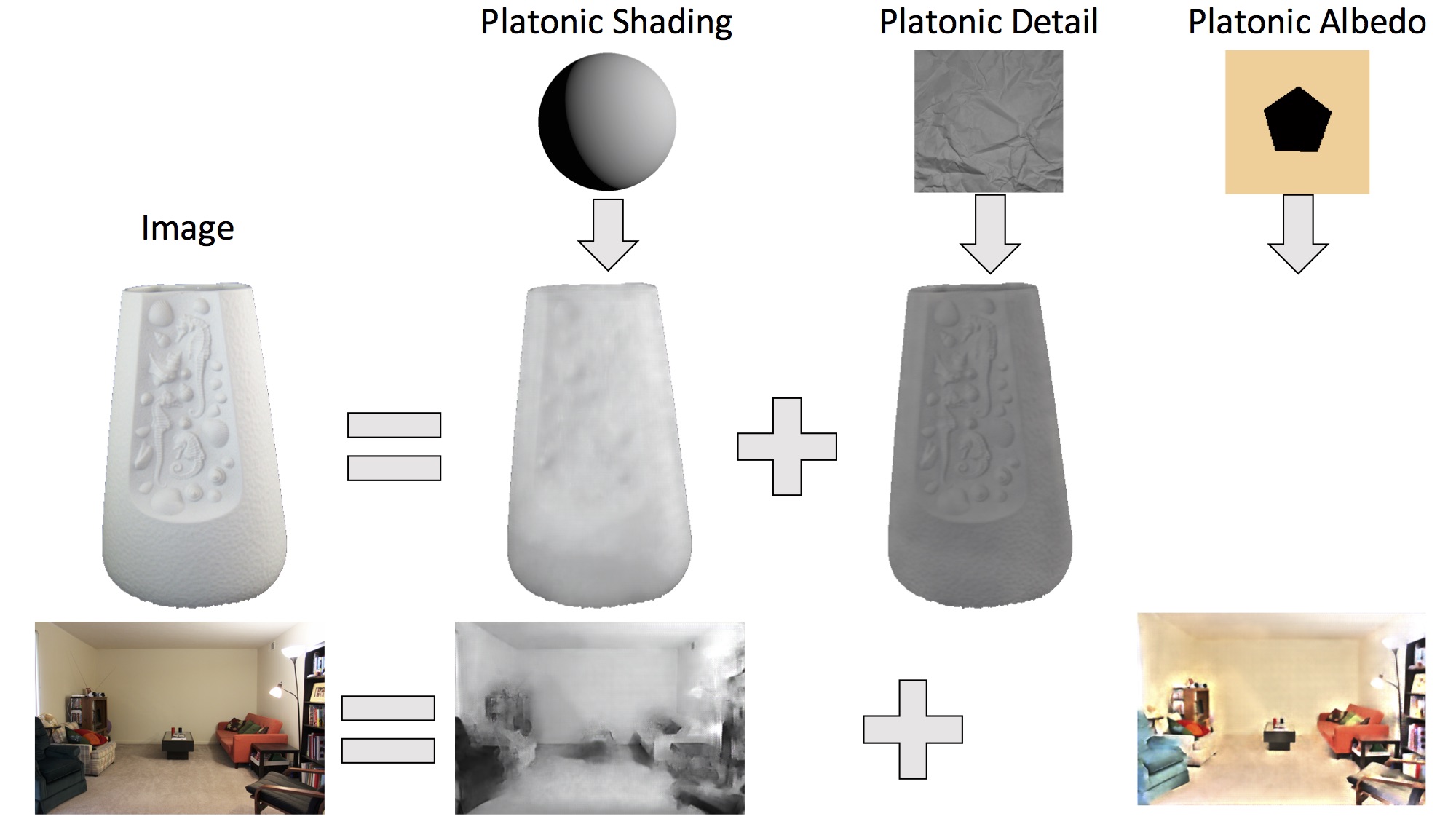}
\caption{We learn models for platonic albedo, shading, and detail independently
from platonic images.  We can then combine them as shown to perform decompositions
tasks.  For example, we can decompose the shading and detail from the vase, or the
albedo and shading of the room using the same shading model trained on platonic shading.
Note that the decompositions of shading for the vase captures the generalized cylindrical shape,
and the shading of the corners of the room are noticably darker.  Figures best viewed
in high resolution in color.}
\label{fig:teaser}
\end{center}
\end{figure}

Unfortunately, collecting images and decomposing them into ground truth albedo, shading, and shading detail
appears to be impossible. It is comparably easier to collect datasets which represent the \textit{Platonic ideals}
for each layer independently.  For example, albedo images are piecewise constant Mondrians,
shading images are realistically rendered 3D primitives, and shading detail (roughly the contribution
of surface bumps) is swatches of reasonable materials (eg. stucco walls, sand, crumpled paper)
that have minimal long scale shading effects.  These independently trained
generative models form a generative basis for images that can be combined to produce layers
that explain a full image.

Using generative models for the decomposition task won't work unless the models are capable
of producing something that looks like images.  This requires significant architectural innovation,
since current generative models create rather small and blurry images.  Our innovation is the
convolutional variant of the Variational Auto Encoder (VAE)
that is capable of producing high quality images.  We call this variant, a conv-VAE and
evaluate its representational power on images.

{\bf Contributions:} 1) We describe a Convolutional Variational Auto Encoder that is capable of representing high frequency image information.
2) We author models using the conv-VAE for specific platonic phenomena which generalize to the real phenomena.
3) We decompose images into intrinsic layers even when ground truth decomposition of images into the layers cannot be constructed.

\section{Background}

{\bf Image Prediction using Neural Networks:}
Neural networks have been applied in relatively direct ways to predict various per pixel measures
including colorization \cite{colorfuleccvpd:6G30l5H6,Larsson:2016vu}, superresolution \cite{Dong:2016fd},
intrinsic image decomposition \cite{Narihira:2015td}, depth \cite{Eigen:2014vq}, surface normals, semantic labels \cite{Long:2015cs},
pixel values \cite{PixelRecurrentNeur:bxPLT9sG}, various combinations \cite{vpdf:dFLy6Xro,Bansal:2016wt}, and \cite{pdf:dvut8KmF} introduce a generally applicable image-to-image translation tool.

Minimizing perceptual losses allows for the production of stylized images \cite{Gatys:2016ul, Combini:PUIx8BTU}
and textures \cite{Gatys:2015wf, Ustyuzhaninov:2016tc}. These perceptual losses
can be used to train feedforward networks as in \cite{Johnson:2016wm} and \cite{Li:2016wo}.
Perceptual losses can also be learned with GANs as in \cite{Dosovitskiy:2016wp}.

{\bf Generative Models for images}
Other recent work builds on generative models like encode-decoders, VAEs~\cite{AutoEncodingVariat:w8hr1WZn},
or GANs \cite{Goodfellow:2014td}.
VAEs have been used on images \cite{Kingma:2016uo}, faces \cite{Kulkarni:2015ui, AttributeImageCon:Cqjisrcl},
inpainting \cite{Pathak:2016vz}, prediction of motion \cite{AnUncertainFuture:xZxFTUB6,vpdf:SWulAZ4a}, and
room surface normals and textures \cite{Wang:2016tt}.  GANs have been used
to generate images \cite{Salimans:2016wg} and 3D shapes \cite{vpdf:DhgdsfM4}.  When combined with VAEs,
GANs can be used to learn losses \cite{Autoencodingbeyond:VyajmJEZ}.  The most similar work in this space to
ours is \cite{Anonymous:860DN96n} who use VAEs to represent an image manifold.  However, our conv-VAE
framework allows us to generate high resolution images directly from a VAE.

{\bf Intrinsic Image Decomposition:}
Splitting an image into shading and albedo components is a classical computer vision problem~\cite{LAND:71}, as is explaining shading
by reconstructing surface normals~\cite{horn1970shape}.
There is a strong tradition of obtaining intrinsic images as inference on a
generative physical model.  Write ${\cal I}$ for the log image, ${\cal A}$
for the log albedo image, and ${\cal S}$ for the log shading image.
One then assumes that ${\cal I}={\cal A}+{\cal S}$, and seeks
solutions for ${\cal A}$ and ${\cal S}$ that maximize priors.  Similarly, reconstructing surface
normals from shading is seen as inference on a generative physical model -- one seeks a normal field ${\bf n}({\bf x})$ that
explains the shading ${\cal S}$ under some rendering model~\cite{horn1970shape}.   Barron and Malik show that
attacking these problems together yields an extremely strong method for recovering albedo and shading, as well
as the best current shape reconstructions~\cite{Barron:2013uc}.

These traditions are odd.  It is easily verified with current datasets
(eg.~\cite{Grosse:2009ji,bell14intrinsic}) that ${\cal I} \neq {\cal A}+{\cal S}$ (as a result of
mesostructure, glossy, and subsurface scattering phenomena).
Even in constrained lab environments built to determine ground truth,
it is difficult to fully separate the phenomena into the correct layers, in ground truth images from
\cite{Grosse:2009ji} the mesostructure shading bleeds into the albedo layer.
Similarly, all shading models that yield tractable reconstruction methods are physically
incorrect~\cite{Forsyth:1989bq}.

Prior to neural networks, methods for intrinsic images~\cite{Gehler:2011vc, Barron:2013uc, Shen:2011gf} used
hand defined priors for the albedo and shading channels in \cite{Grosse:2009ji}.
Recent works, which use neural networks to predict intrinsic image decompositions \cite{Narihira:2015td, Zoran:2015ee}
directly use SINTEL \cite{Butler:ECCV:2012} or IIW \cite{bell14intrinsic} to augment the MIT dataset
because they otherwise will not have enough data.  Our method, does not need ground truth decompositions
because we do not train the albedo, shading, or shading detail models jointly.  Instead, 
we train independent representations on Platonic ideals, and then combine them to form
a decomposition model.

\section{VAE Architecture}
We will briefly describe VAEs in practice to motivate our conv-VAE.  In depth theoretical
discussion can be found in~\cite{AutoEncodingVariat:w8hr1WZn} and a nice tutorial is \cite{Doersch:2016vb}.

In essence, one can think about training a VAE as training two networks, an encoder $E(I)$ which is trained
to map images $I$ to latent variables, usually called codes $z$, and a decoder $D(z)$ that is trained
to map these codes to images.  A variational criterion is used to ensure that (a) codes are distributed as
$z \sim \mathcal{N}(0,1)$ (b) decoding a code $D(z)$, with $z=E(I)$ yields the image $I$ and (c) decoding a code
near some $z=E(I)$ yields an image close to $I$.

\begin{figure}[h!]
\centering
\includegraphics[trim=0 210 200 0,width=\linewidth]{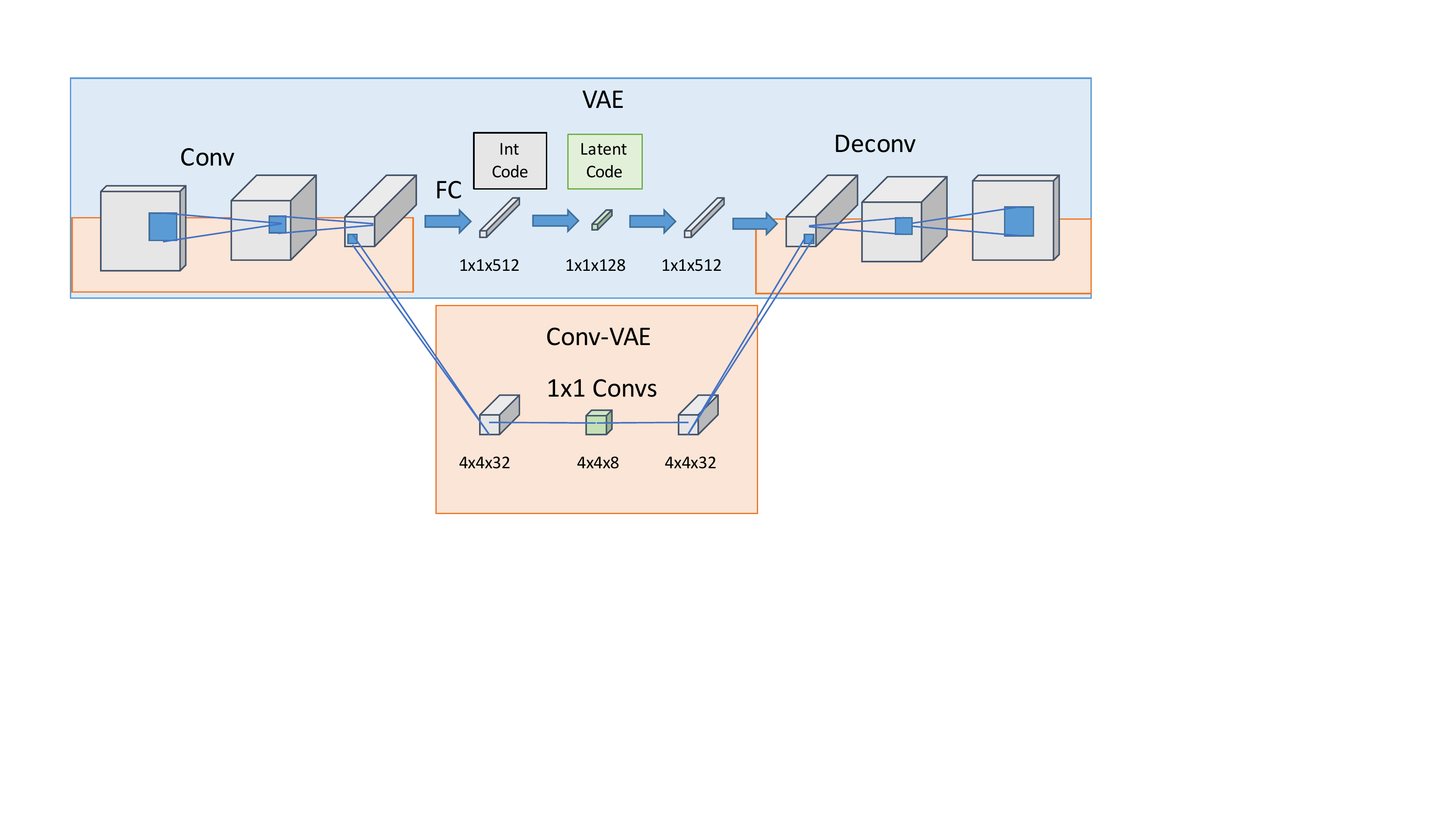}
\caption{The difference between a VAE and convolutional VAE is entirely in the ``code'' layers.  Notice that the dimensionality of the
latent representation doesn't change, but the size of the representation is drastically reduced because very large
fully connected layers are replaced by 1x1 convolutions.  Replacing the fully connected layers with convolutions can be thought of
as constructing a diagonal matrix, and does not significantly impact the VAE theory.}
\label{fig:conv_vae_arch}
\end{figure}

\subsection{Convolutional VAE}
The basic VAE has some problems as a model for images, especially high resolution images.
First, the VAE model has difficulties producing high spatial frequencies. Second,
the global codes make it difficult to learn that images are shift and rotationally
invariant.  Third, there is no way to apply a VAE to images of varying sizes.  Previous
works have tried to solve these issues by generating images pixel-by-pixel, conditioning
on previously seen pixels.\cite{PixelRecurrentNeur:bxPLT9sG}

We propose the conv-VAE for achieving these goals.  Rather than creating a single global code
for an image, we create a field of codes that describe local regions.
This means that our latent space is a code ``image'' rather than a code vector as shown in figure~\ref{fig:conv_vae_arch}.
For example, on a 64x64 image, we would write a 128 bit code as a 4x4x8 code, where each ``pixel'' location impacts about one
quarter of the output image.  Since we replaced a fully connected layer that produced a 128 bit code
with a convolution that produces 8 bit codes, it reduces the number of parameters in the network.
However, the conv-VAE is still capable of reproducing images better because it strictly enforces locality in
the latent space.  This comes at the cost of independence between the dimensions of the code,
which makes drawing valid codes from the latent space more difficult. However, 
for decomposition this is not a problem because we have other constraints.

\subsection{Laplacian Conv-VAE}
Training conv-VAEs to produce sharp images is still difficult because the typical reconstruction loss for $D(z)$ is
$\|D(E(I))-I\|^2$ which does not capture the importance of high frequency edges.
As such, generated images are often blurry. Techniques for improving encoder-decoder results
include filtering the image with a Laplacian filter to emphasize edges~\cite{AttributeImageCon:Cqjisrcl} and
discretizing the continuous color space and using a cross entropy loss~\cite{PixelRecurrentNeur:bxPLT9sG}.

We extend the Laplacian filter idea by learning a VAE for each layer of a Laplacian Pyramid.
This is good for three reasons.  First, predicting the high frequency
layers of a Laplacian Pyramid is easy because they are often 0.  Second, an L2 Loss
can be applied at each layer of the Laplacian Pyramid, requiring that all frequencies of the input
image are correctly captured.  Third, predicting each layer with a conv-VAE explicitly places
additional code capacity on the higher frequencies since they are encoded as larger images.
We are not the first to apply Laplacian Pyramids in generative networks. \cite{Denton:2015wp} train
a GAN to predict a Laplacian Pyramid. Our approach differs as we do not condition across scales
and instead treat our conv-VAEs at each level as independent predictors.

\subsection{Modeling Examples with VAEs}
We validate conv-VAEs by comparing to conventional VAEs reproduction on images
from ImageNet.   Unlike previous works, which resize images to a small size, we
take 64x64 pixel crops during training.  This makes the assumption that images are translation
invariant explicit.  We compare a conv-VAE with and without the Laplacian Pyramid to two conventional VAEs.
VAE-1 has roughly the same number of network parameters, and VAE-2 has roughly the same
number of latent values as our conv-VAEs.  Due to the fully connected layers in the conventional VAE, it is impossible to provide
both at once. Details of the full parameterization is provided in table~\ref{tab:vae_archs}.
We train each model for 25 epochs using Adam with an exponentially decreasing step length.
Other hyperparameter details are the same as for our other experiments and are given in section~\ref{sec:vae_train_detail}

\begin{figure}
\centering
\includegraphics[width=.19\linewidth]{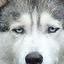}
\includegraphics[width=.19\linewidth]{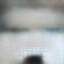}
\includegraphics[width=.19\linewidth]{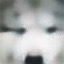}
\includegraphics[width=.19\linewidth]{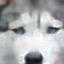}
\includegraphics[width=.19\linewidth]{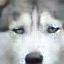}\\

\includegraphics[width=.19\linewidth]{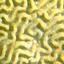}
\includegraphics[width=.19\linewidth]{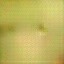}
\includegraphics[width=.19\linewidth]{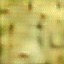}
\includegraphics[width=.19\linewidth]{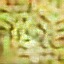}
\includegraphics[width=.19\linewidth]{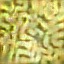}\\

\includegraphics[width=.19\linewidth]{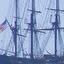}
\includegraphics[width=.19\linewidth]{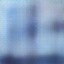}
\includegraphics[width=.19\linewidth]{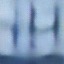}
\includegraphics[width=.19\linewidth]{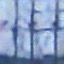}
\includegraphics[width=.19\linewidth]{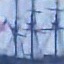}\\

\begin{subfigure}[t]{.19\linewidth}
\includegraphics[width=\linewidth]{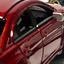}
\caption{Image}
\end{subfigure}
\begin{subfigure}[t]{.19\linewidth}
\includegraphics[width=\linewidth]{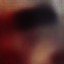}
\caption{vae-1}
\end{subfigure}
\begin{subfigure}[t]{.19\linewidth}
\includegraphics[width=\linewidth]{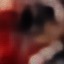}
\caption{vae-2}
\end{subfigure}
\begin{subfigure}[t]{.19\linewidth}
\includegraphics[width=\linewidth]{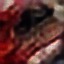}
\caption{cvae}
\end{subfigure}
\begin{subfigure}[t]{.19\linewidth}
\includegraphics[width=\linewidth]{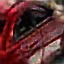}
\caption{lcvae}
\end{subfigure}

\caption{Conv-VAEs (d) and (e) always outperform traditional VAEs (b) and (c) even though
(c) has as large a latent space, and the conv-VAEs have far fewer parameters.
The Laplacian conv-VAE (e) is more capable of reconstructing small features of images than the conv-VAE (d).
Even for relative failure cases for the method, like the last row, the conv-VAEs perform significantly better.
Best viewed in color at high resolution}
\label{fig:ImagenetResults}
\end{figure}

Recall that a VAE has an encoder $E(I)$ and a decoder $D(z)$.  To compare representational power,
we take an image patch $I$ and look at the result of encoding and then decoding it: $D(E(I))$.
We do this for patches of the 200 held out images select so that they lie on interesting portion of the
image, rather than the background.
Quantitively on an L2 error measure on the held-out patches, VAE-1 has error .23, VAE-2 has error .18, conv-VAE has error .13, and
Laplacian conv-VAE has error .11.  Qualitative results, shown in figure~\ref{fig:ImagenetResults}
validate these findings.

\begin{figure}
\centering
\includegraphics[width=.49\linewidth]{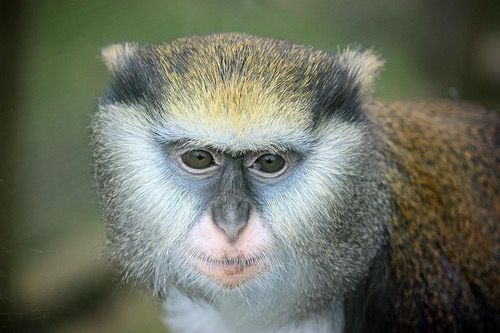}
\includegraphics[width=.49\linewidth]{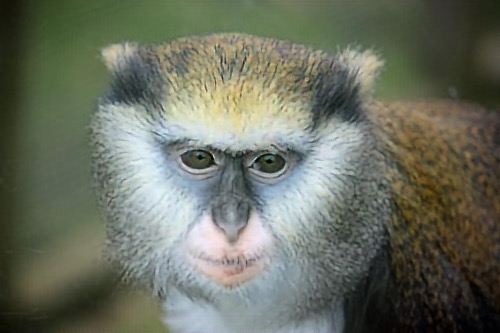}

\includegraphics[width=.49\linewidth]{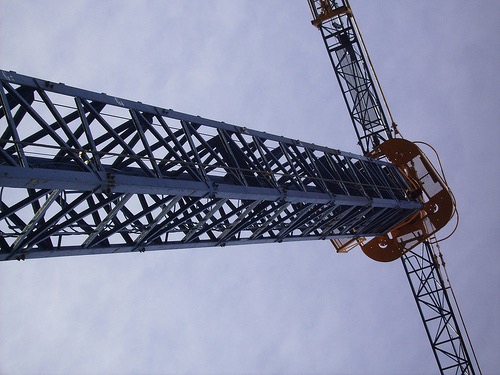}
\includegraphics[width=.49\linewidth]{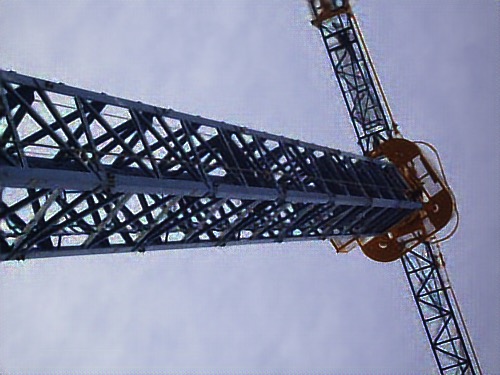}

\includegraphics[width=.49\linewidth]{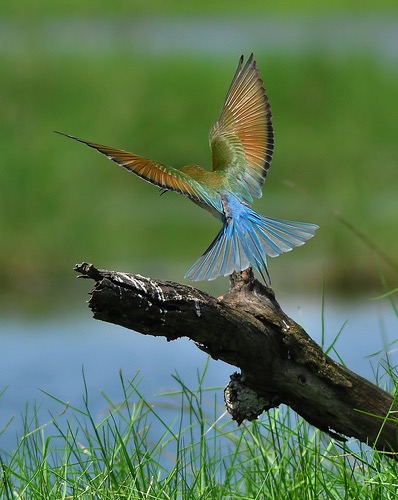}
\includegraphics[width=.49\linewidth]{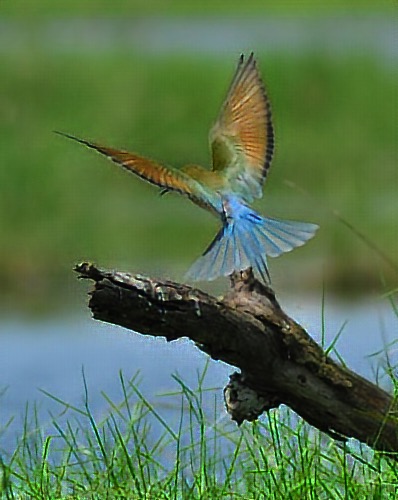}

\caption{Laplacian conv-VAE trained with 64x64 patches on full images.
On the left is the input image.  On the right is an image created by encoding and then decoding the input image.
Our model typically does a good job of handling image phenomena, though
there are some mild checkerboard patterns.  Best viewed in color at high resolution.}
\label{fig:ImagenetFullImages}
\end{figure}

Our convolutional models are also capable of modeling full ImageNet images with no additional training.
We encode and decode using full images rather than patches as above.  Results
for the Laplacian conv-VAE on full images is shown in figure~\ref{fig:ImagenetFullImages}.  We emphasize
that our model only saw 64x64 patches during training, but it correctly reproduces the long-scale structure
of the image.

\subsection{Decomposition with conv-VAE Models}
We have described a method that can learn high quality representations for images.
Assume that we have a generative model for log Platonic albedo and a generative model for log Platonic shading.
We want to use these models to decompose an image into its component parts.  We note that some albedo
(resp. shading) codes are more common than others.  Furthermore, when there are phenomena
that can be explained by either layer (eg. cast shadows) we want to force our decomposition
to choose -- the layers should not have strong correlation.
We would like to obtain a decomposition that (a) uses common codes for each layer
(b) explains the image and (c) produces decorrelated layers.

The conv-VAE models are composed of $E_a(\cdot)$ and $D_a(\cdot)$ for the albedo encoder and decoder.
Similarly, $E_s(\cdot)$ and $D_s(\cdot)$ describe the VAE for shading.
For the Laplacian conv-VAE, the encoder and decoder produce a set of code images and a Laplacian
decomposition respectively.
We build a probabilistic model of common codes per code ``pixel'' for albedo and shading.
First, we encode ground truth patches to recover code images. 
For the Laplacian conv-VAE, we recover the set of code ``images'', and 
concatenate the codes, resizing the smaller ``images'' to the size of the largest ``image''.
We then treat each pixel as an iid draw from a distribution and fit a probability density model to the
set.  In practice we use a Gaussian model with a spherical covariance.
Let the probability models for albedo and shading be $P_a(z)$ and $P_s(z)$.  Let $z^a_i$ be the $i$th pixel of the albedo code, 
(resp $z^s_i$).
We write the full code loss as a negative log likelihood $P(z^a, z^s) = -1/N \sum_i \log(P_a(z^a_i)) + \log(P_s(z^s_i))$.

As is traditional in intrinsic images,
we enforce the image be explained by the codes by minimizing the residual
$\mathcal{R}(z^s, z^a, I) = \|\log(I) - D_s(z^s) - D_a(z^a)\|^2$.

To decorrelate the layers, we introduce a term which is meant to force the
decomposition to ``make up its mind'' about where a signal should live.
We define an upper bound on the correlation with the Frobenius
norm of the 3x3 covariance matrix of the spatially corresponding pixel values in the albedo and shading.
This correlation measure is attractive because it is simple and can be applied patch-wise,
at varying scales, and across Laplacian layers.
Let $\text{cov}(\cdot,\cdot)$ be the operation that computes the covariance matrix by centering two signals,
and then taking the mean of the outer products.
Let $P$ be the number of patches and $L$ the number of Laplacian layers.  Let $A^{(l)}_p$ be
the $p$th patch in the $l$th Laplacian layer of the albedo prediction.
The correlation is $\text{corr}(A,S) = \frac{1}{L\cdot P}\sum_{l} \sum_{p} \| \text{cov}(A^{(l)}_p, S^{(l)}_p)\|_F$

Our full decomposition equation is
\begin{equation}
\begin{aligned}
\argmin_{z^a, z^s} & \mathcal{R}(z^s, z^a, I) \\
& - \lambda_p P(z^a, z^s)\\
& + \lambda_c \text{corr}(D_s(z_s), D_a(z_a))
\end{aligned}
\end{equation}

\section{Authored Training Data and Models}
\label{sec:training_data}
\begin{table}[]
\centering
\begin{tabular}{|lllll|}
\hline
\multicolumn{1}{|c}{Filter} & \multicolumn{4}{c|}{Layer}                \\ \hline
\rowcolor[HTML]{BBDAFF} 
                            & 128x128x3 & 64x64x3  & 32x32x3  & 16x16x3 \\
5x5                         & 64x64x64  & 32x32x64 & 16x16x64 & 8x8x64  \\
5x5                         & 32x32x64  & 16x16x64 & 8x8x64   & 4x4x64  \\
3x3                         & 32x32x64  & 16x16x64 & 8x8x64   & 4x4x64  \\
4x4                         & 29x29x64  & 13x13x64 & 5x5x64   & 1x1x64  \\
\rowcolor[HTML]{B8D8A4}
1x1                         & 29x29x4   & 13x13x4  & 5x5x4    & 1x1x4   \\ \hline
\end{tabular}
\caption{The VAE architecture we use for all of our authored models is a Laplacian conv-VAE with 4 Laplacian layers,
and a code size of 4 per code ``pixel''.  As in table~\ref{tab:vae_archs}, we show only the encoder portion, as the decoder is
the reverse.  For a 128x128 image, this compresses to about 8\% of the original image size.
In the limit as image size increases, the compression rate is about 11\%.}
\label{tab:AuthoredVAEModel}
\end{table}

\begin{figure}[h!]
\begin{subfigure}[b]{\linewidth}
\includegraphics[width=.24\linewidth]{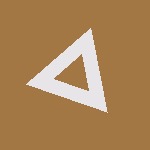}
\includegraphics[width=.24\linewidth]{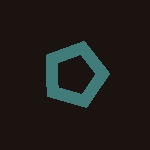}
\includegraphics[width=.24\linewidth]{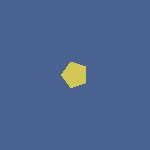}
\includegraphics[width=.24\linewidth]{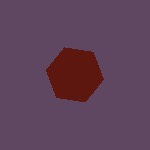}
\caption{Platonic albedo is two color Mondrian images.}
\label{fig:td:mondrian}
\end{subfigure}
\begin{subfigure}[b]{\linewidth}
\includegraphics[width=.24\linewidth]{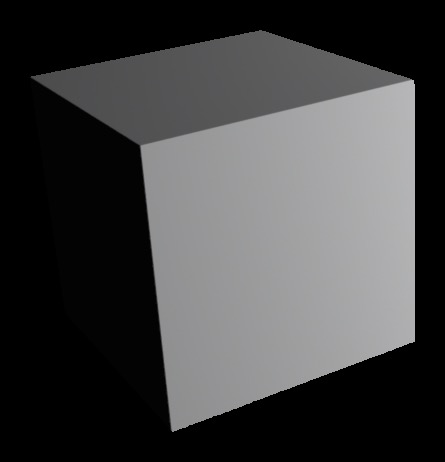}
\includegraphics[width=.24\linewidth]{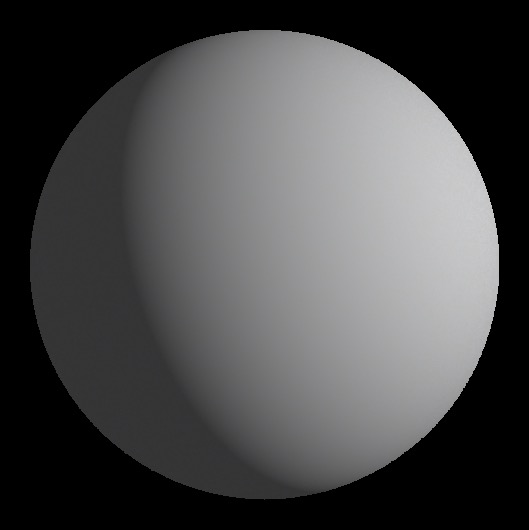}
\includegraphics[width=.24\linewidth]{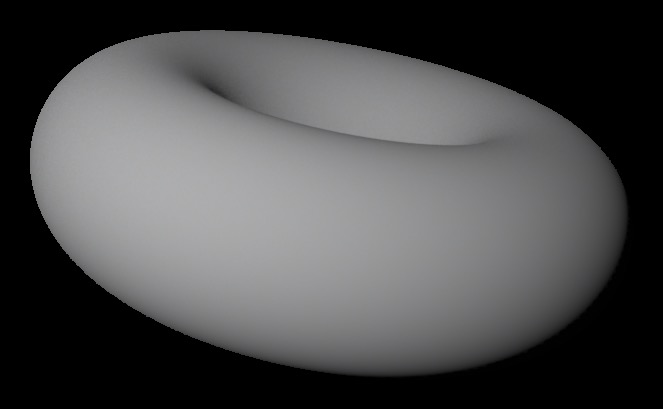}
\includegraphics[width=.24\linewidth]{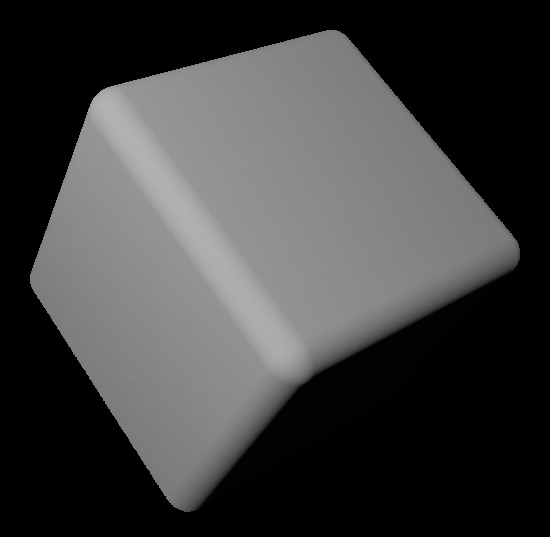}
\caption{Platonic shading is rendered primitive shapes under varying lighting conditions.}
\label{fig:td:shading}
\end{subfigure}
\begin{subfigure}[b]{\linewidth}
\includegraphics[width=.24\linewidth]{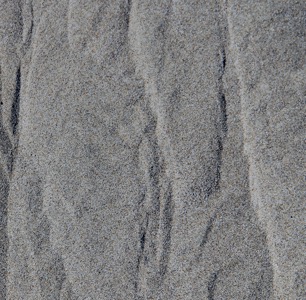}
\includegraphics[width=.24\linewidth]{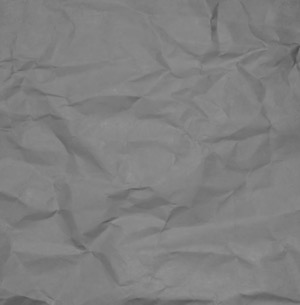}
\includegraphics[width=.24\linewidth]{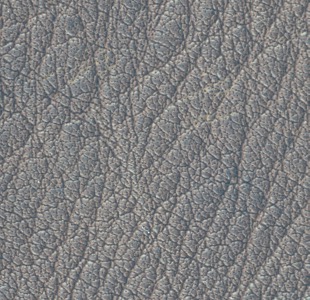}
\includegraphics[width=.24\linewidth]{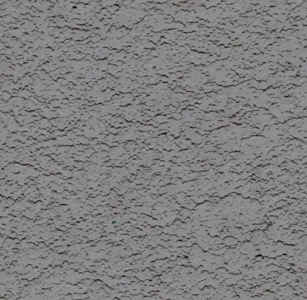}
\caption{Platonic shading detail is caused by small bumps on an otherwise flat smooth surface under uniform lighting.}
\label{fig:td:material}
\end{subfigure}
\caption{Example Platonic training data images are used to train VAE models.  Notice that the representation
each captures is unique.}
\end{figure}

\subsection{Albedo}
Platonic albedo, figure~\ref{fig:td:mondrian}, is piecewise constant.  We create a set of
2-color Mondrian images, where a polygon at the center
has a different color than the rest of the image.  The colors for these
are drawn from the palette of colors from the 10-train set of MIT images.
We generate 500 Mondrian images of size 150x150 to allow for
128x128 crops that are shifted or rotated.

\subsection{Shading}
Platonic shading, figure~\ref{fig:td:shading}, is the effect of light on a simple smooth surface.  We generate
platonic shadings by rendering 3D primitives with no surface color 
using LuxRender.  We use a directional light source which rotates about the the
object with a fixed camera view.  We either provide a fill light where dark pixels get about 10\% of pure white, 
or a weak fill light where dark pixels get about 1\% of pure white.
The weak fill light matches the lighting conditions found in MIT, while the stronger fill light
is a good proxy for more typical lighting, where diffuse components are quite large.
There are 70 images, cropped 
to the bounding box of the rendered object so that the images are about 500x500 pixels.  During training, we
take 128x128 crops from these, such that the crops lie mostly inside of the masked object.

\subsection{Shading Detail}
Platonic shading detail, figure~\ref{fig:td:material} is the effect from small bumps on a flat smooth surface under uniform lighting.
We collect swatches of materials from the Internet including sand which has wavy texture, stucco which has repetitive bumps,
and creased paper which varies between the two.  These images are similar to those used in video games for texturing.
These swatches are mostly planar and have minimal long-scale shading effects
and almost no albedo effects.\footnote{While sand technically has albedo effects caused by the varying colors
of sand grains, this is different from what we think of as platonic albedo in a similar manner to how
shading detail is different from platonic shading.} We remove contributions of shading and center
the images at .5 by fitting a linear shading gradient to the image.  We also crop to $[0,1]$, to guarantee
existence of the log image. Our dataset consists of 45 images of about 300x300 pixels. 

\subsection{Laplacian conv-VAE Training Details}
\label{sec:vae_train_detail}
Our VAE model's architecture is described in table~\ref{tab:AuthoredVAEModel}.  We use a 4-layer Laplacian decomposition,
but it is otherwise the same as the Laplacian conv-VAE used on ImageNet.  We train the model to encode and decode
log versions of the images so that they can be used directly for intrinsic image decomposition,
where it is common to write $\log(I) = \log(A) + \log(S)$.  This also means that we do not use a sigmoid activation layer
for the final output layer. We use the Adam optimizer with an exponentially decaying learning rate starting at .001 
decaying every 500 iterations by .9.  For $t$, the iteration and $\sigma(\cdot)$ the sigmoid function,
our KL divergence terms weight is  $20 \sigma(.02\cdot t-.5)$.  The image residual loss is weighted by
$1000$.  An image prior, the L1 of image gradients, is weighted by $.1$.

\section{Experimental Results}
We emphasize that all results are obtained with the same models of Platonic
albedo (A), shading (S), and shading detail (D).
Different decompositions are obtained by using different combinations of models (i.e.
A,S; S,D; A,S,D; etc).

\subsection{Albedo and Shading Decomposition}
\begin{figure*}[!h]
\centering

\begin{subfigure}[b]{.18\textwidth}
\includegraphics[width=\linewidth]{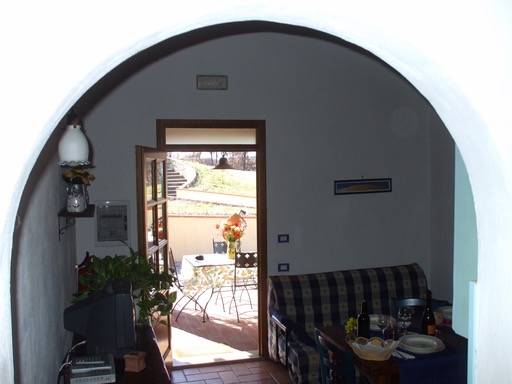}
\includegraphics[width=\linewidth]{Figures/IIW/im_6_99_0}
\includegraphics[width=\linewidth]{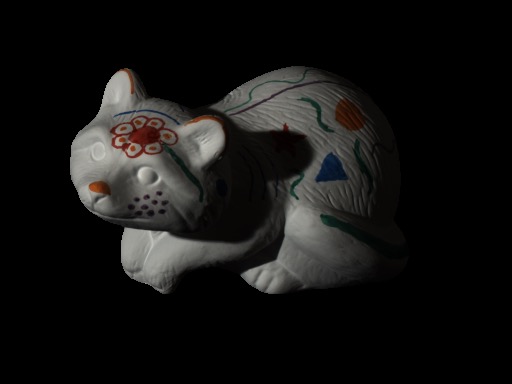}
\includegraphics[width=\linewidth]{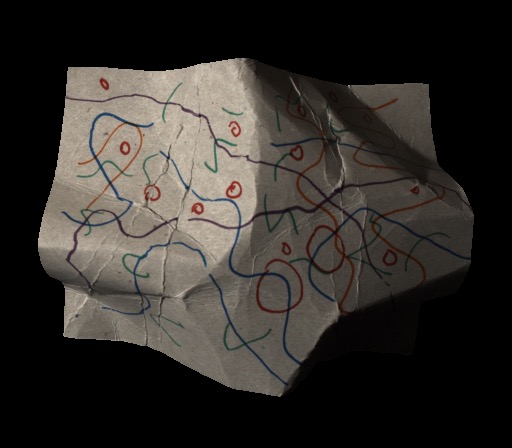}
\includegraphics[width=\linewidth]{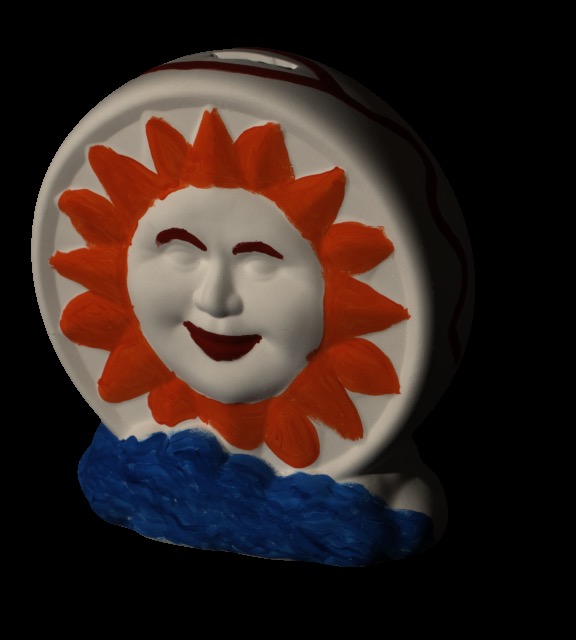}
\includegraphics[width=\linewidth]{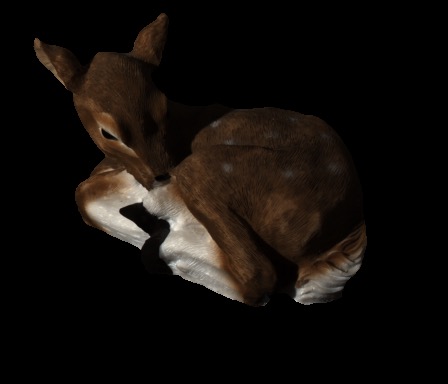}
\caption{Input Image}
\end{subfigure}
\begin{subfigure}[b]{.18\linewidth}
\includegraphics[width=\linewidth]{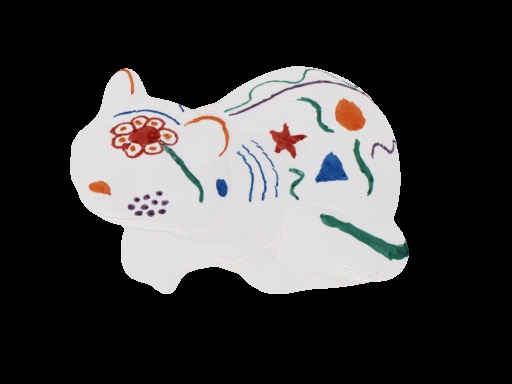}
\includegraphics[width=\linewidth]{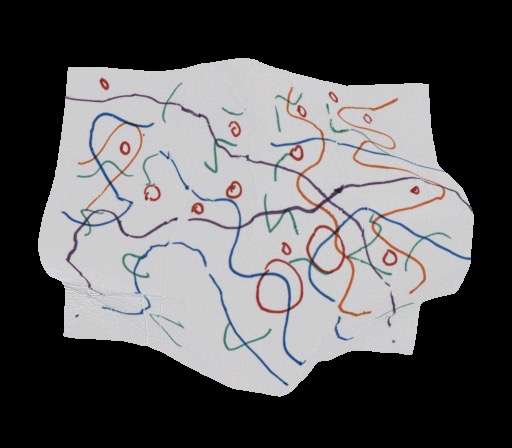}
\includegraphics[width=\linewidth]{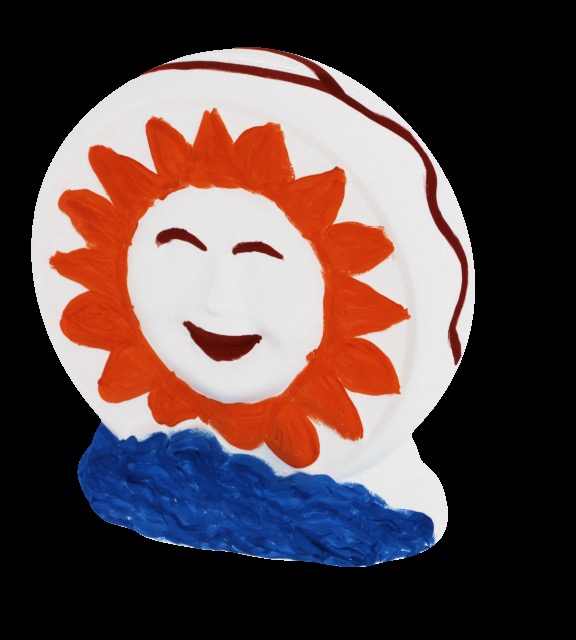}
\includegraphics[width=\linewidth]{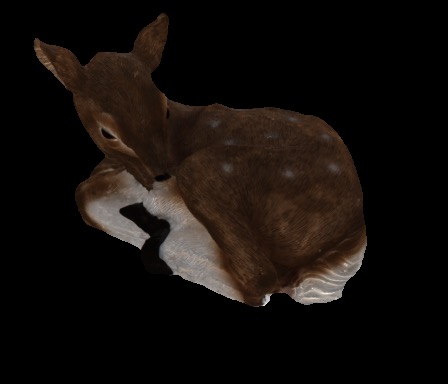}
\caption{GT Albedo}
\end{subfigure}
\begin{subfigure}[b]{.18\linewidth}
\includegraphics[width=\linewidth]{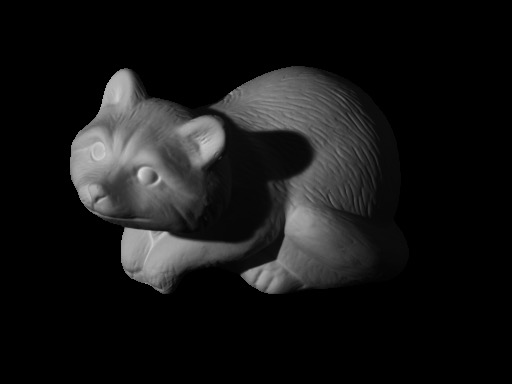}
\includegraphics[width=\linewidth]{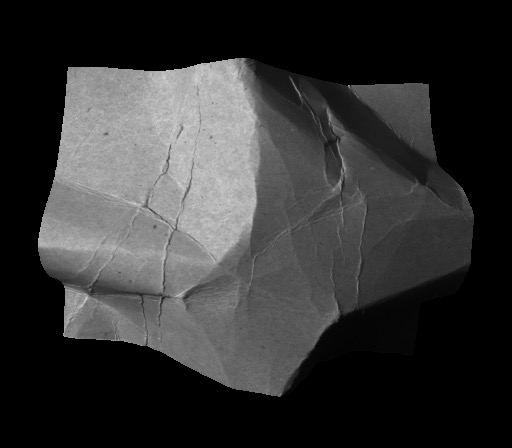}
\includegraphics[width=\linewidth]{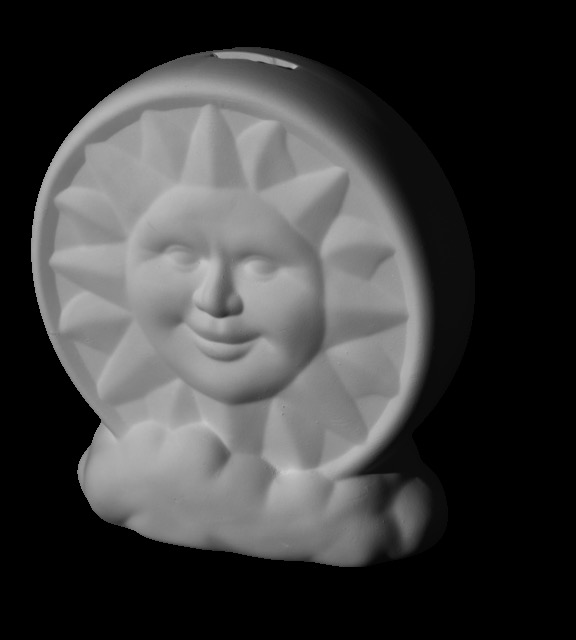}
\includegraphics[width=\linewidth]{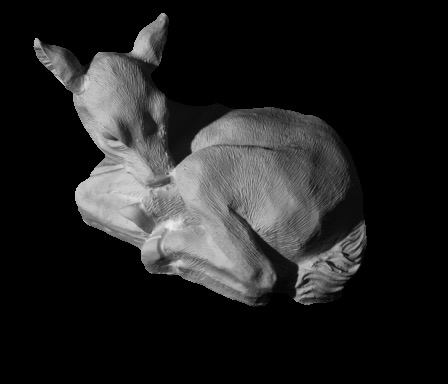}
\caption{GT Shading}
\end{subfigure}
\begin{subfigure}[b]{.18\linewidth}
\includegraphics[width=\linewidth]{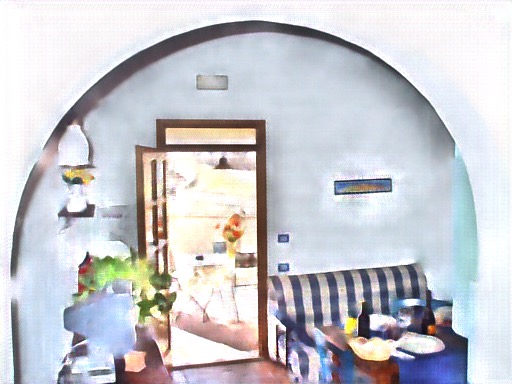}
\includegraphics[width=\linewidth]{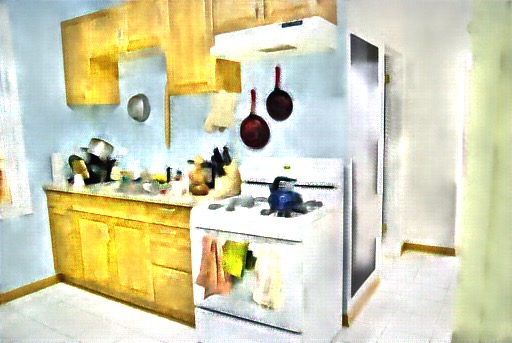}
\includegraphics[width=\linewidth]{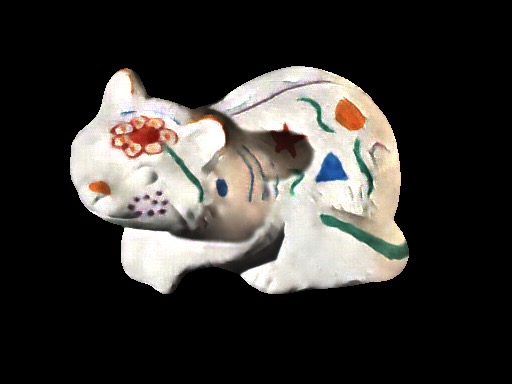}
\includegraphics[width=\linewidth]{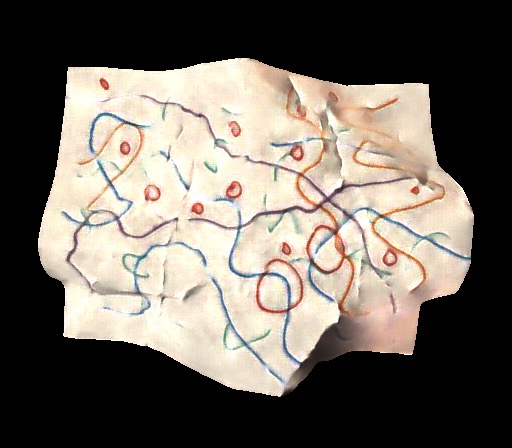}
\includegraphics[width=\linewidth]{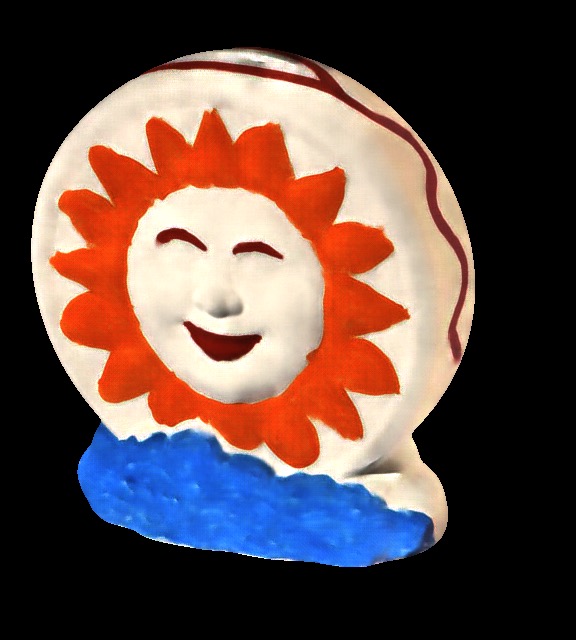}
\includegraphics[width=\linewidth]{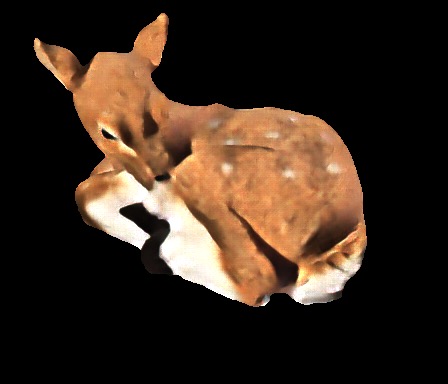}
\caption{Pred Albedo}
\end{subfigure}
\begin{subfigure}[b]{.18\linewidth}
\includegraphics[width=\linewidth]{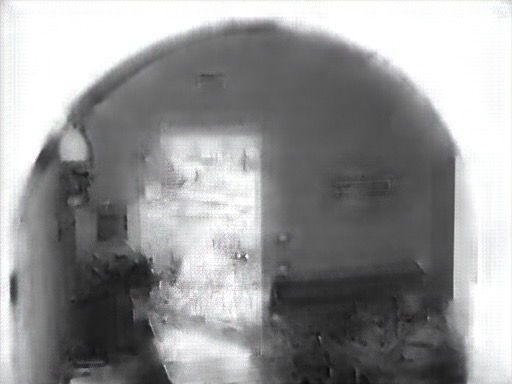}
\includegraphics[width=\linewidth]{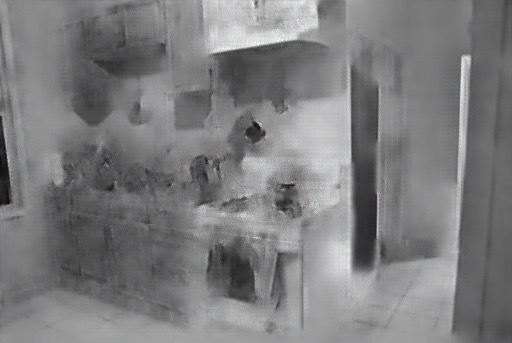}
\includegraphics[width=\linewidth]{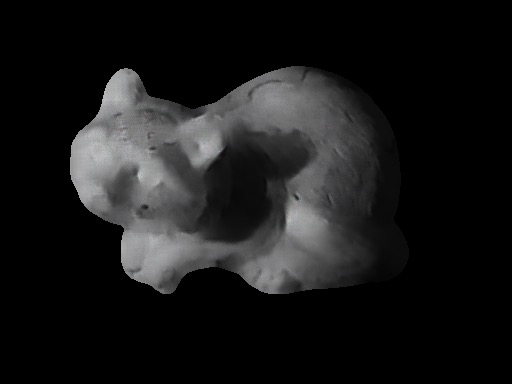}
\includegraphics[width=\linewidth]{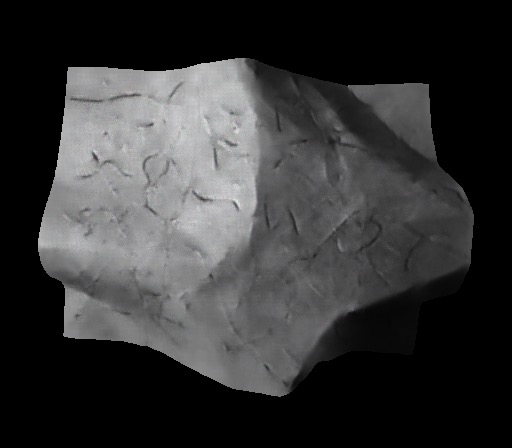}
\includegraphics[width=\linewidth]{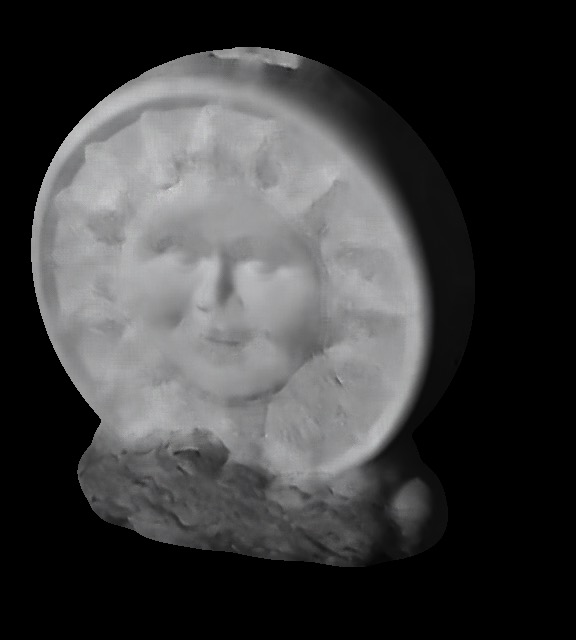}
\includegraphics[width=\linewidth]{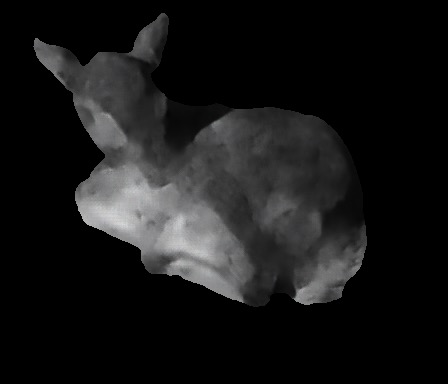}

\caption{Pred Shading}
\end{subfigure}

\caption{Our Intrinsic image decomposition results on MIT test data (bottom 3) and IIW images (top 2) are
qualitatively good even though our model was never trained to do decomposition
on images.  Our model produces albedo (resp. shading) fields with a strong
similarity to  the ideal (eg. constant patches of color).
This is pronounced compared even to the MIT ground truth.  For example, is the roughness at
the base of the sun in gt really an albedo effect?  Best viewed in high resolution in color.
All images produced using the same platonic models, but in different combinations (A,S here).}
\label{fig:MIT_AS_qual}
\end{figure*}

\begin{figure*}
\centering

\begin{subfigure}[b]{.18\linewidth}
\includegraphics[width=\linewidth]{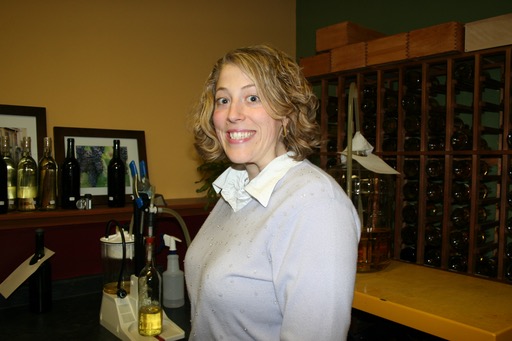}
\includegraphics[width=\linewidth]{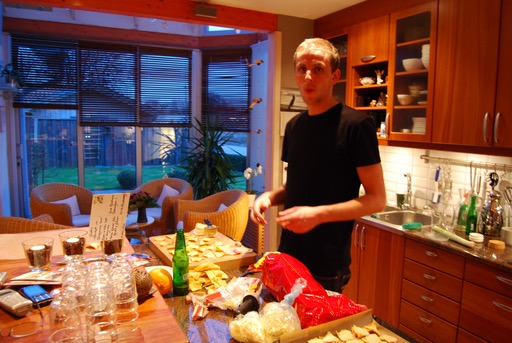}
\includegraphics[width=\linewidth]{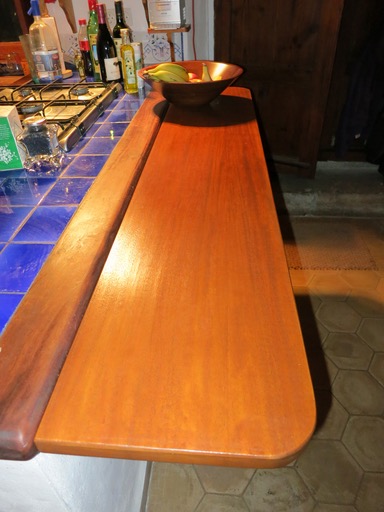}
\includegraphics[width=\linewidth]{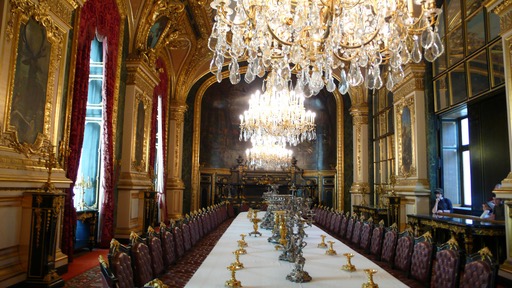}
\includegraphics[width=\linewidth]{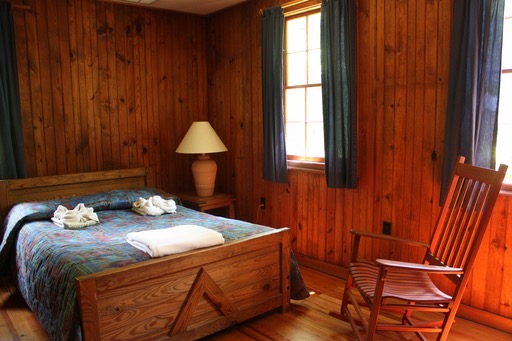}
\includegraphics[width=\linewidth]{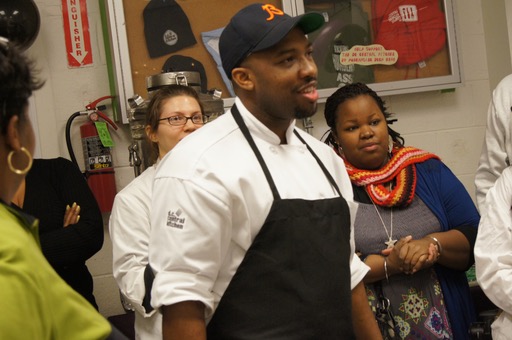}
\includegraphics[width=\linewidth]{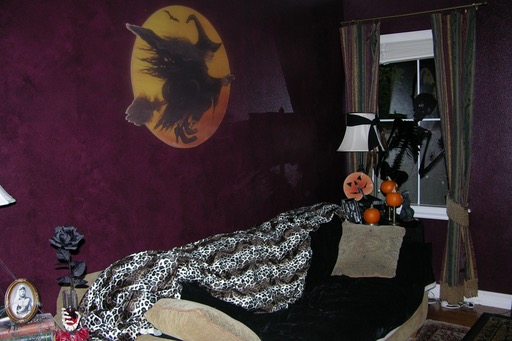}
\includegraphics[width=\linewidth]{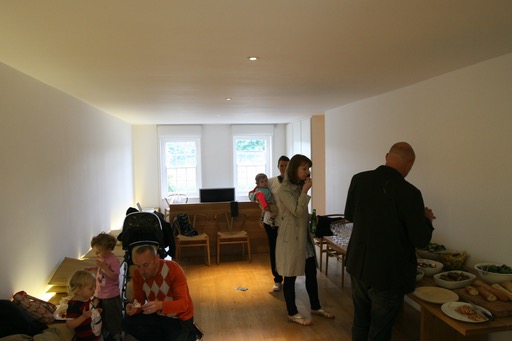}

\caption{Input Image}
\end{subfigure}
\begin{subfigure}[b]{.18\linewidth}
\includegraphics[width=\linewidth]{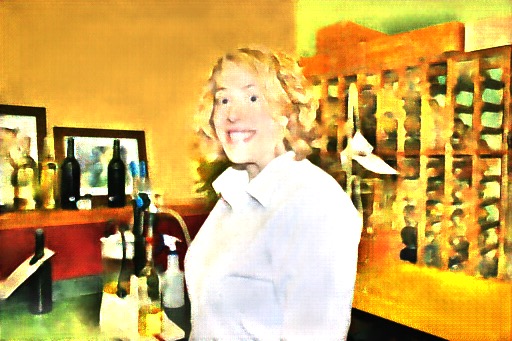}
\includegraphics[width=\linewidth]{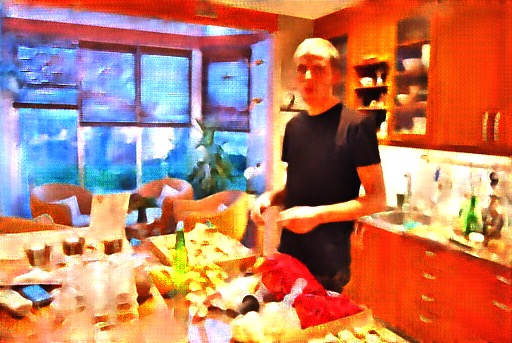}
\includegraphics[width=\linewidth]{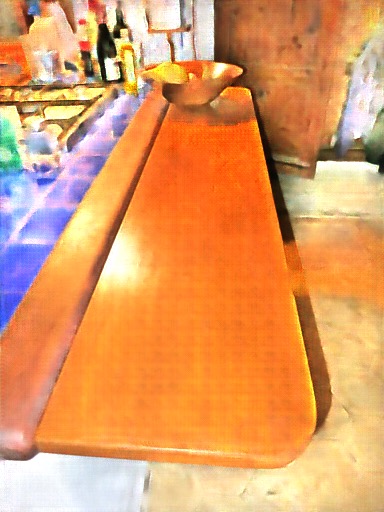}
\includegraphics[width=\linewidth]{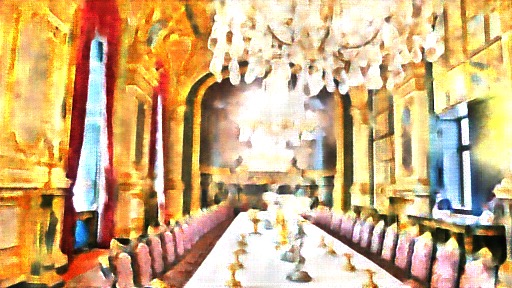}
\includegraphics[width=\linewidth]{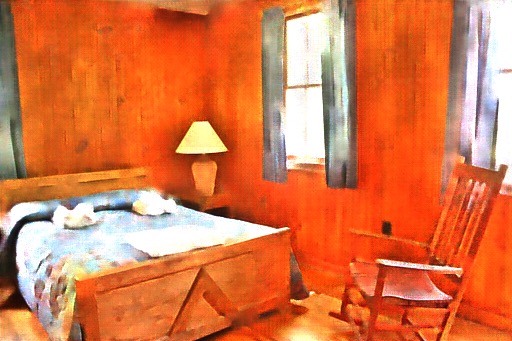}
\includegraphics[width=\linewidth]{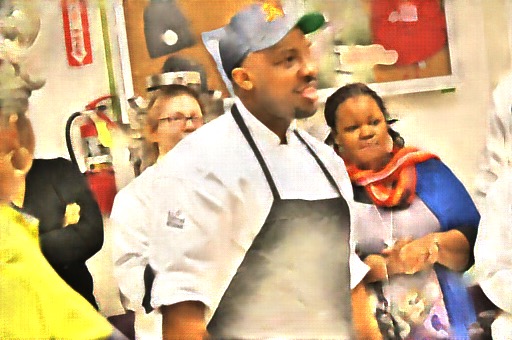}
\includegraphics[width=\linewidth]{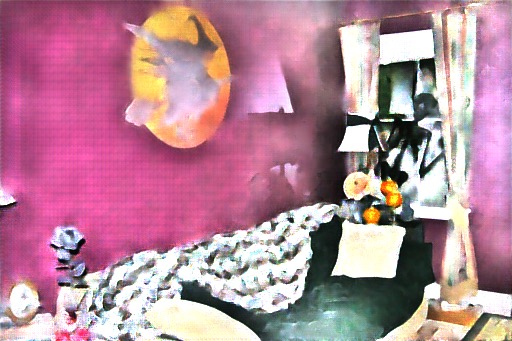}
\includegraphics[width=\linewidth]{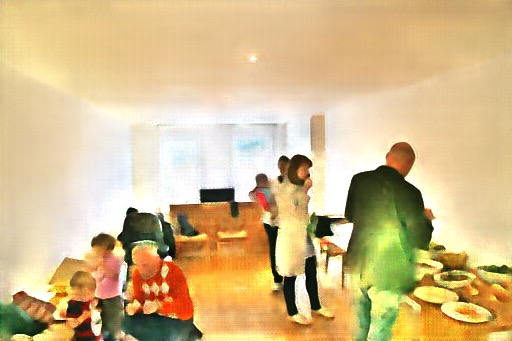}

\caption{Our Albedo}
\end{subfigure}
\begin{subfigure}[b]{.18\linewidth}
\includegraphics[width=\linewidth]{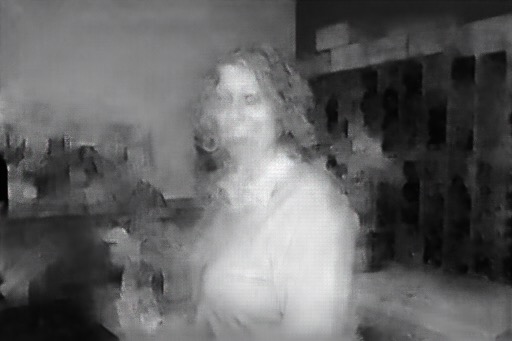}
\includegraphics[width=\linewidth]{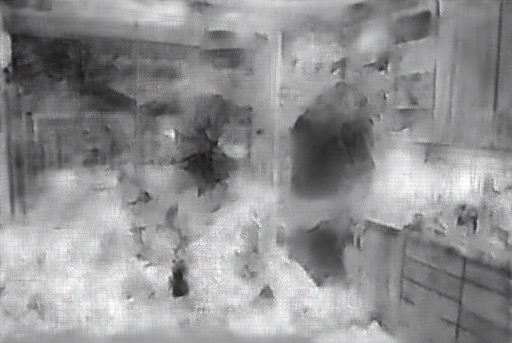}
\includegraphics[width=\linewidth]{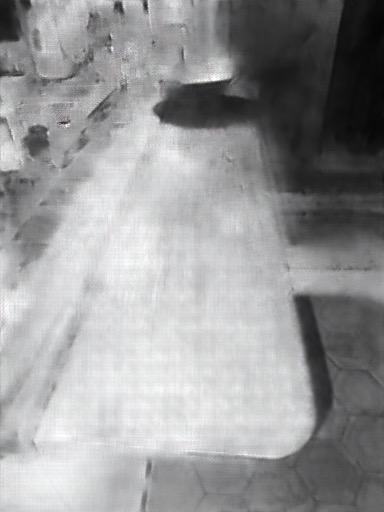}
\includegraphics[width=\linewidth]{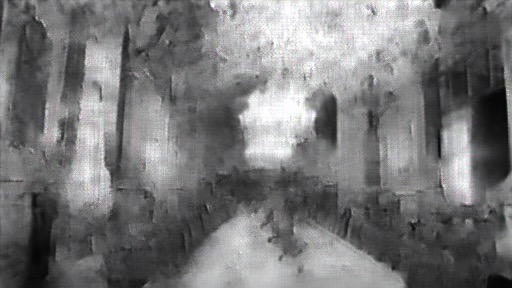}
\includegraphics[width=\linewidth]{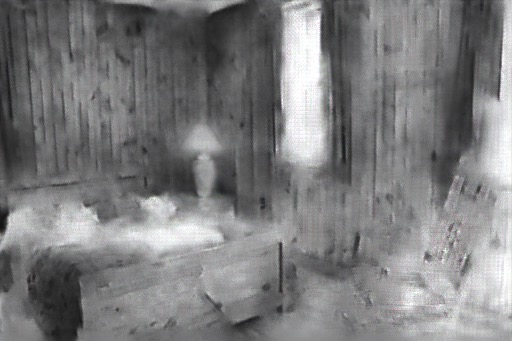}
\includegraphics[width=\linewidth]{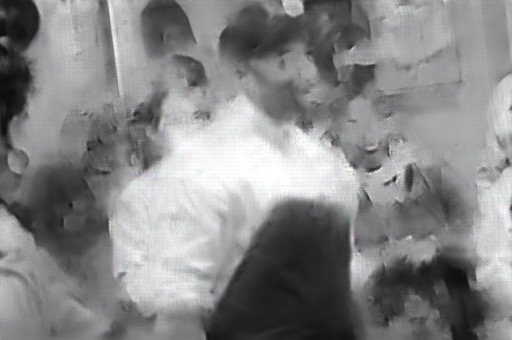}
\includegraphics[width=\linewidth]{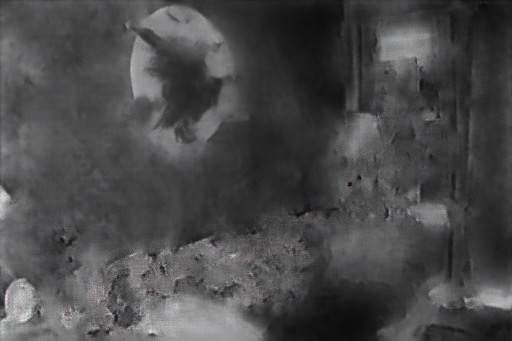}
\includegraphics[width=\linewidth]{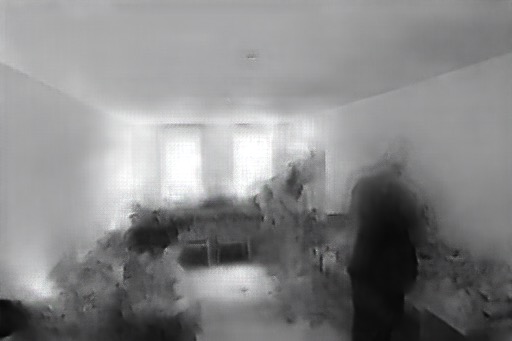}

\caption{Our Shading}
\end{subfigure}
\begin{subfigure}[b]{.18\linewidth}
\includegraphics[width=\linewidth]{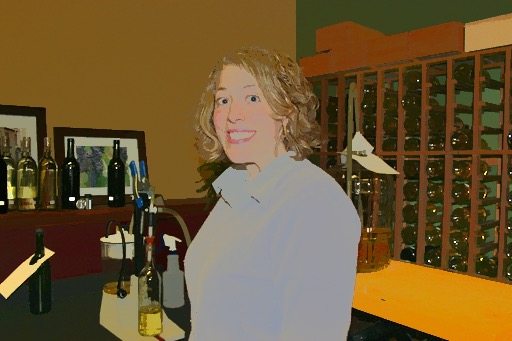}
\includegraphics[width=\linewidth]{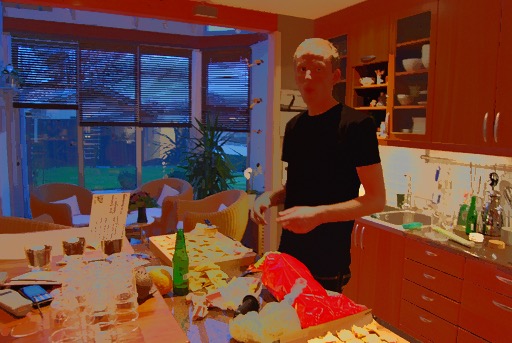}
\includegraphics[width=\linewidth]{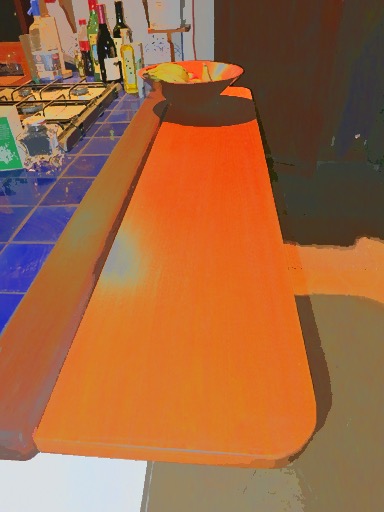}
\includegraphics[width=\linewidth]{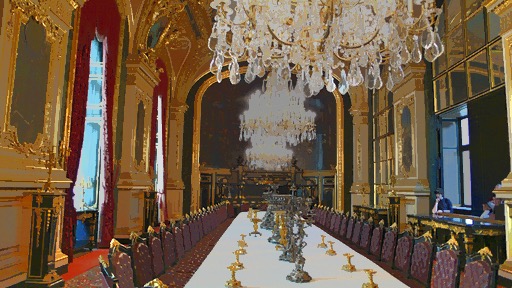}
\includegraphics[width=\linewidth]{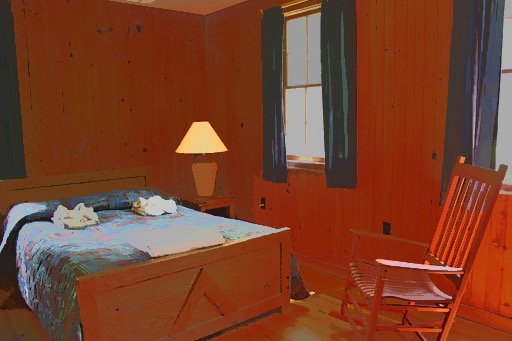}
\includegraphics[width=\linewidth]{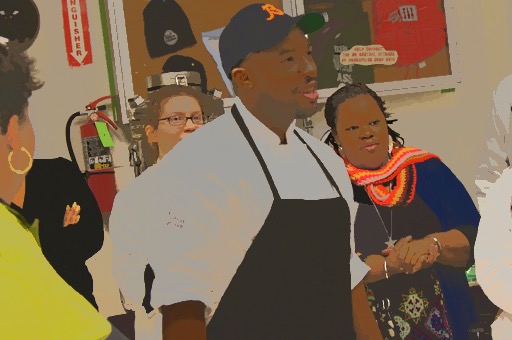}
\includegraphics[width=\linewidth]{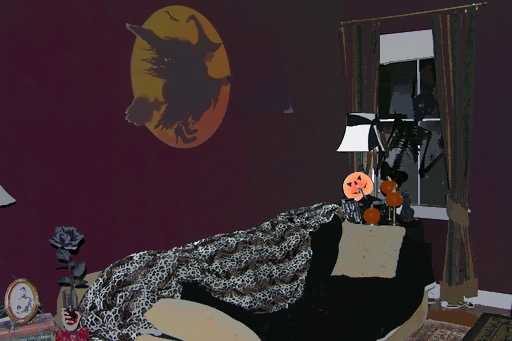}
\includegraphics[width=\linewidth]{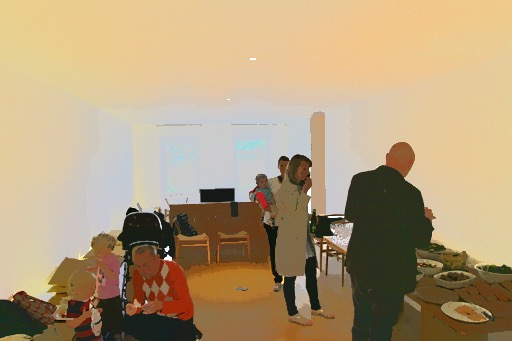}

\caption{Bell et al. Albedo}
\end{subfigure}
\begin{subfigure}[b]{.18\linewidth}
\includegraphics[width=\linewidth]{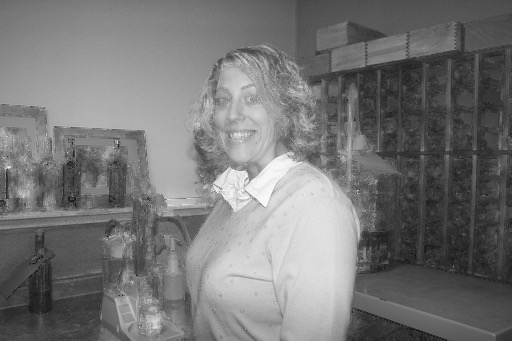}
\includegraphics[width=\linewidth]{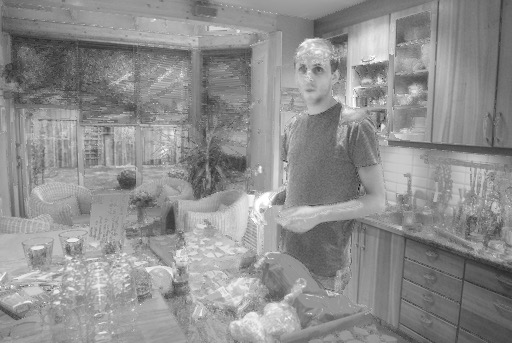}
\includegraphics[width=\linewidth]{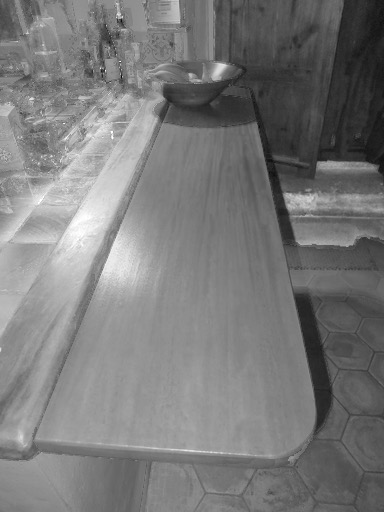}
\includegraphics[width=\linewidth]{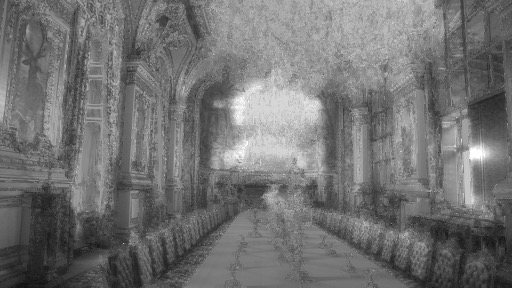}
\includegraphics[width=\linewidth]{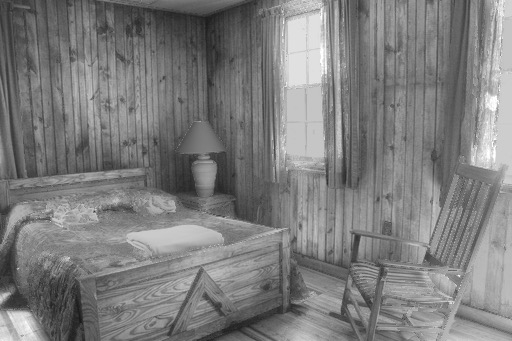}
\includegraphics[width=\linewidth]{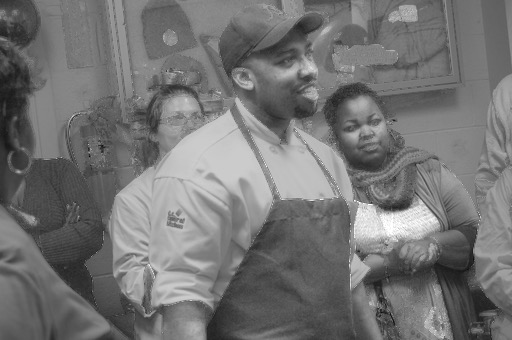}
\includegraphics[width=\linewidth]{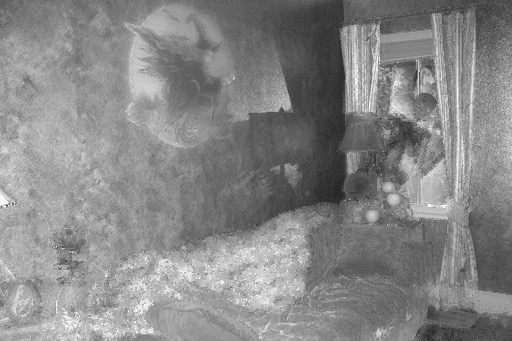}
\includegraphics[width=\linewidth]{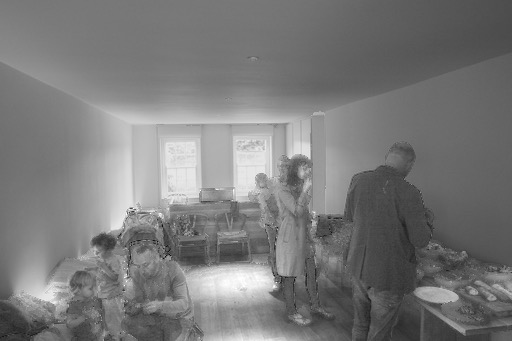}

\caption{Bell et al. Shading}
\end{subfigure}

\caption{We compare the best results (top 5 rows, WHDR = 0.0\%)
    from Bell et al. to our results on the same images.
    We also compare to good results (bottom 3 rows, WHDR = 6.5\%).
    We did not pick any results based on our WHDR scores.
    Notice that our results are occasionally locally better.  For example,
    in the third row, look at the shadow from the bowl, in the last row look
at the bright light on the center of the floor.}
\label{fig:IIW_qual}
\end{figure*}

We decompose images into Albedo and shading using our authored models.
We use the 10-image MIT train set to perform a cursory search of parameters
for weighting the probability models, correlation, and image reconstruction term.
We present quantitative results in table~\ref{tab:MITQuant} on the 10-MIT test dataset
for real color images.  We use the typical scaled MSE measures from \cite{Barron:2013uc}.
Our qualitative results are shown in figure~\ref{fig:MIT_AS_qual}.  We also present quantitative results on WHDR on IIW from \cite{bell14intrinsic} using the same model in table~\ref{tab:iiw_results}.  Since our model is not trained to perform decompositions which
maximize the WHDR measure, we search for an optimal threshold using the first 100 images from
IIW. We compare to decompositions from \cite{bell14intrinsic} in figure~\ref{fig:IIW_qual}.

We set parameters as follows: .0001 for the probability models, 10k for the correlation, and 100
for the image reconstruction term. While the loss we are optimizing is drastically
nonconvex, a trick that we found helpful for determining weights
was to initialize the model near the ground truth decomposition, and look for
weights that were stable at that point.  Another important trick was finding a good
initialization.  For albedo and shading, we initialize shading at a smoothed
version of the image, and albedo to the residual.

\begin{table}[]
\centering
\begin{tabular}{l|l|l|l|}
\cline{2-4}
                                     & S-MSE & R-MSE & RS-MSE \\ \hline
\multicolumn{1}{|l|}{Naive Baseline} & .0577 & .0455 & .0354  \\ \hline
\multicolumn{1}{|l|}{Retinex}        & .0204 & .0186 & .0163  \\ \hline
\multicolumn{1}{|l|}{Gehler et. al. \cite{Gehler:2011vc}}  & .0106 & .0101 & .0131  \\ \hline
\multicolumn{1}{|l|}{Barron et. al. \cite{Barron:2013uc}}  & .0064 & .0098 & .0125  \\ \hline
\multicolumn{1}{|l|}{Ours S, A}           & .0134 & .0175 & .0253  \\ \hline
\multicolumn{1}{|l|}{Ours S+Sd, A}           & .0131& .0160 & .0250  \\ \hline
\end{tabular}

\caption{We compare to intrinsic image techniques on the MIT dataset. Similar to the method
presented in \cite{bell14intrinsic} our method beats Retinex.  We achieve this result
without architecting a solution directly on the decomposition of MIT images as other
prior works have. We perform relatively poorly on RS-MSE because it heavily penalizes
small wiggles in constant value regions.}
\label{tab:MITQuant}
\end{table}

\begin{table}[]
\centering
\begin{tabular}{ll}
    Algorithm          & WHDR \\ \hline
    Bell et al. \cite{bell14intrinsic}        & 21.1 \\
    Zhao et al. \cite{zhao2012closed}       & 23.7 \\
    Garces et al. \cite{garces2012intrinsic} & 25.9 \\
    Retinex (gray) \cite{Grosse:2009ji}    & 27.3 \\
    Retinex (color)    & 27.4 \\
    Ours (Linear .25)  & 27.9 \\
    Shen et al. \cite{shen2011intrinsic}  & 32.4 \\
    Ours (Linear .1)   & 34.9 \\
    Baseline (const R) & 36.6 \\
    Ours (sRGB .1)     & 47.3 \\
    Baseline (const S) & 51.6
\end{tabular}
\caption{Results on IIW.  All other algorithms were tuned for
WHDR with a threshold of .1, ours was not.  As such, we search for
an optimal value for the threshold (.25) on the first 100 images from IIW, and
report results.  Other numbers from \cite{bell14intrinsic}.}
\label{tab:iiw_results}
\end{table}

\begin{figure*}[!h]
\centering
\begin{subfigure}[b]{.14\linewidth}
\includegraphics[width=\linewidth]{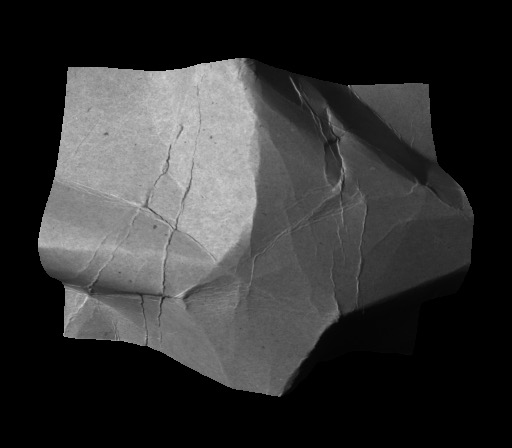}
\includegraphics[width=\linewidth]{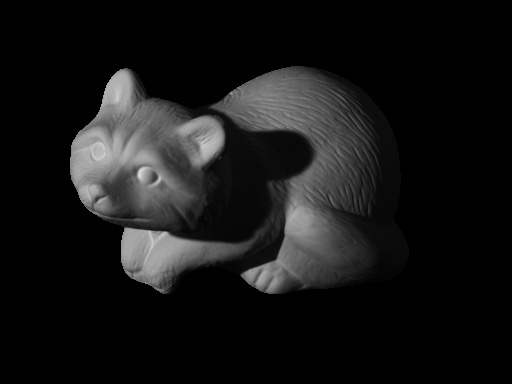}
\includegraphics[width=\linewidth]{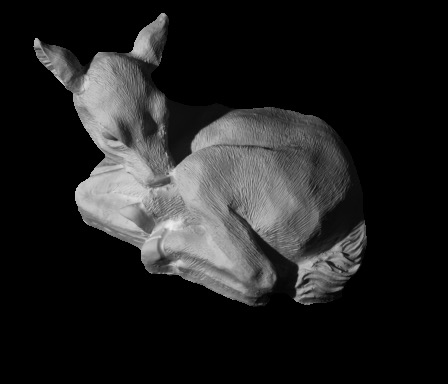}
\caption{Input Image}
\end{subfigure}
\begin{subfigure}[b]{.14\linewidth}
\includegraphics[width=\linewidth]{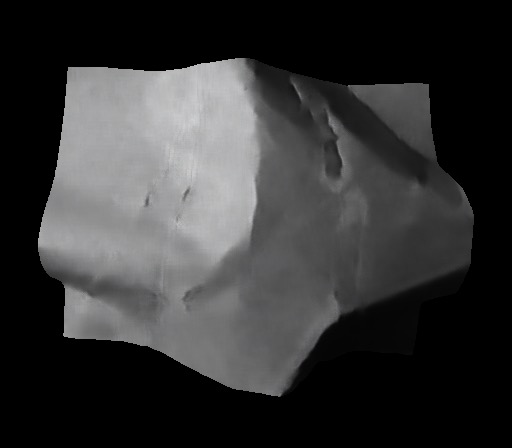}
\includegraphics[width=\linewidth]{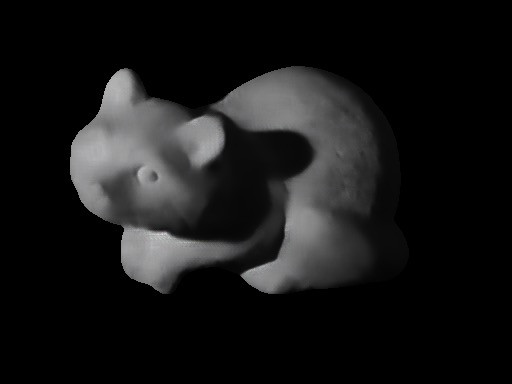}
\includegraphics[width=\linewidth]{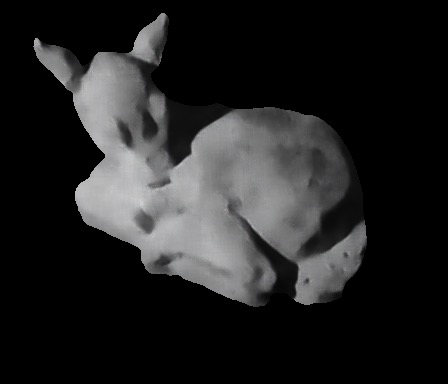}
\caption{Shading}
\end{subfigure}
\begin{subfigure}[b]{.14\linewidth}
\includegraphics[width=\linewidth]{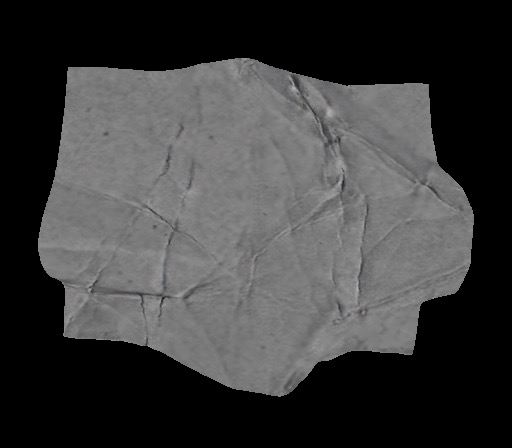}
\includegraphics[width=\linewidth]{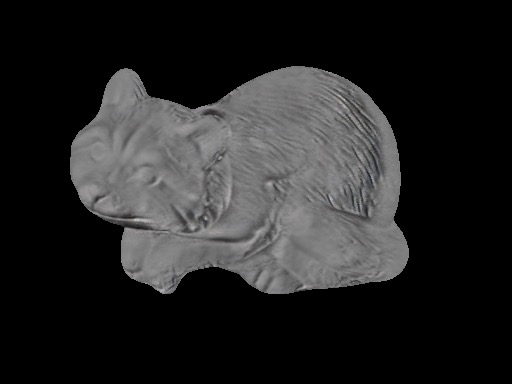}
\includegraphics[width=\linewidth]{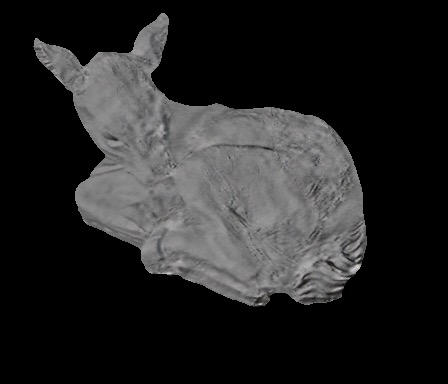}
\caption{Shading Detail}
\end{subfigure}
\begin{subfigure}[b]{.14\linewidth}
\includegraphics[width=\linewidth]{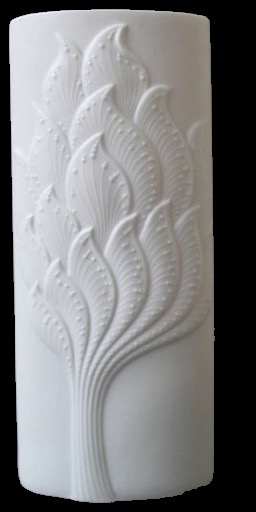}
\caption{Input Image}
\end{subfigure}
\begin{subfigure}[b]{.14\linewidth}
\includegraphics[width=\linewidth]{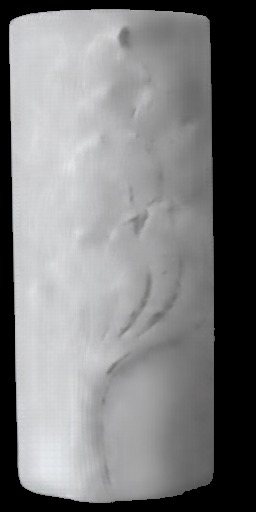}
\caption{Shading}
\end{subfigure}
\begin{subfigure}[b]{.14\linewidth}
\includegraphics[width=\linewidth]{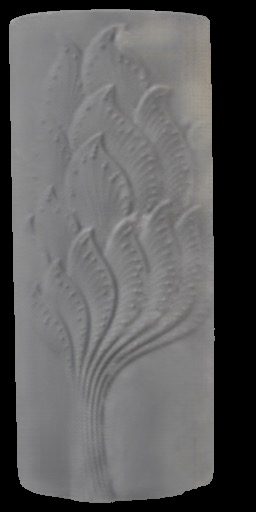}
\caption{Shading Detail}
\end{subfigure}

\caption{Decomposition of shading into shading from shape and shading detail.
Notice that the shading detail recovers texture even in the dark shading
on the raccoon's back.  Notice also that we can decompose real images outside
the MIT dataset with the same model.  All images produced using the same
platonic models, but in different combinations (here S,D).}
\label{fig:MIT_SSD_QUAL}
\end{figure*}

\subsection{Shading and Shading Detail}
Another cool application is decomposing shading into shading and shading detail.
It is especially interesting because, unlike
albedo and shading there is no easy way to capture images of an object with and without 
shading detail.  As such, there are no ground truth decompositions of images into shading
and shading detail.  However, the effect of the decomposition is obvious.  The shading
image captures the global shape of an object, while the shading detail
captures the texture or ``feel'' of the object.  It is also worth noting that it works well on
MIT even when the detail is in deep shadow.  It is also general, and can decompose images
of white generalized cylinders (vases). We show both in figure~\ref{fig:MIT_SSD_QUAL}.  Additional
results on vases can be found in figure~\ref{fig:app_vase}, all results on MIT can be found in the appendix.

The details of our model are similar to the albedo and shading decomposition.
We use
.0001 for the probability weights, 10k for the correlation, and 100 for the image
reconstruction term.  We initialize the shading image to be a smoothed version of the input
image on MIT (the image for vases), and the material to be a constant image.

\subsection{Albedo, Shading, and Shading Detail}
Finally, we can generalize our model to decompose images into three
phenomena at once. We evaluate how well this works on MIT
by composing shading and shading detail to form a single shading
image.  As we can see in table~\ref{tab:MITQuant},
incorporating these three channels improves quantitative performance,
since we accurately determine that shading detail should be attributed to shading.
We show output images in figure~\ref{fig:qual_MIT_asm} and
additional results can be found in the appendix.

We use parameters .0001 for the probability models, 10k for the correlation, and
we have to increase the image reconstruction term to 10000 because the correlation model contains
the contribution from all pairs.

\begin{figure*}[!h]
\centering
\begin{subfigure}{.16\textwidth}
\includegraphics[width=\linewidth]{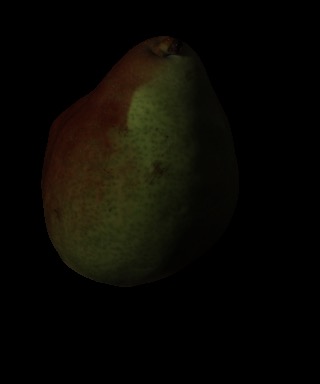}
\includegraphics[width=\linewidth]{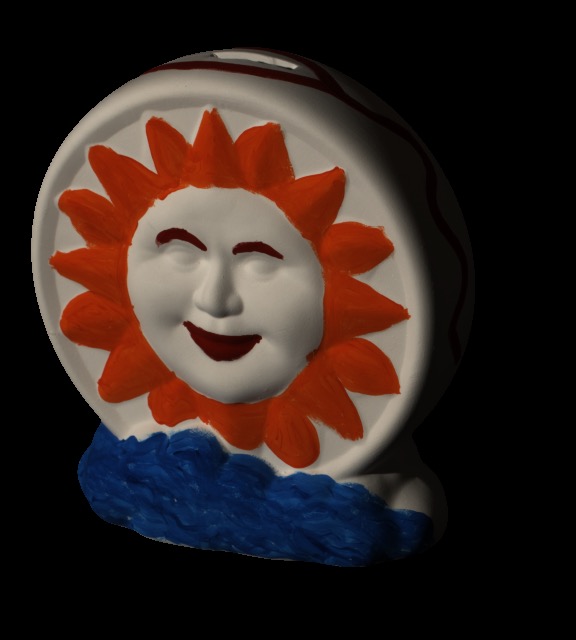}
\includegraphics[width=\linewidth]{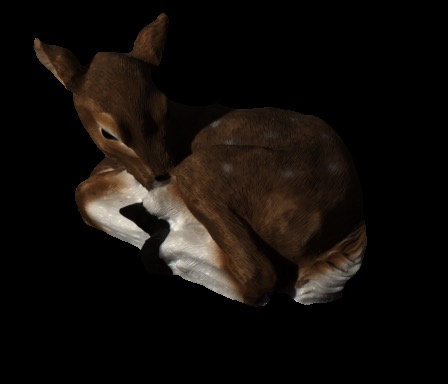}
\caption{Input Image}
\end{subfigure}
\begin{subfigure}{.16\linewidth}
\includegraphics[width=\linewidth]{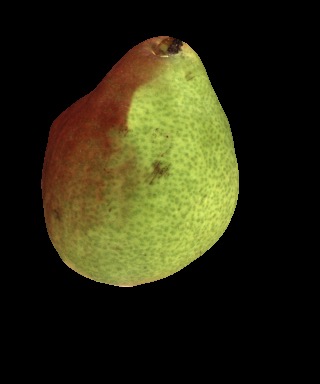}
\includegraphics[width=\linewidth]{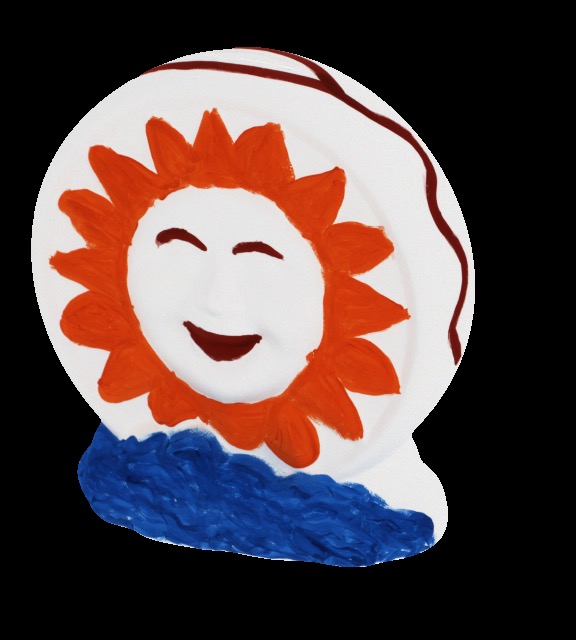}
\includegraphics[width=\linewidth]{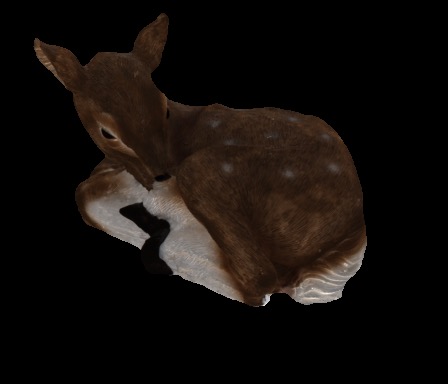}
\caption{GT Albedo}
\end{subfigure}
\begin{subfigure}{.16\linewidth}
\includegraphics[width=\linewidth]{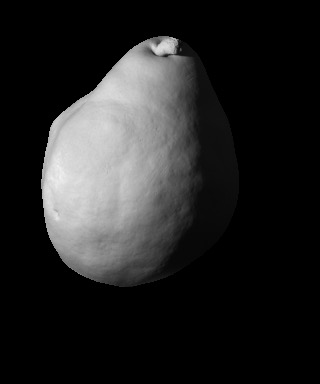}
\includegraphics[width=\linewidth]{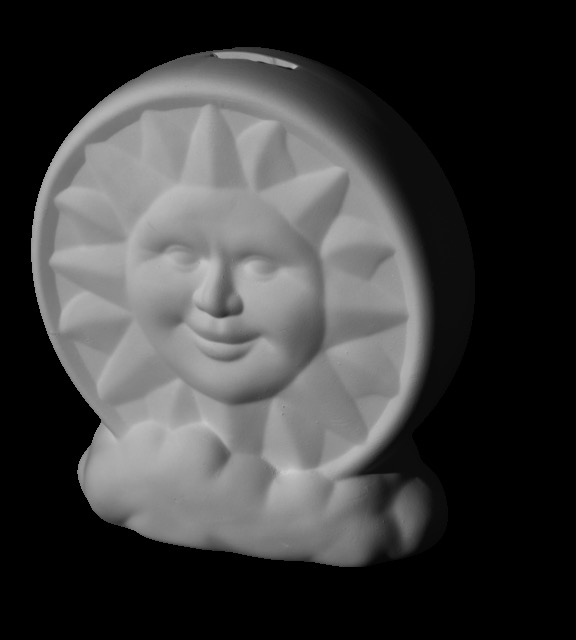}
\includegraphics[width=\linewidth]{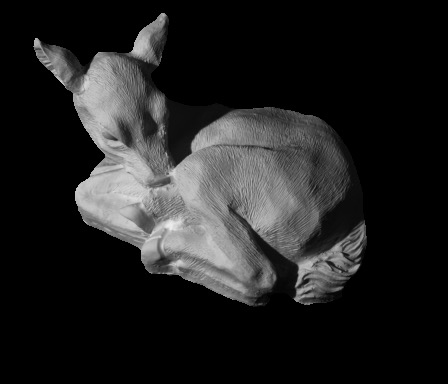}
\caption{GT Shading}
\end{subfigure}
\begin{subfigure}{.16\linewidth}
\includegraphics[width=\linewidth]{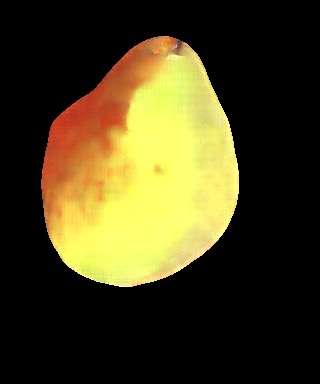}
\includegraphics[width=\linewidth]{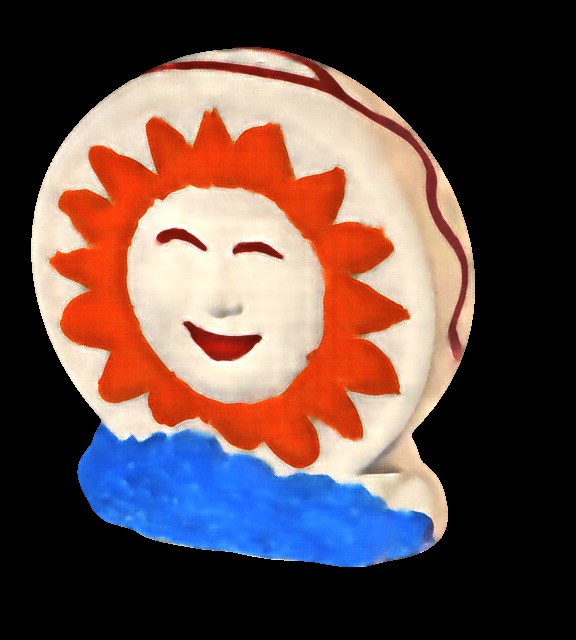}
\includegraphics[width=\linewidth]{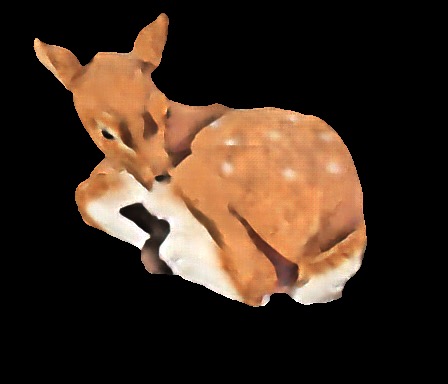}
\caption{Pred Albedo}
\end{subfigure}
\begin{subfigure}{.16\linewidth}
\includegraphics[width=\linewidth]{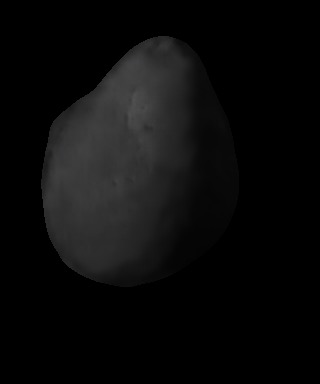}
\includegraphics[width=\linewidth]{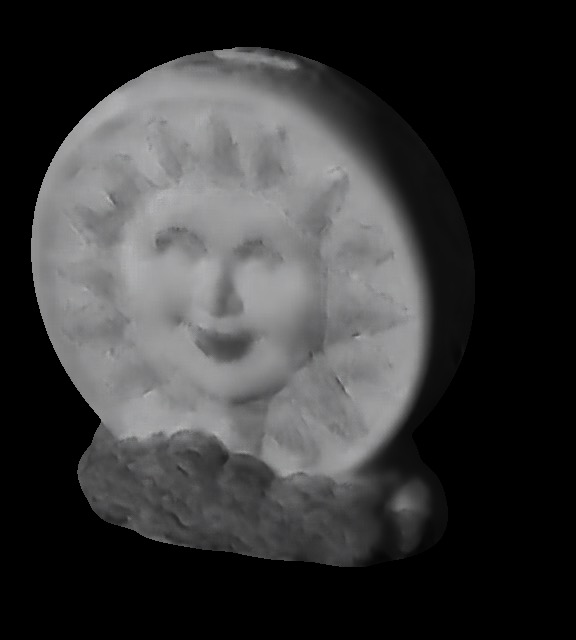}
\includegraphics[width=\linewidth]{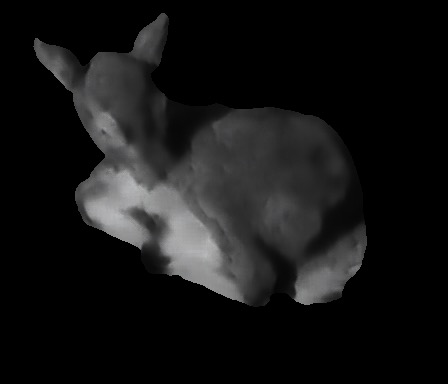}
\caption{Pred Shading}
\end{subfigure}
\begin{subfigure}{.16\linewidth}
\includegraphics[width=\linewidth]{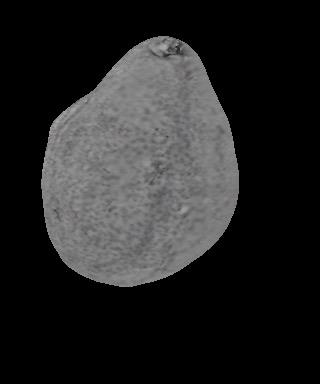}
\includegraphics[width=\linewidth]{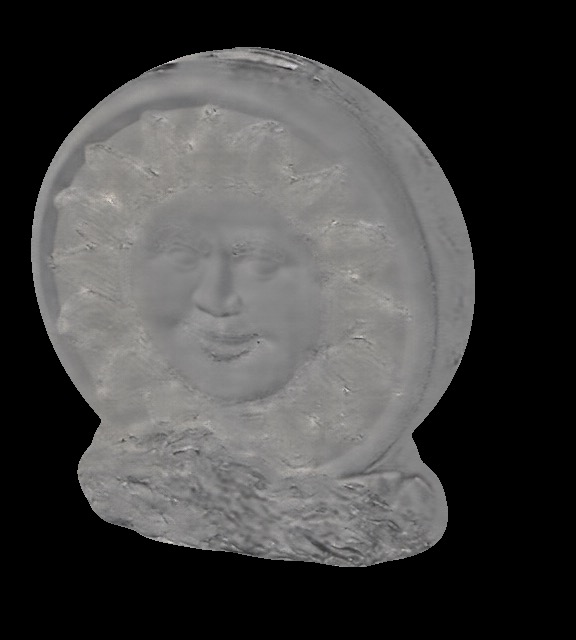}
\includegraphics[width=\linewidth]{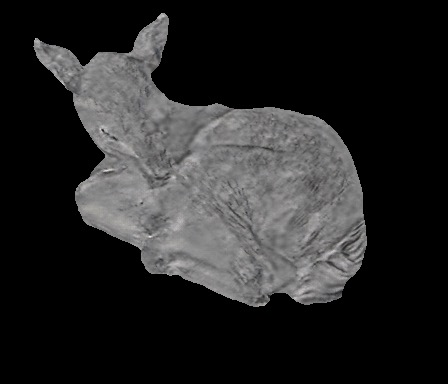}
\caption{Shading Detail}
\end{subfigure}
\caption{Our model can extend to decompose three layers: shading, detail, and albedo.
The outputs are qualitatively very good, especially the albedos which are generally constant
and the shading details which take small bumpy details. Best viewed in high resolution in color.
All images are produced using the same platonic models, but in different combinations (here A,S,D).
}
\label{fig:qual_MIT_asm}
\end{figure*}

\section{Discussion}

We have demonstrated a method for decomposing images into component channels using authored models to determine the channels.
We represent albedo (resp. shading, shading detail) by the signals that a learned generative model is capable of producing.
We obtain strong performance, because our models use a novel pyramid conv-VAE architecture that can
reproduce fine image detail.  A particular attraction of our approach is that it applies in the absence of canonical or ground-truth
decompositions of images; instead, one provides platonic ideals for the layers envisaged.  In the future we want to explore the tradeoffs between bias and variance in authoring the models, either
in adjustment of the conv-VAE hyperparameters, or choosing different sets of images as our platonic
examples. We plan to investigate decomposing images into different layers, possibly even layers that are not
compositionally related to the image. Introducing other losses on the produced images could also
lead to improvements in decomposition, for example, a bias for specific colors in albedos.
Conv-VAE's involve a compromise; it is hard to sample a conv-VAE, because codes at each pixel
are not independent. In future, we will investigate sampling conv-VAE's to produce a fully generative model
that can produce high resolution images.

%

{\small
\bibliographystyle{ieee}
\bibliography{bib_all}
}

\clearpage

\appendix
\begin{appendices}
\section{VAE Architectures}
Our VAE and conv-VAE architectures used for comparison are described in detail in table~\ref{tab:vae_archs}.

\section{Comparison to Isola et. al.}
\begin{figure*}
\centering
\begin{subfigure}[b]{.15\linewidth}
\includegraphics[width=\linewidth]{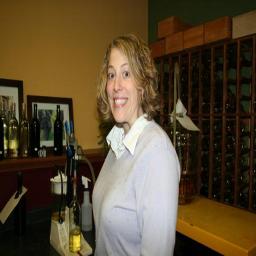}
\includegraphics[width=\linewidth]{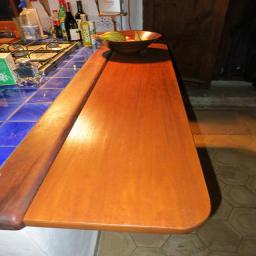}
\includegraphics[width=\linewidth]{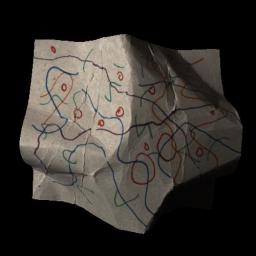}
\includegraphics[width=\linewidth]{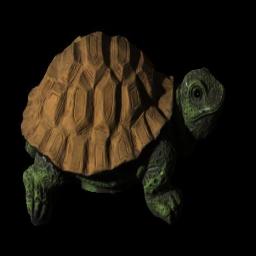}
\caption{Input Image}
\end{subfigure}
\begin{subfigure}[b]{.15\linewidth}
\includegraphics[width=\linewidth]{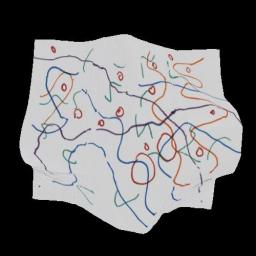}
\includegraphics[width=\linewidth]{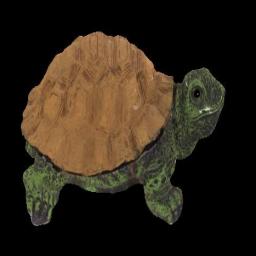}
\caption{Albedo GT}
\end{subfigure}
\begin{subfigure}[b]{.15\linewidth}
\includegraphics[width=\linewidth]{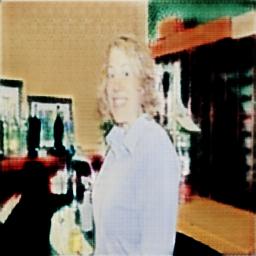}
\includegraphics[width=\linewidth]{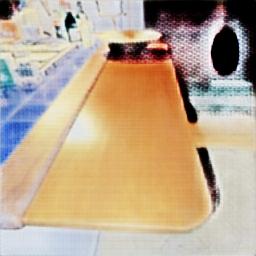}
\includegraphics[width=\linewidth]{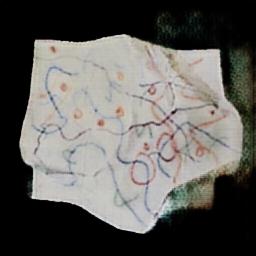}
\includegraphics[width=\linewidth]{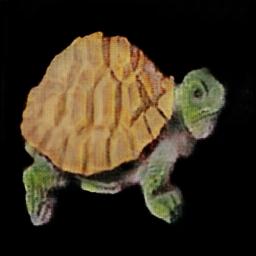}
\caption{Albedo Pred}
\end{subfigure}
\begin{subfigure}[b]{.15\linewidth}
\includegraphics[width=\linewidth]{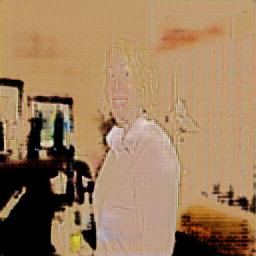}
\includegraphics[width=\linewidth]{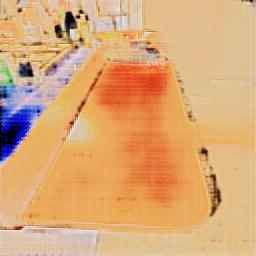}
\includegraphics[width=\linewidth]{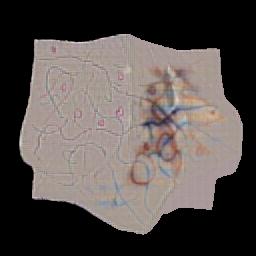}
\includegraphics[width=\linewidth]{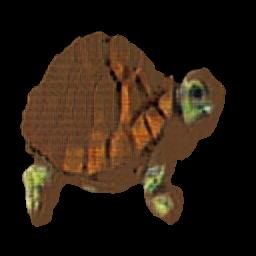}
\caption{Patch Pred}
\end{subfigure}
\begin{subfigure}[b]{.15\linewidth}
\includegraphics[width=\linewidth, height=\linewidth]{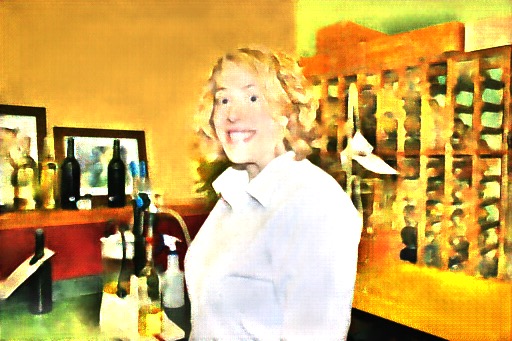}
\includegraphics[width=\linewidth, height=\linewidth]{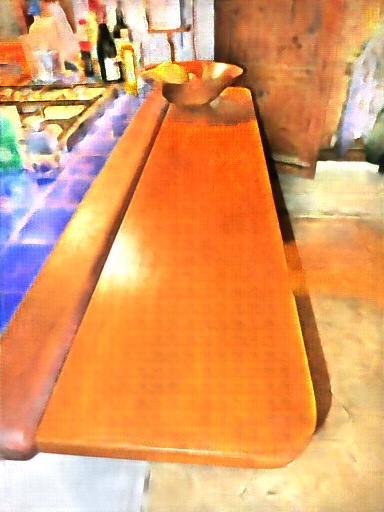}
\includegraphics[width=\linewidth, height=\linewidth]{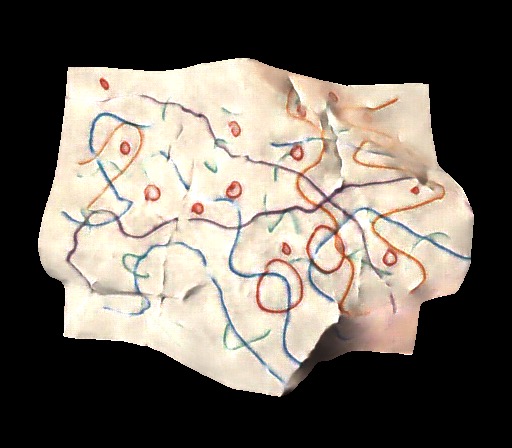}
\includegraphics[width=\linewidth, height=\linewidth]{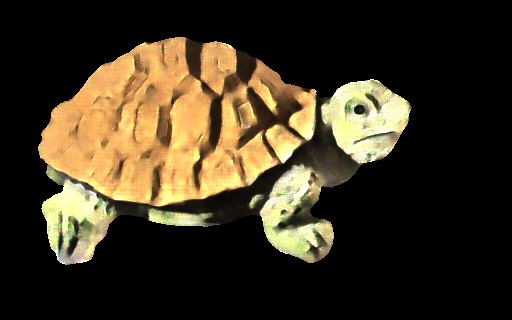}
\caption{Ours}
\end{subfigure}
\caption{While Isola et. al. \cite{pdf:dvut8KmF} performs well on MIT when trained with MIT (rows 3 and 4, column 3).  However, the model does not generalize well to IIW as small details on the wine rack or table are eliminated (rows 1 and 2).
When trained on images composed from Mondrian and shading images,
the results are poor on both IIW and MIT as colors bleed across boundaries (column d).  Our result is shown in column e for comparison.}
\label{fig:pix2pix}
\end{figure*}

Image-to-Image translation \cite{pdf:dvut8KmF}, a concurrent paper with ours, demonstrates
how to predict images from other images.  We evaluate image-to-image translation on MIT and IIW
qualitatively in figure~\ref{fig:pix2pix}. We find that the model is capable of learning MIT
albedo prediction when trained on MIT, but the model does not generalize
to IIW.  

\begin{figure}
\centering
\begin{subfigure}[b]{.25\linewidth}
\includegraphics[width=\linewidth]{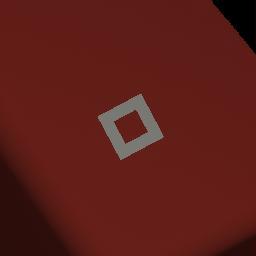}
\includegraphics[width=\linewidth]{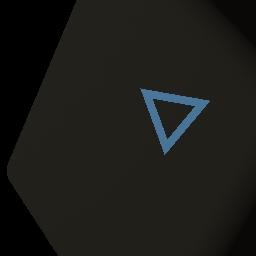}
\includegraphics[width=\linewidth]{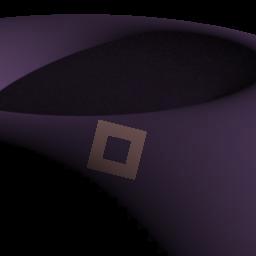}
\includegraphics[width=\linewidth]{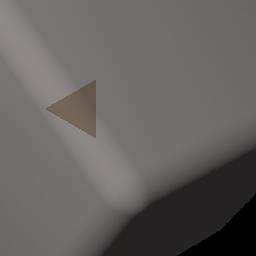}
\includegraphics[width=\linewidth]{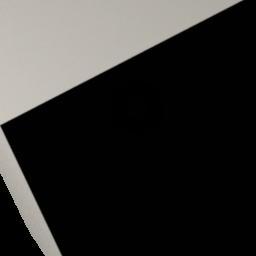}
\includegraphics[width=\linewidth]{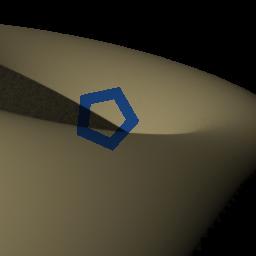}
\includegraphics[width=\linewidth]{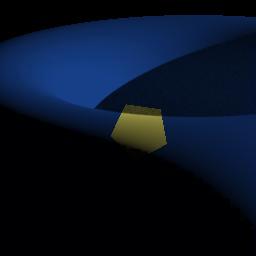}
\caption{input image}
\end{subfigure}
\begin{subfigure}[b]{.25\linewidth}
\includegraphics[width=\linewidth]{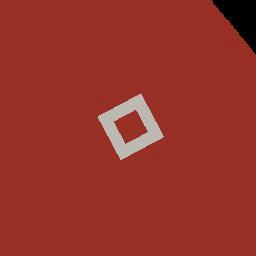}
\includegraphics[width=\linewidth]{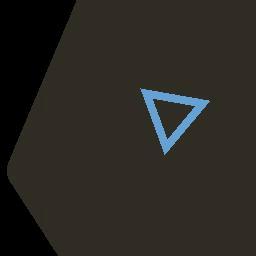}
\includegraphics[width=\linewidth]{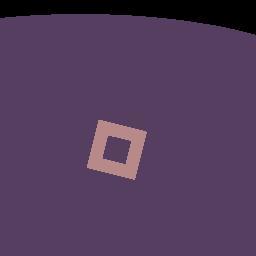}
\includegraphics[width=\linewidth]{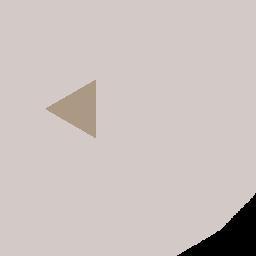}
\includegraphics[width=\linewidth]{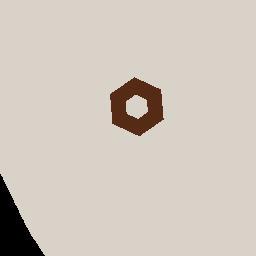}
\includegraphics[width=\linewidth]{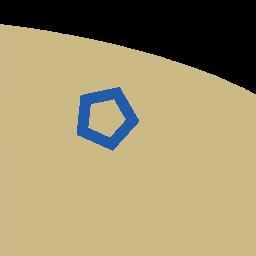}
\includegraphics[width=\linewidth]{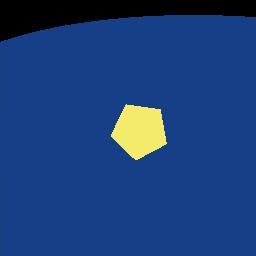}
\caption{GT Albedo}
\end{subfigure}
\begin{subfigure}[b]{.25\linewidth}
\includegraphics[width=\linewidth]{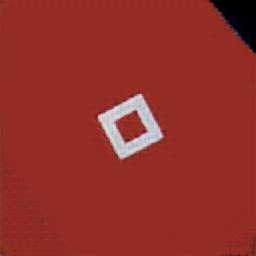}
\includegraphics[width=\linewidth]{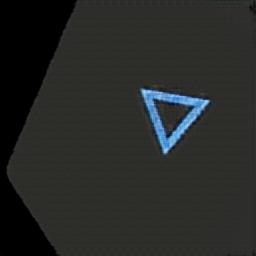}
\includegraphics[width=\linewidth]{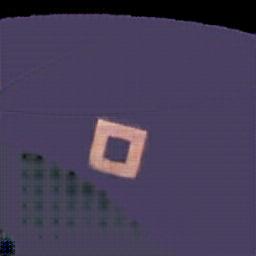}
\includegraphics[width=\linewidth]{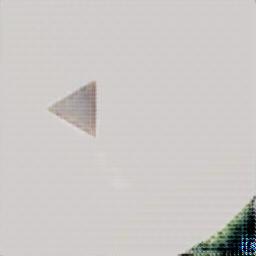}
\includegraphics[width=\linewidth]{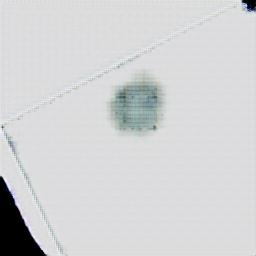}
\includegraphics[width=\linewidth]{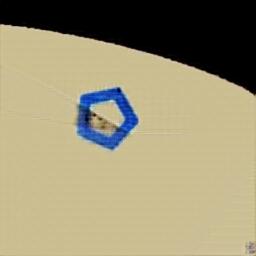}
\includegraphics[width=\linewidth]{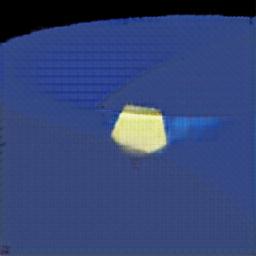}
\caption{Prediction}
\end{subfigure}
\caption{Using Image-to-Image \cite{pdf:dvut8KmF} to learn to predict Mondrian albedos
during training works well.  However, as shown in figure~\ref{fig:pix2pix}, this
model does not generalize well to real images.}
\label{fig:pix2pix_mondrian}
\end{figure}

We also train an image-to-image prediction model on our authored data
by creating shaded Mondrian images.  These are formed by compositing Mondrians and shading images.
As shown in figure~\ref{fig:pix2pix_mondrian}, the model can predict the Mondrian albedos,
but the model does not generalize to predicting albedos on MIT or IIW (figure~\ref{fig:pix2pix}).

\begin{table*}[]
\centering
\begin{tabular}{l|ll|ll|ll|llll|}
\cline{2-11}
      & \multicolumn{2}{c|}{VAE-1}                              & \multicolumn{2}{c|}{VAE-2}                              & \multicolumn{2}{c|}{conv-VAE}                           & \multicolumn{4}{c|}{Laplacian conv-VAE}                    \\ \cline{2-11} 
      & \multicolumn{1}{c}{Filter} & \multicolumn{1}{c|}{Layer} & \multicolumn{1}{c}{Filter} & \multicolumn{1}{c|}{Layer} & \multicolumn{1}{c}{Filter} & \multicolumn{1}{c|}{Layer} & \multicolumn{1}{c}{Filter} & \multicolumn{3}{c|}{Layer}    \\
\rowcolor[HTML]{BBDAFF} 
Image &                            & 64x64x3                    &                            & 64x64x3                    &                            & 64x64x3                    &                            & 64x64x3  & 32x32x3  & 16x16x3 \\
      & 5x5                        & 32x32x64                   & 5x5                        & 32x32x64                   & 5x5                        & 32x32x64                   & 5x5                        & 32x32x64 & 16x16x64 & 8x8x64  \\
      & 5x5                        & 16x16x128                  & 5x5                        & 16x16x128                  & 5x5                        & 16x16x128                  & 5x5                        & 16x16x64 & 8x8x64   & 4x4x64  \\
      & 3x3                        & 16x16x128                  & 3x3                        & 16x16x128                  & 3x3                        & 16x16x128                  & 3x3                        & 16x16x64 & 8x8x64   & 4x4x64  \\
      & fc                         & 128                        & fc                         & 4096                       & 4x4                        & 13x13x20                   & 4x4                        & 13x13x16 & 5x5x16   & 1x1x16  \\
\rowcolor[HTML]{B8D8A4}
Code  & fc                         & 32                         & fc                         & 832                        & 1x1                        & 13x13x8                    & 1x1                        & 13x13x4  & 5x5x4    & 1x1x4   \\ \cline{2-11} 
\end{tabular}

\caption{Architecture for the encoding portion of the VAE, conv-VAE, and Laplacian conv-VAE for comparison.  The decoding portion is
the same in reverse. For a fair comparison, we compare two VAEs.  VAE-1 has roughly the same number of parameters
as the conv-VAE, while VAE-2 has the same code dimensionality, but drastically more parameters.
When converting from Laplacian conv-VAE to conv-VAE we matched the intermediate sizes when possible,
but always rounded up when necessary.}
\label{tab:vae_archs}
\end{table*}

\section{Albedo Shading on MIT}
\begin{figure*}
\centering
\begin{subfigure}[b]{.12\linewidth}
\includegraphics[width=\linewidth]{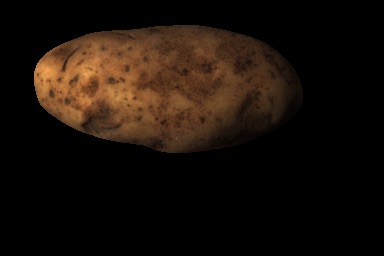}
\includegraphics[width=\linewidth]{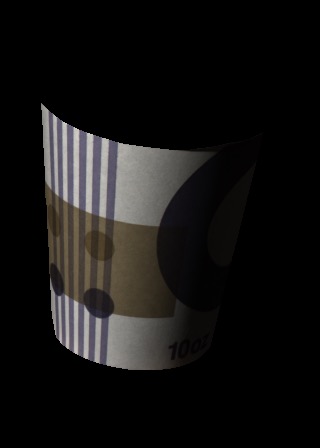}
\includegraphics[width=\linewidth]{Figures/AlbedoShading/im_2_99_0}
\includegraphics[width=\linewidth]{Figures/AlbedoShading/im_3_99_0}
\includegraphics[width=\linewidth]{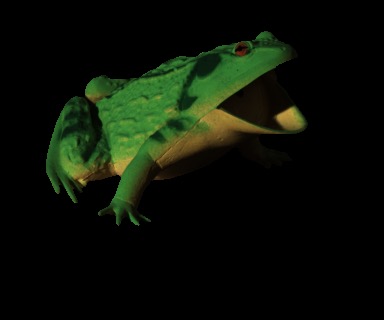}
\includegraphics[width=\linewidth]{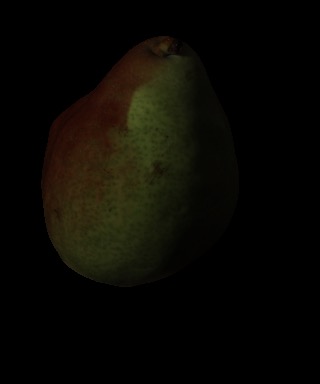}
\includegraphics[width=\linewidth]{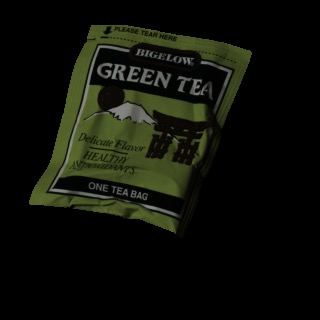}
\includegraphics[width=\linewidth]{Figures/AlbedoShading/im_7_99_0}
\includegraphics[width=\linewidth]{Figures/AlbedoShading/im_8_99_0}
\includegraphics[width=\linewidth]{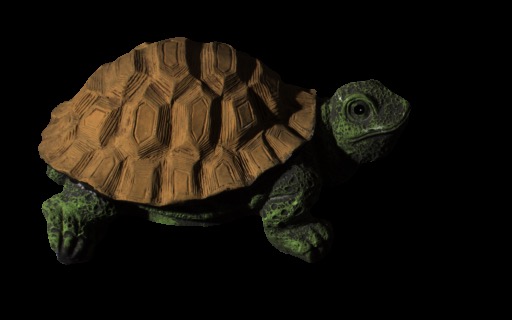}
\caption{Input Image}
\end{subfigure}
\begin{subfigure}[b]{.12\linewidth}
\includegraphics[width=\linewidth]{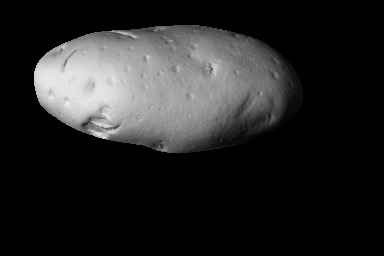}
\includegraphics[width=\linewidth]{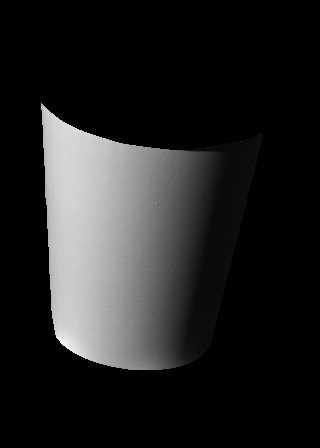}
\includegraphics[width=\linewidth]{Figures/AlbedoShading/shading_gt_2_99_0}
\includegraphics[width=\linewidth]{Figures/AlbedoShading/shading_gt_3_99_0}
\includegraphics[width=\linewidth]{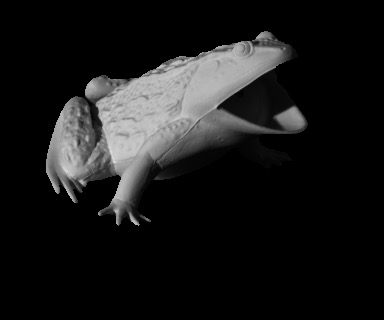}
\includegraphics[width=\linewidth]{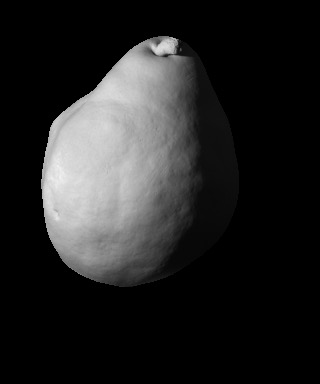}
\includegraphics[width=\linewidth]{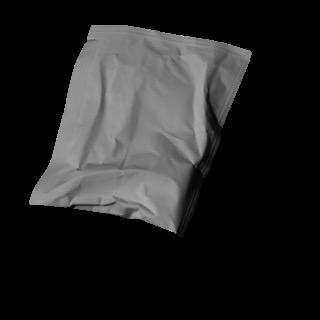}
\includegraphics[width=\linewidth]{Figures/AlbedoShading/shading_gt_7_99_0}
\includegraphics[width=\linewidth]{Figures/AlbedoShading/shading_gt_8_99_0}
\includegraphics[width=\linewidth]{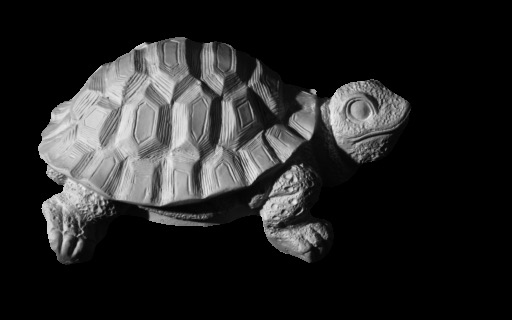}
\caption{Shading GT}
\end{subfigure}
\begin{subfigure}[b]{.12\linewidth}
\includegraphics[width=\linewidth]{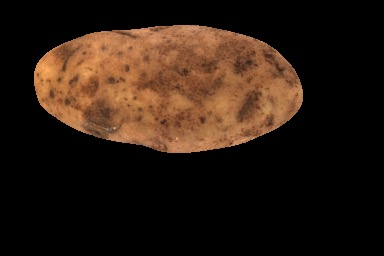}
\includegraphics[width=\linewidth]{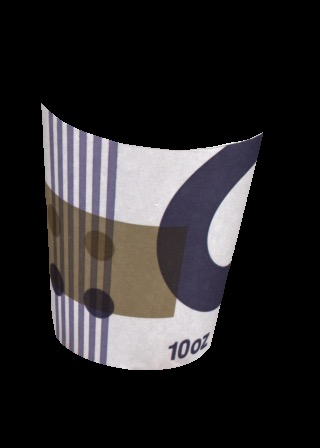}
\includegraphics[width=\linewidth]{Figures/AlbedoShading/albedo_gt_2_99_0}
\includegraphics[width=\linewidth]{Figures/AlbedoShading/albedo_gt_3_99_0}
\includegraphics[width=\linewidth]{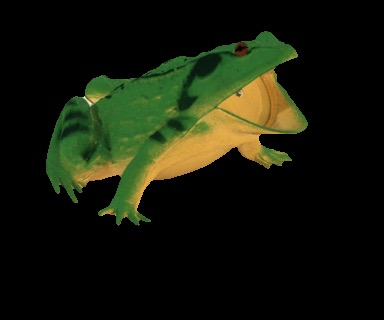}
\includegraphics[width=\linewidth]{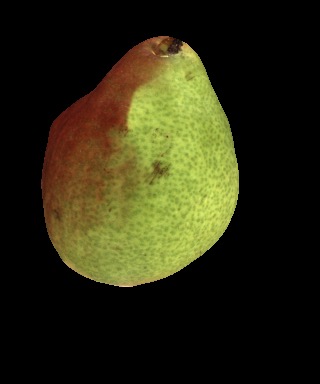}
\includegraphics[width=\linewidth]{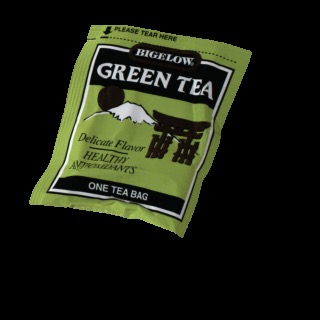}
\includegraphics[width=\linewidth]{Figures/AlbedoShading/albedo_gt_7_99_0}
\includegraphics[width=\linewidth]{Figures/AlbedoShading/albedo_gt_8_99_0}
\includegraphics[width=\linewidth]{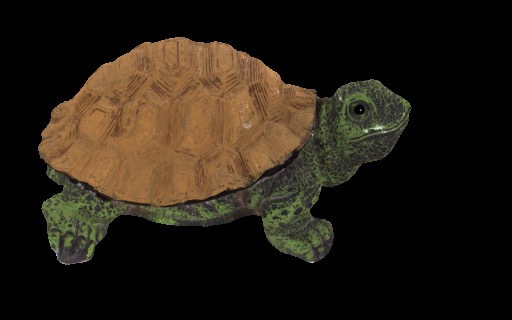}
\caption{Albedo GT}
\end{subfigure}
\begin{subfigure}[b]{.12\linewidth}
\includegraphics[width=\linewidth]{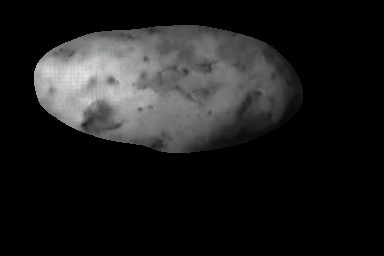}
\includegraphics[width=\linewidth]{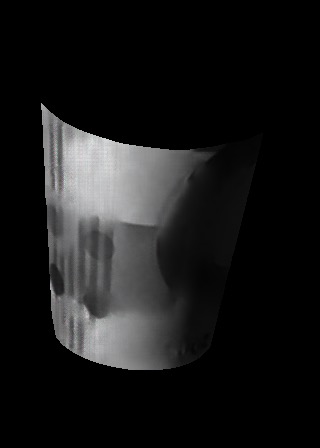}
\includegraphics[width=\linewidth]{Figures/AlbedoShading/shading_2_99_0}
\includegraphics[width=\linewidth]{Figures/AlbedoShading/shading_3_99_0}
\includegraphics[width=\linewidth]{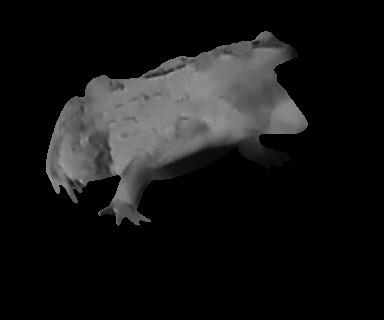}
\includegraphics[width=\linewidth]{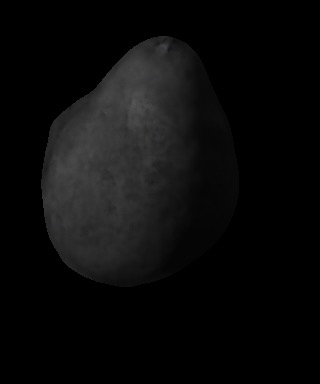}
\includegraphics[width=\linewidth]{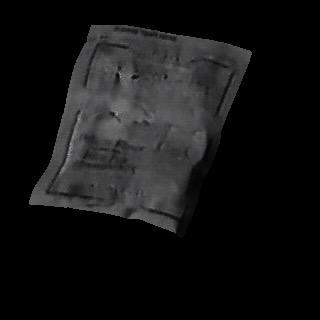}
\includegraphics[width=\linewidth]{Figures/AlbedoShading/shading_7_99_0}
\includegraphics[width=\linewidth]{Figures/AlbedoShading/shading_8_99_0}
\includegraphics[width=\linewidth]{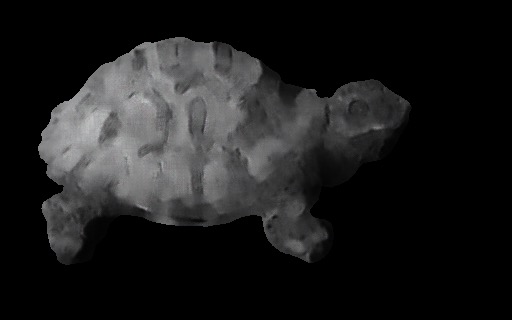}
\caption{Shading}
\end{subfigure}
\begin{subfigure}[b]{.12\linewidth}
\includegraphics[width=\linewidth]{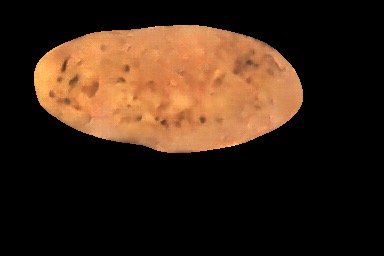}
\includegraphics[width=\linewidth]{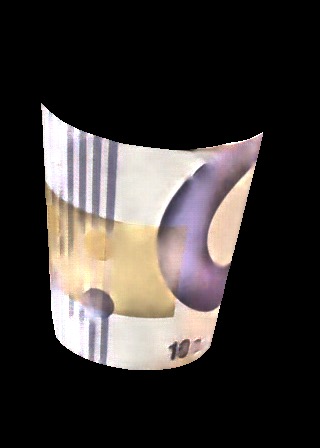}
\includegraphics[width=\linewidth]{Figures/AlbedoShading/albedo_2_99_0}
\includegraphics[width=\linewidth]{Figures/AlbedoShading/albedo_3_99_0}
\includegraphics[width=\linewidth]{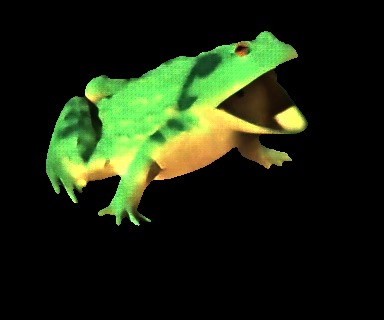}
\includegraphics[width=\linewidth]{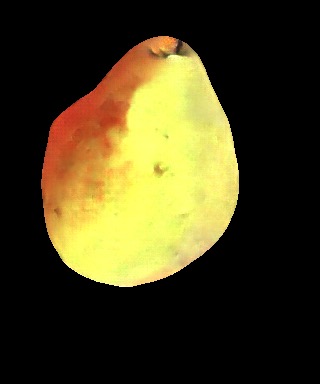}
\includegraphics[width=\linewidth]{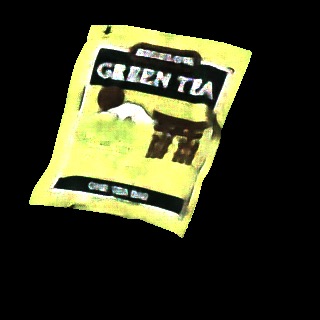}
\includegraphics[width=\linewidth]{Figures/AlbedoShading/albedo_7_99_0}
\includegraphics[width=\linewidth]{Figures/AlbedoShading/albedo_8_99_0}
\includegraphics[width=\linewidth]{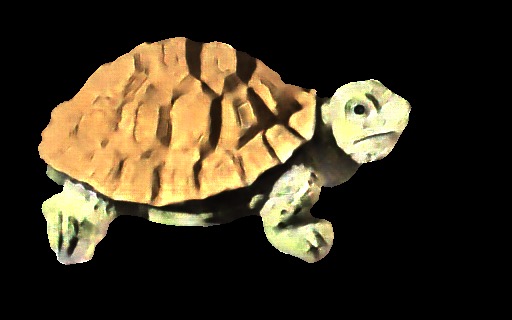}
\caption{Albedo}
\end{subfigure}
\caption{Albedo and Shading decompositions on all images from MIT Test for our method.}
\label{fig:mit_albedo_shading_test}
\end{figure*}

All decompositions for MIT test are shown in figure~\ref{fig:mit_albedo_shading_test}.

\section{Albedo, Shading, and Shading Detail on MIT}
\begin{figure*}
\centering
\begin{subfigure}[b]{.12\linewidth}
\includegraphics[width=\linewidth]{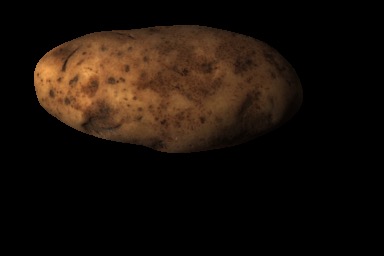}
\includegraphics[width=\linewidth]{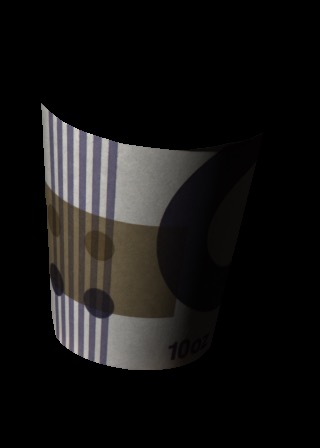}
\includegraphics[width=\linewidth]{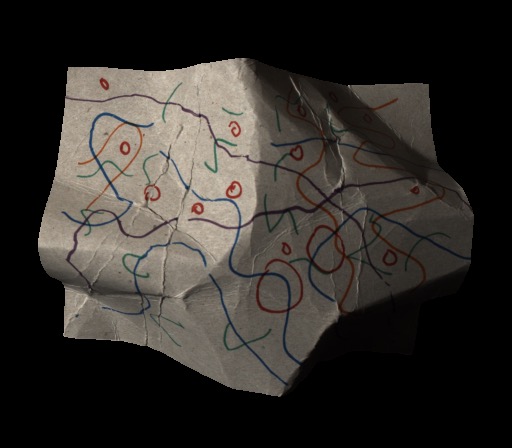}
\includegraphics[width=\linewidth]{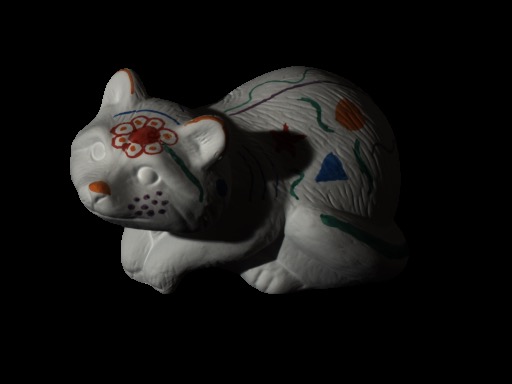}
\includegraphics[width=\linewidth]{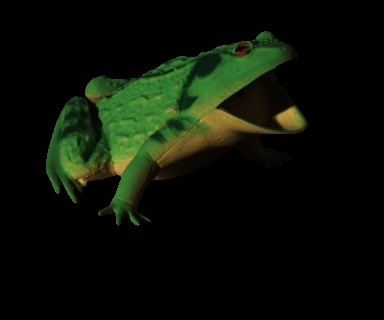}
\includegraphics[width=\linewidth]{Figures/AMSImages/test/im_5_99_0}
\includegraphics[width=\linewidth]{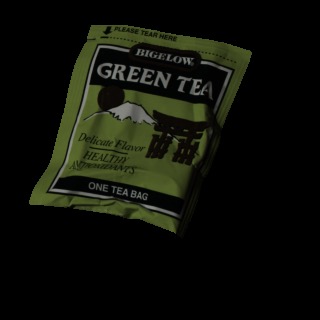}
\includegraphics[width=\linewidth]{Figures/AMSImages/test/im_7_99_0}
\includegraphics[width=\linewidth]{Figures/AMSImages/test/im_8_99_0}
\includegraphics[width=\linewidth]{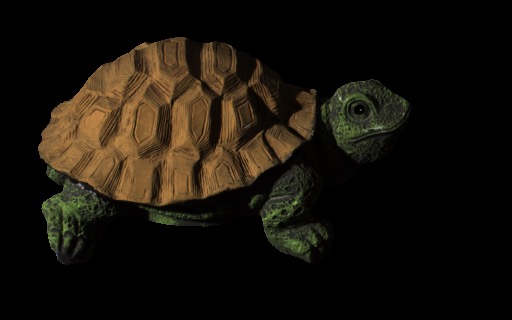}
\caption{Input Image}
\end{subfigure}
\begin{subfigure}[b]{.12\linewidth}
\includegraphics[width=\linewidth]{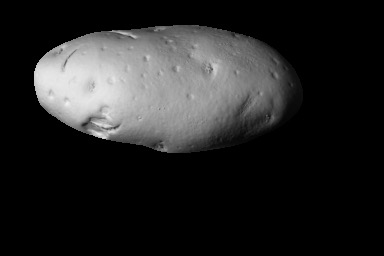}
\includegraphics[width=\linewidth]{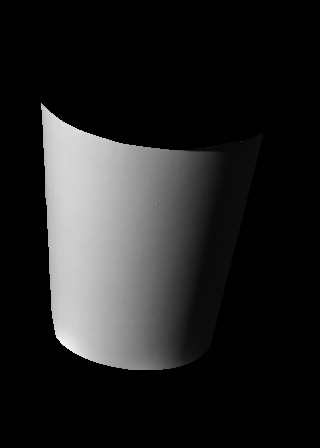}
\includegraphics[width=\linewidth]{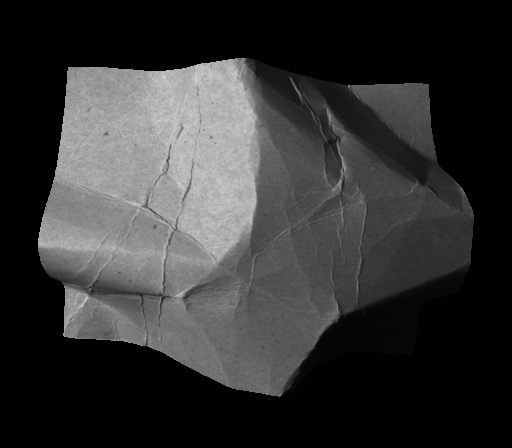}
\includegraphics[width=\linewidth]{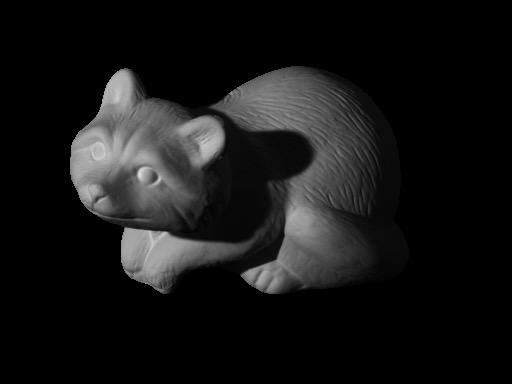}
\includegraphics[width=\linewidth]{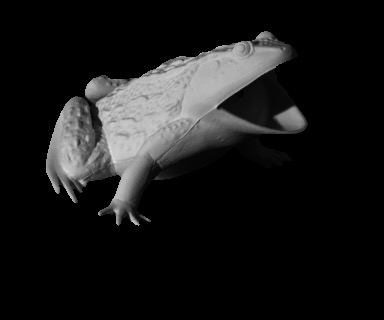}
\includegraphics[width=\linewidth]{Figures/AMSImages/test/shading_gt_5_99_0}
\includegraphics[width=\linewidth]{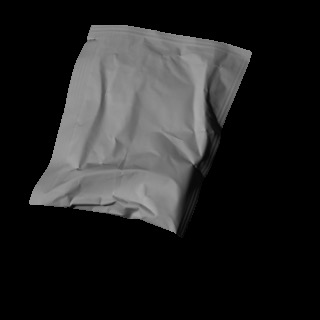}
\includegraphics[width=\linewidth]{Figures/AMSImages/test/shading_gt_7_99_0}
\includegraphics[width=\linewidth]{Figures/AMSImages/test/shading_gt_8_99_0}
\includegraphics[width=\linewidth]{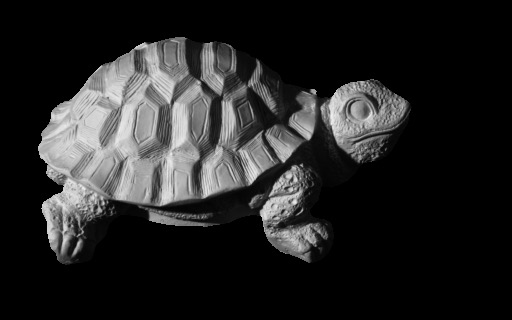}
\caption{Shading GT}
\end{subfigure}
\begin{subfigure}[b]{.12\linewidth}
\includegraphics[width=\linewidth]{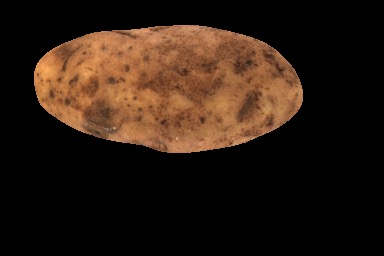}
\includegraphics[width=\linewidth]{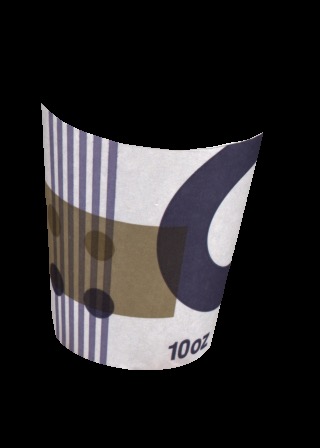}
\includegraphics[width=\linewidth]{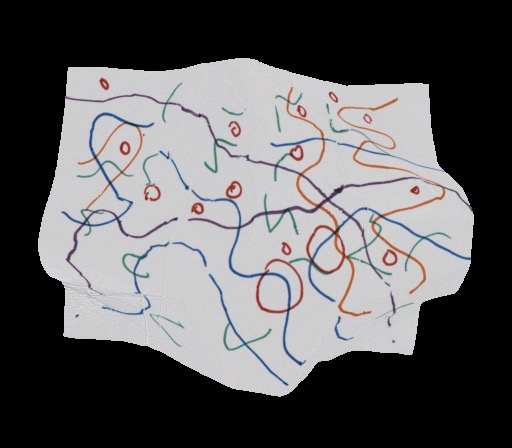}
\includegraphics[width=\linewidth]{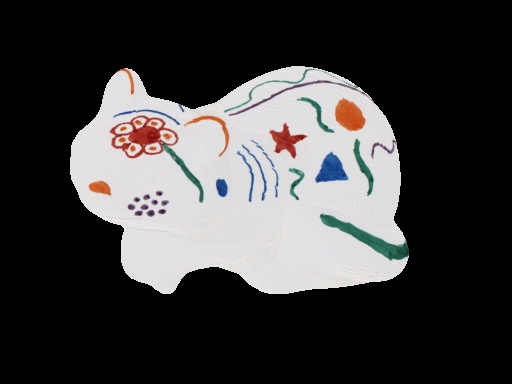}
\includegraphics[width=\linewidth]{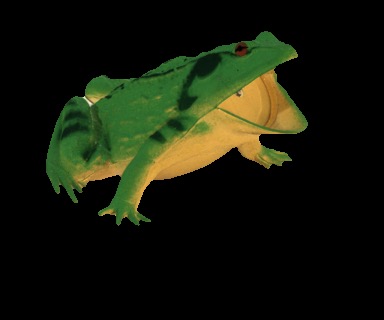}
\includegraphics[width=\linewidth]{Figures/AMSImages/test/albedo_gt_5_99_0}
\includegraphics[width=\linewidth]{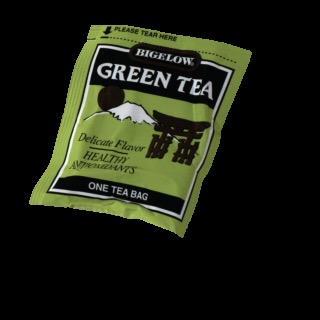}
\includegraphics[width=\linewidth]{Figures/AMSImages/test/albedo_gt_7_99_0}
\includegraphics[width=\linewidth]{Figures/AMSImages/test/albedo_gt_8_99_0}
\includegraphics[width=\linewidth]{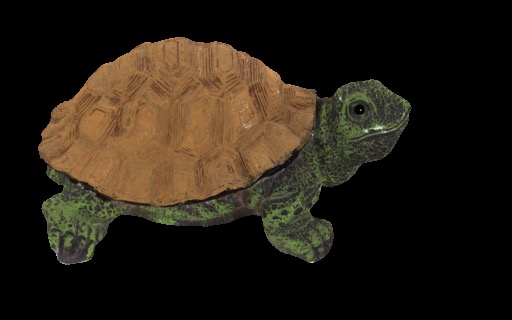}
\caption{Albedo GT}
\end{subfigure}
\begin{subfigure}[b]{.12\linewidth}
\includegraphics[width=\linewidth]{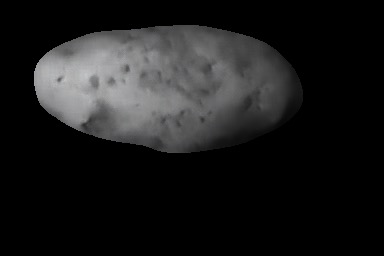}
\includegraphics[width=\linewidth]{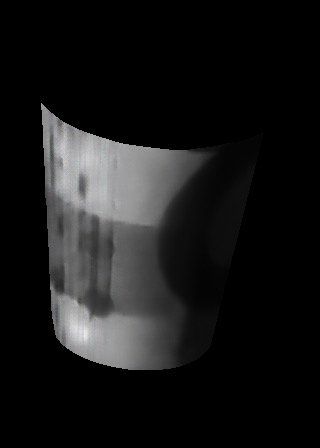}
\includegraphics[width=\linewidth]{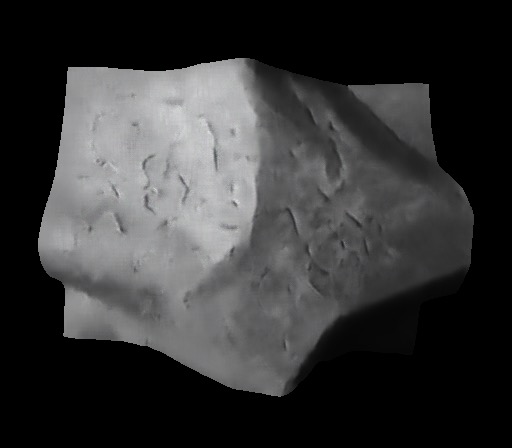}
\includegraphics[width=\linewidth]{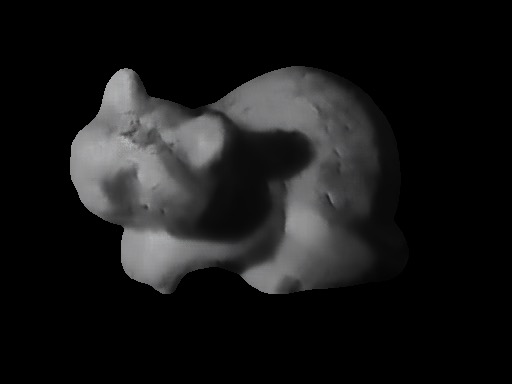}
\includegraphics[width=\linewidth]{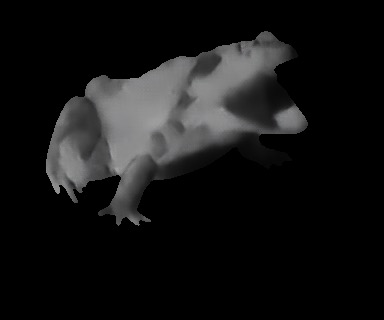}
\includegraphics[width=\linewidth]{Figures/AMSImages/test/shading_5_99_0}
\includegraphics[width=\linewidth]{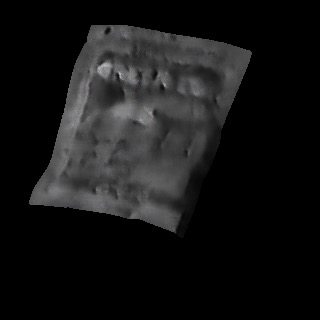}
\includegraphics[width=\linewidth]{Figures/AMSImages/test/shading_7_99_0}
\includegraphics[width=\linewidth]{Figures/AMSImages/test/shading_8_99_0}
\includegraphics[width=\linewidth]{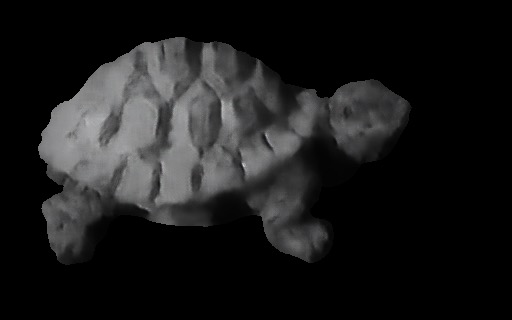}
\caption{Shading}
\end{subfigure}
\begin{subfigure}[b]{.12\linewidth}
\includegraphics[width=\linewidth]{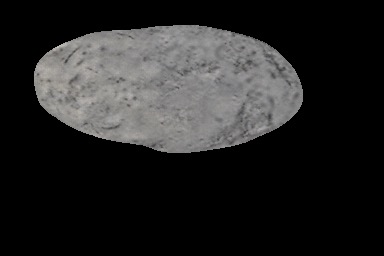}
\includegraphics[width=\linewidth]{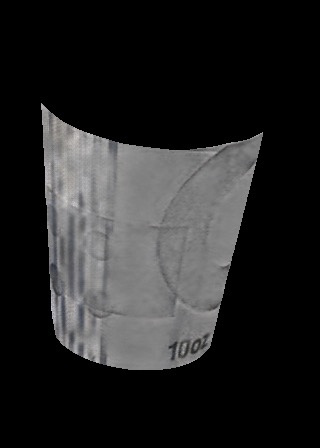}
\includegraphics[width=\linewidth]{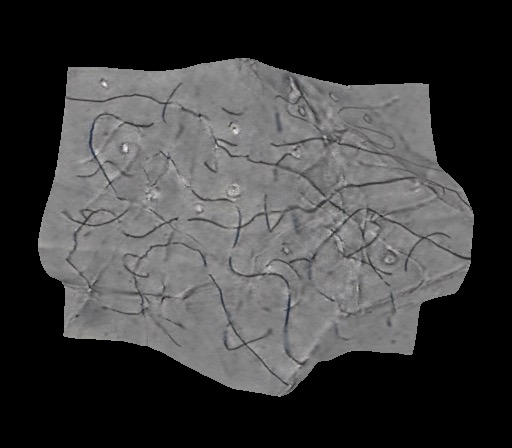}
\includegraphics[width=\linewidth]{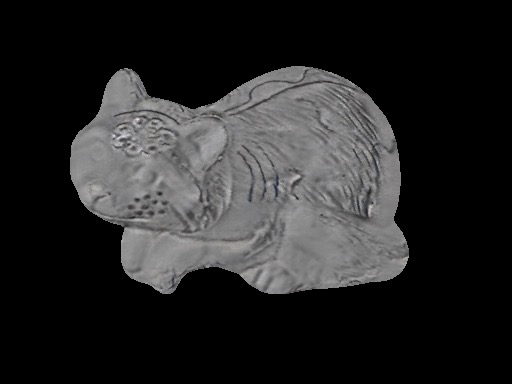}
\includegraphics[width=\linewidth]{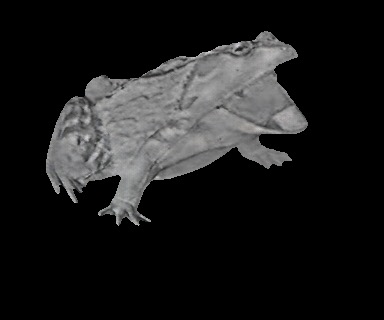}
\includegraphics[width=\linewidth]{Figures/AMSImages/test/material_5_99_0}
\includegraphics[width=\linewidth]{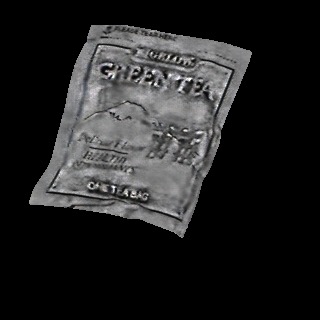}
\includegraphics[width=\linewidth]{Figures/AMSImages/test/material_7_99_0}
\includegraphics[width=\linewidth]{Figures/AMSImages/test/material_8_99_0}
\includegraphics[width=\linewidth]{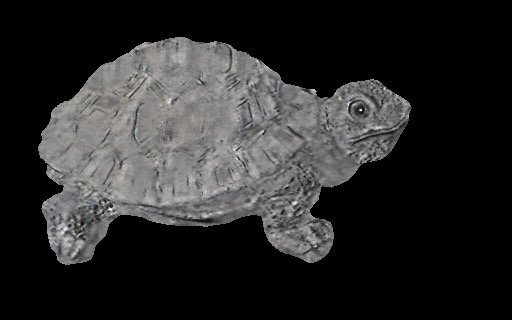}
\caption{Material}
\end{subfigure}
\begin{subfigure}[b]{.12\linewidth}
\includegraphics[width=\linewidth]{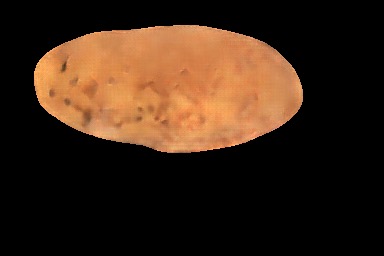}
\includegraphics[width=\linewidth]{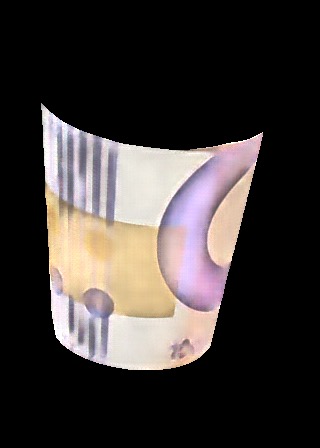}
\includegraphics[width=\linewidth]{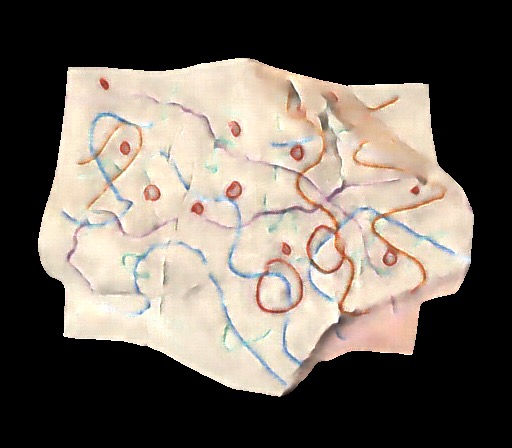}
\includegraphics[width=\linewidth]{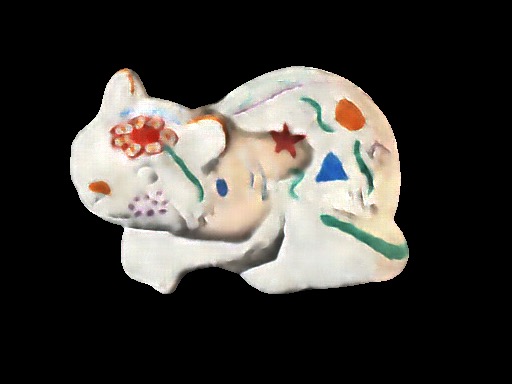}
\includegraphics[width=\linewidth]{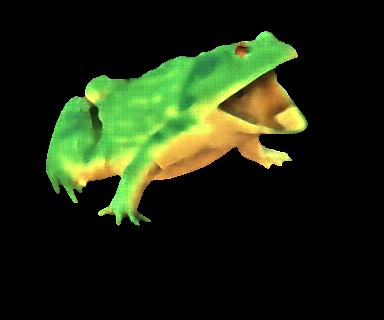}
\includegraphics[width=\linewidth]{Figures/AMSImages/test/albedo_5_99_0}
\includegraphics[width=\linewidth]{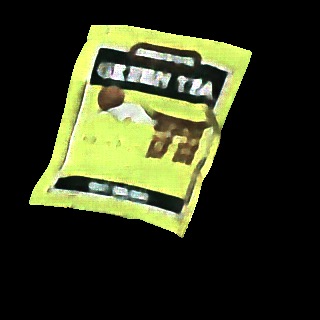}
\includegraphics[width=\linewidth]{Figures/AMSImages/test/albedo_7_99_0}
\includegraphics[width=\linewidth]{Figures/AMSImages/test/albedo_8_99_0}
\includegraphics[width=\linewidth]{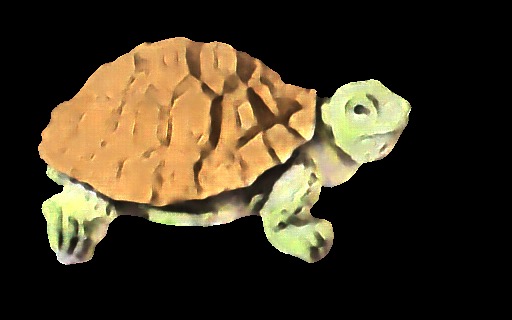}
\caption{Albedo}
\end{subfigure}
\caption{Albedo, Shading, and Shading Detail decompositions on all images from MIT Test for our method.}
\label{fig:mit_AMS_test}
\end{figure*}

All decompositions for MIT test are shown in figure~\ref{fig:mit_AMS_test}.

\section{Shading Shading Detail on MIT}
\begin{figure*}
\centering
\begin{subfigure}[b]{.13\linewidth}
\includegraphics[width=\linewidth]{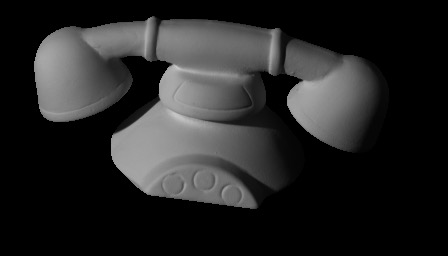}
\includegraphics[width=\linewidth]{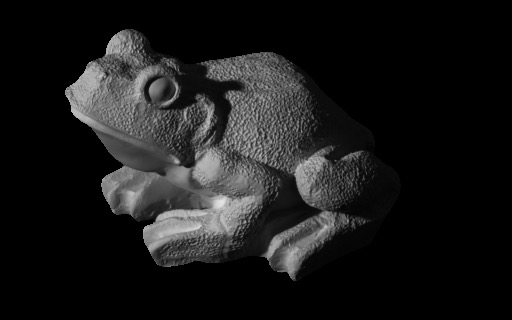}
\includegraphics[width=\linewidth]{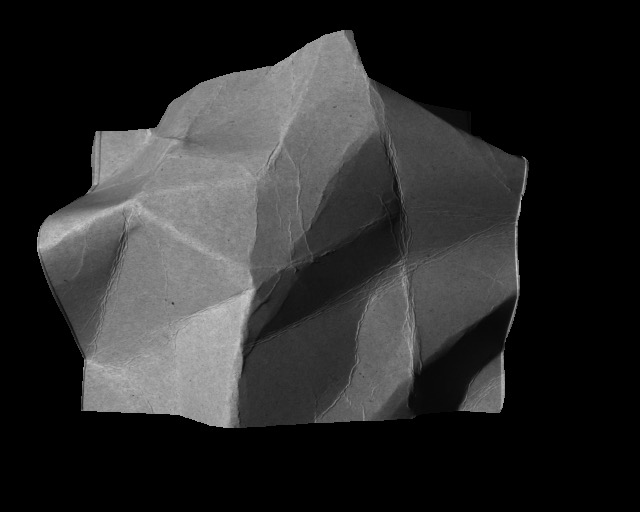}
\includegraphics[width=\linewidth]{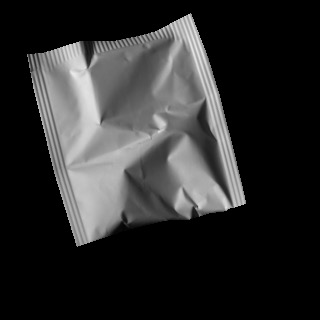}
\includegraphics[width=\linewidth]{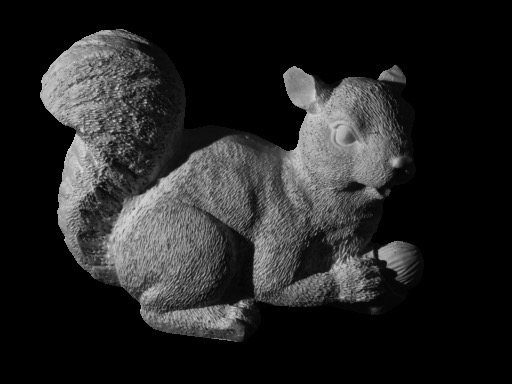}
\includegraphics[width=\linewidth]{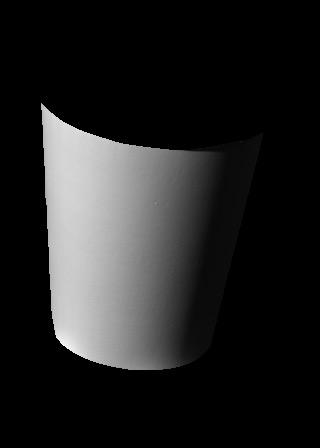}
\includegraphics[width=\linewidth]{Figures/MaterialShading/test/shading_gt_2_99_0}
\includegraphics[width=\linewidth]{Figures/MaterialShading/test/shading_gt_3_99_0}
\includegraphics[width=\linewidth]{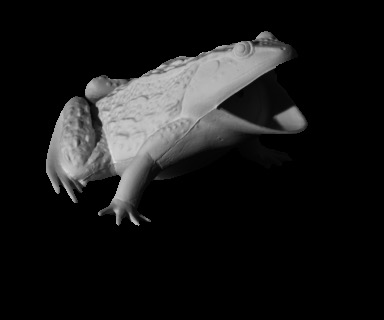}
\includegraphics[width=\linewidth]{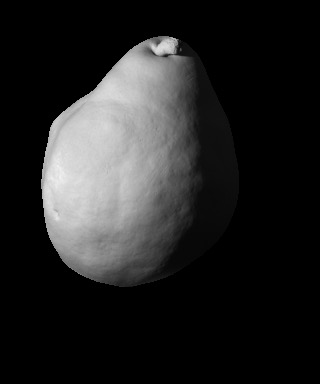}
\caption{Input Image}
\end{subfigure}
\begin{subfigure}[b]{.13\linewidth}
\includegraphics[width=\linewidth]{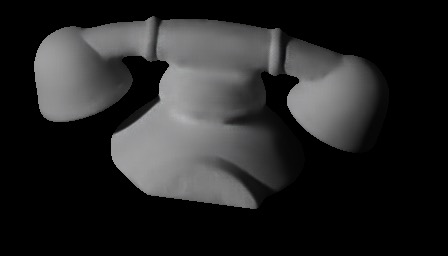}
\includegraphics[width=\linewidth]{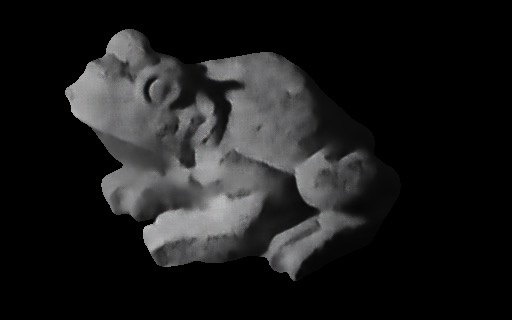}
\includegraphics[width=\linewidth]{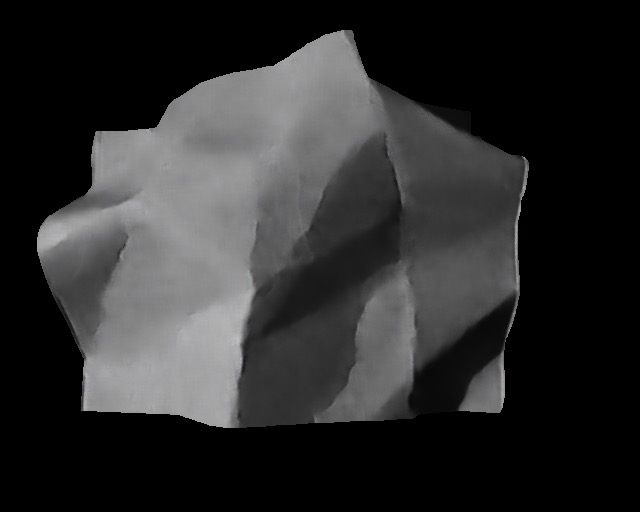}
\includegraphics[width=\linewidth]{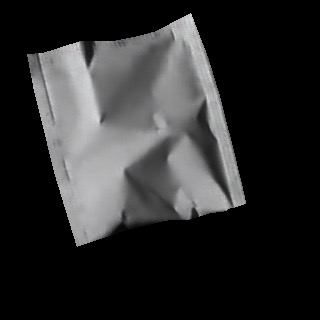}
\includegraphics[width=\linewidth]{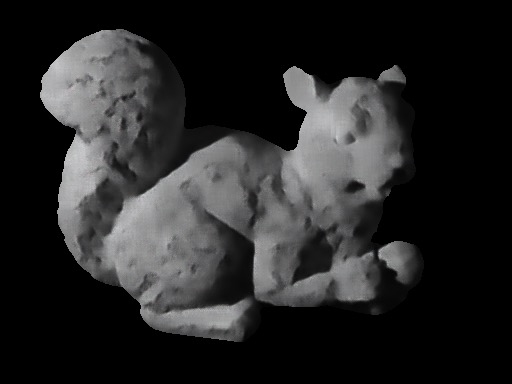}
\includegraphics[width=\linewidth]{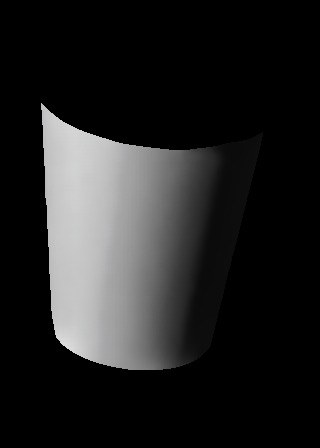}
\includegraphics[width=\linewidth]{Figures/MaterialShading/test/shading_2_99_0}
\includegraphics[width=\linewidth]{Figures/MaterialShading/test/shading_3_99_0}
\includegraphics[width=\linewidth]{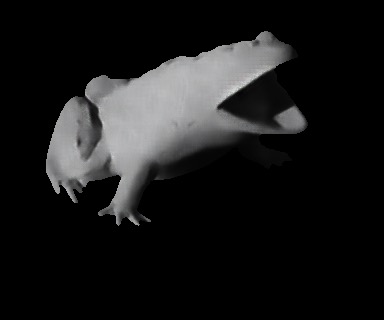}
\includegraphics[width=\linewidth]{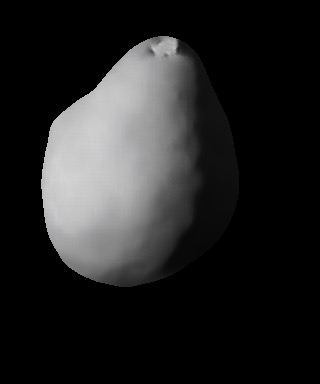}
\caption{Shading}
\end{subfigure}
\begin{subfigure}[b]{.13\linewidth}
\includegraphics[width=\linewidth]{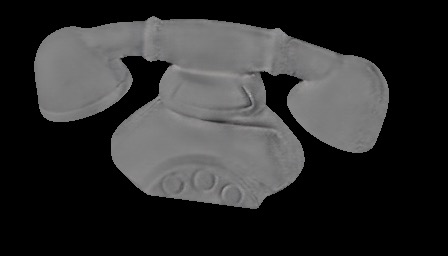}
\includegraphics[width=\linewidth]{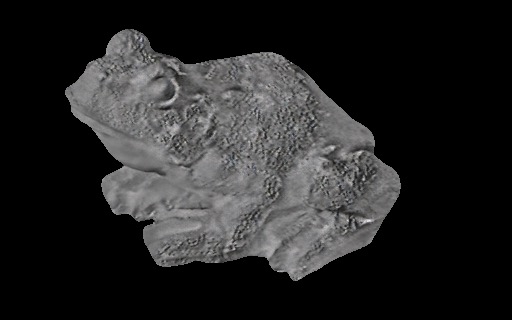}
\includegraphics[width=\linewidth]{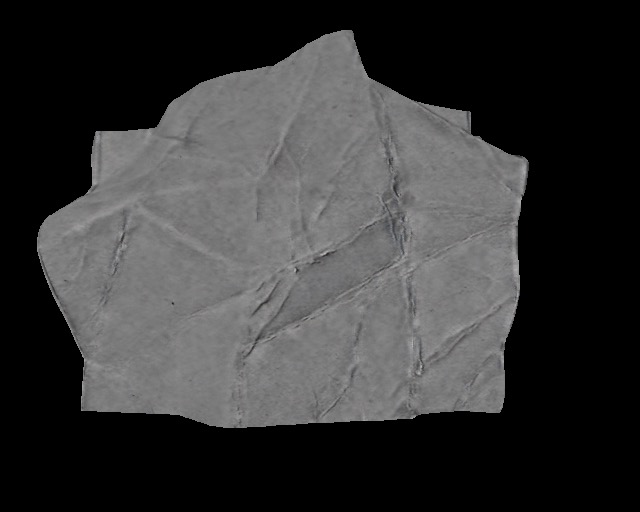}
\includegraphics[width=\linewidth]{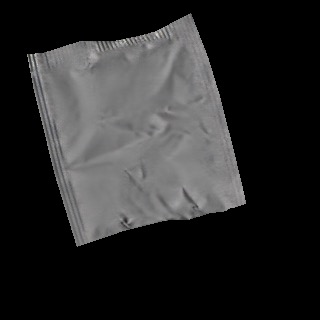}
\includegraphics[width=\linewidth]{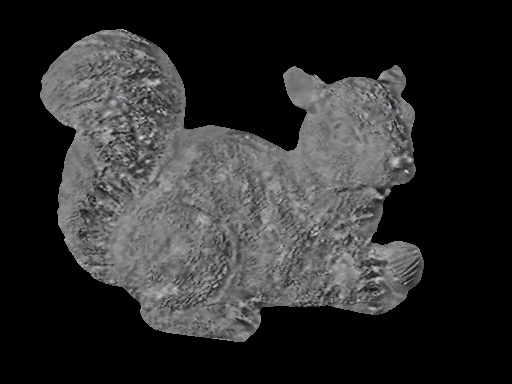}
\includegraphics[width=\linewidth]{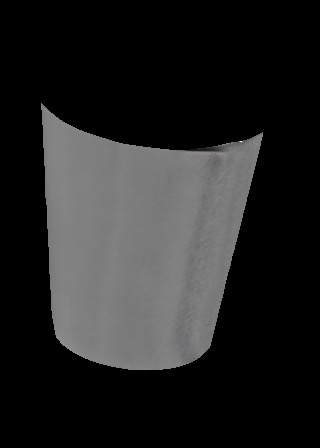}
\includegraphics[width=\linewidth]{Figures/MaterialShading/test/material_2_99_0}
\includegraphics[width=\linewidth]{Figures/MaterialShading/test/material_3_99_0}
\includegraphics[width=\linewidth]{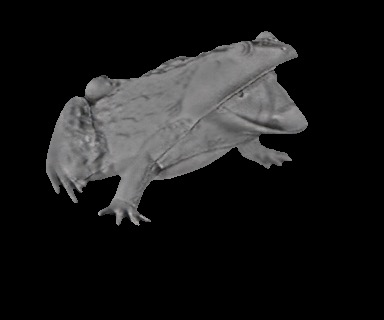}
\includegraphics[width=\linewidth]{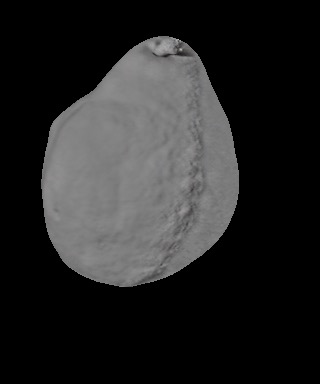}
\caption{Detail}
\end{subfigure}
\begin{subfigure}[b]{.13\linewidth}
\includegraphics[width=\linewidth]{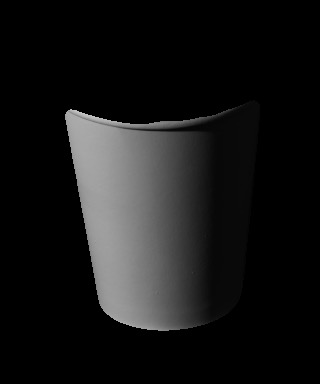}
\includegraphics[width=\linewidth]{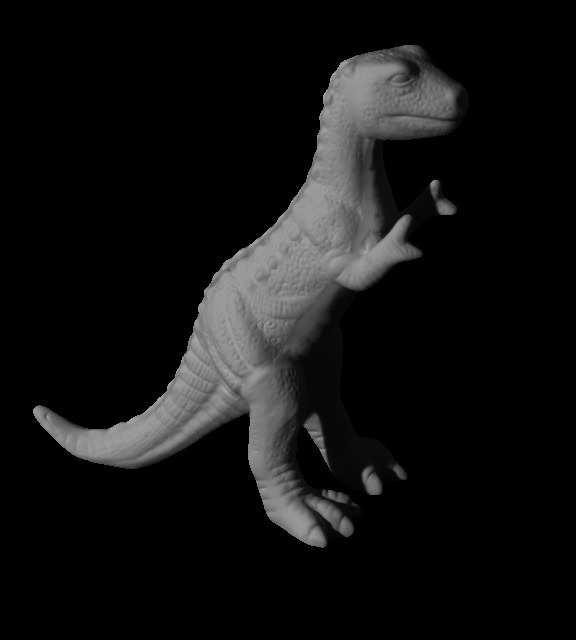}
\includegraphics[width=\linewidth]{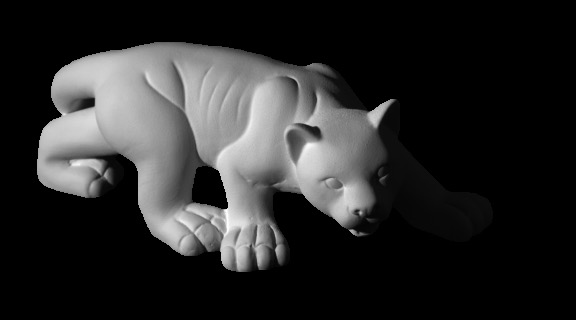}
\includegraphics[width=\linewidth]{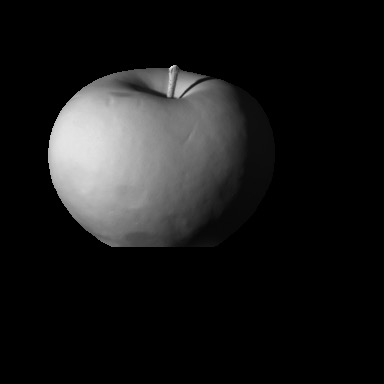}
\includegraphics[width=\linewidth]{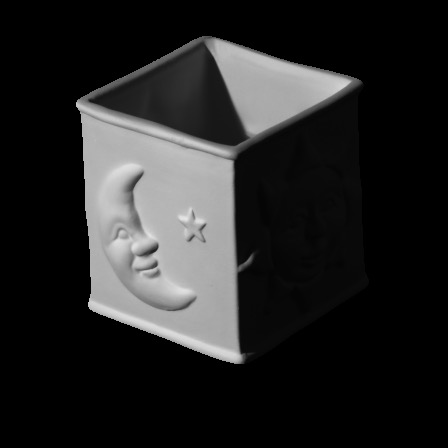}
\includegraphics[width=\linewidth]{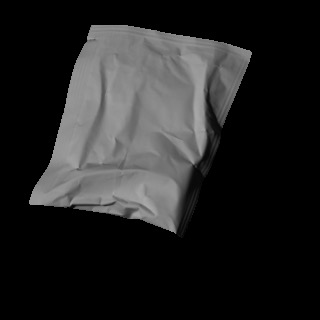}
\includegraphics[width=\linewidth]{Figures/MaterialShading/test/shading_gt_7_99_0}
\includegraphics[width=\linewidth]{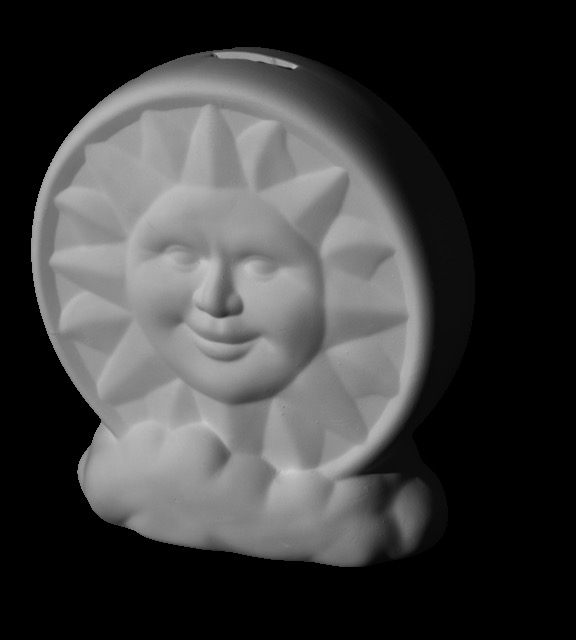}
\includegraphics[width=\linewidth]{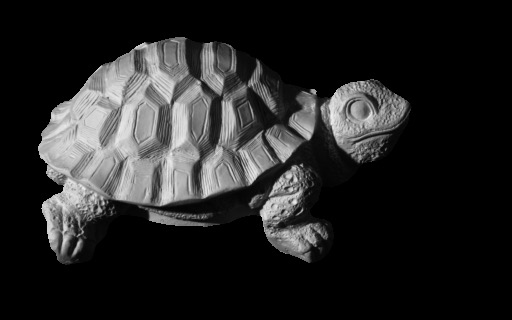}
\includegraphics[width=\linewidth]{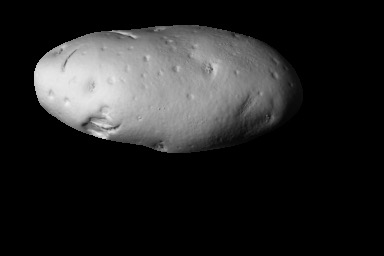}
\caption{Input Image}
\end{subfigure}
\begin{subfigure}[b]{.13\linewidth}
\includegraphics[width=\linewidth]{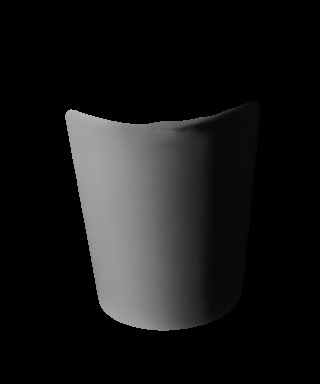}
\includegraphics[width=\linewidth]{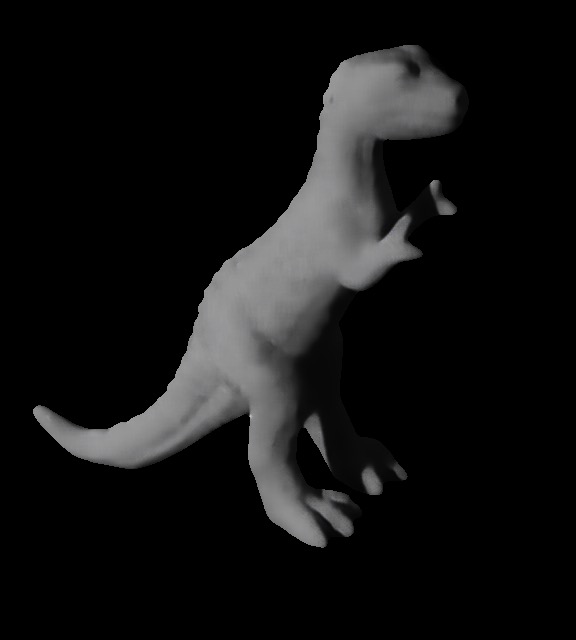}
\includegraphics[width=\linewidth]{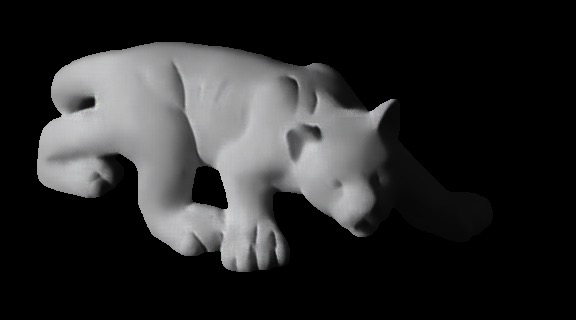}
\includegraphics[width=\linewidth]{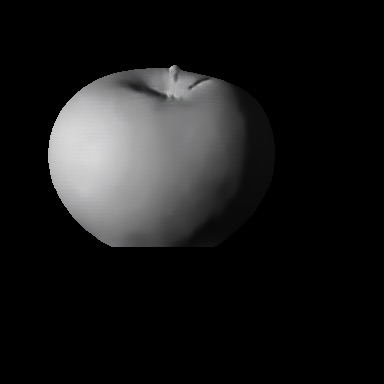}
\includegraphics[width=\linewidth]{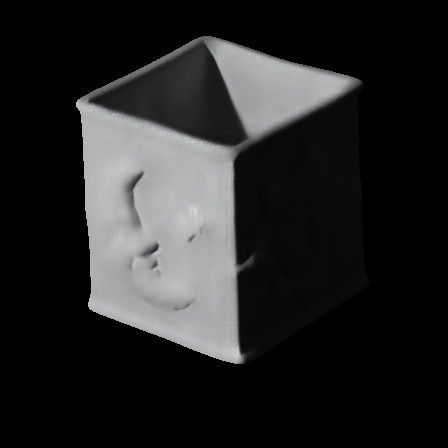}
\includegraphics[width=\linewidth]{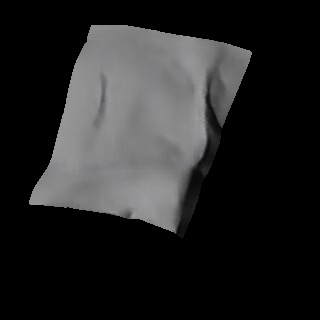}
\includegraphics[width=\linewidth]{Figures/MaterialShading/test/shading_7_99_0}
\includegraphics[width=\linewidth]{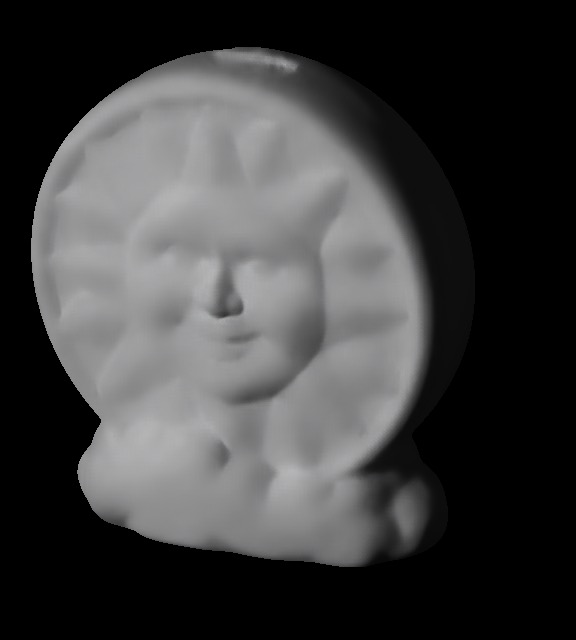}
\includegraphics[width=\linewidth]{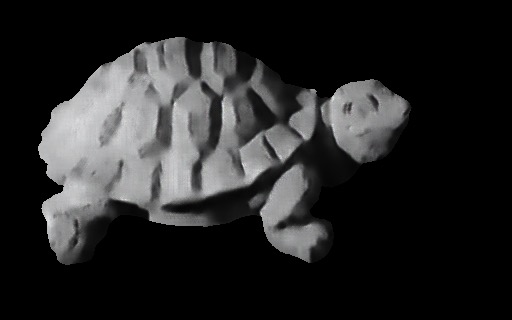}
\includegraphics[width=\linewidth]{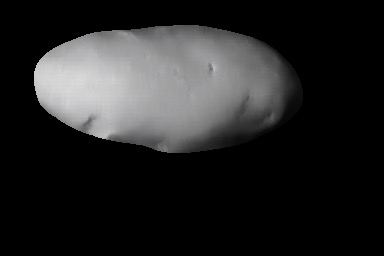}
\caption{Shading}
\end{subfigure}
\begin{subfigure}[b]{.13\linewidth}
\includegraphics[width=\linewidth]{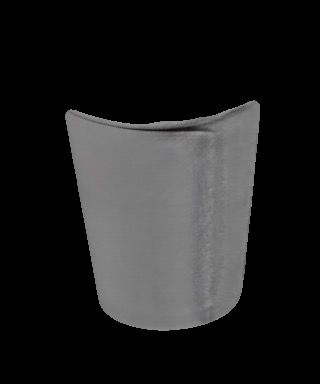}
\includegraphics[width=\linewidth]{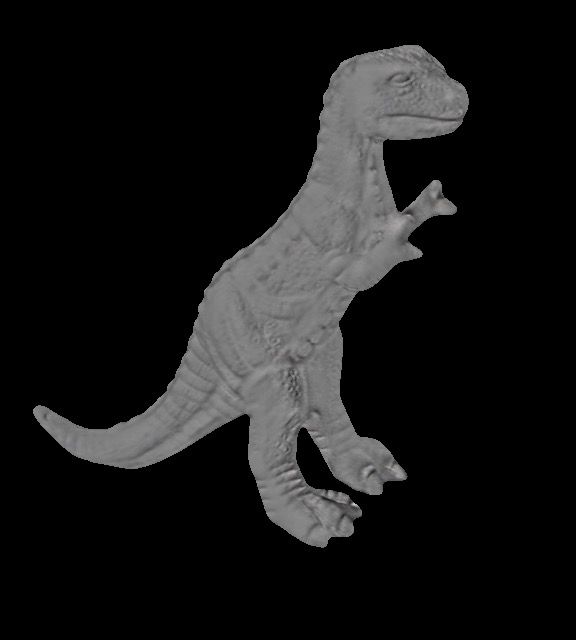}
\includegraphics[width=\linewidth]{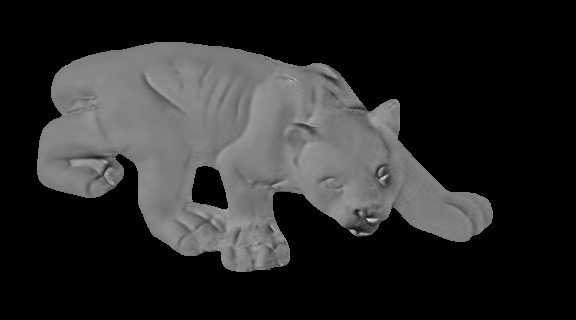}
\includegraphics[width=\linewidth]{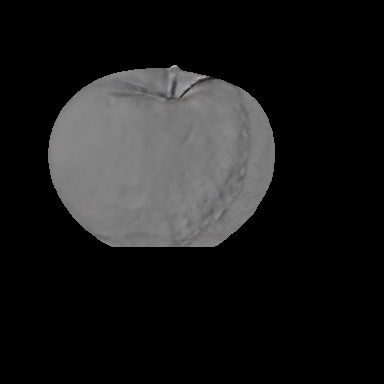}
\includegraphics[width=\linewidth]{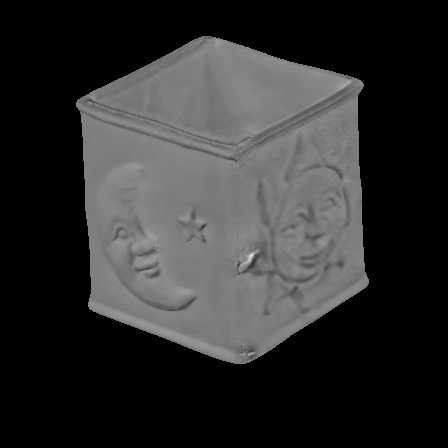}
\includegraphics[width=\linewidth]{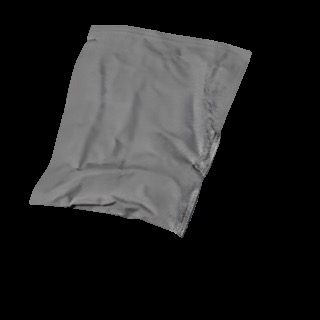}
\includegraphics[width=\linewidth]{Figures/MaterialShading/test/material_7_99_0}
\includegraphics[width=\linewidth]{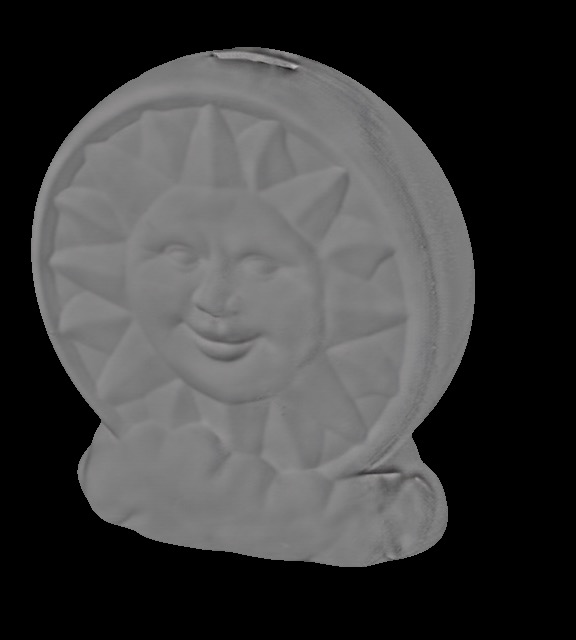}
\includegraphics[width=\linewidth]{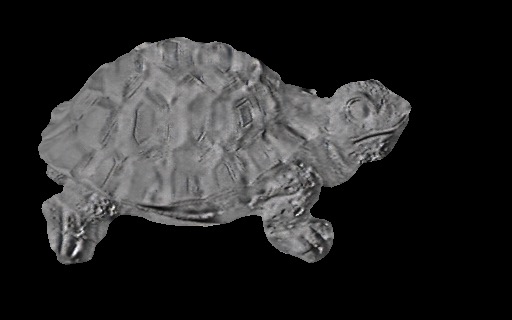}
\includegraphics[width=\linewidth]{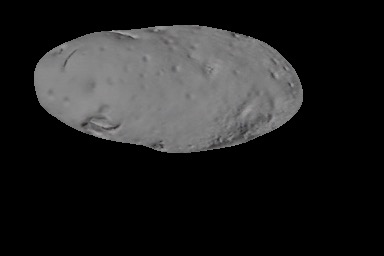}
\caption{Detail}
\end{subfigure}
\caption{Shading and Shading Detail decompositions on MIT from our method.}
\label{fig:app_MIT_shading_detail_train}
\end{figure*}

All decompositions for MIT, test and train, are shown in figure~\ref{fig:app_MIT_shading_detail_train}.

\section{Shading Shading Detail on Vases}
\begin{figure*}
\centering
\begin{subfigure}[b]{.14\linewidth}
\includegraphics[width=\linewidth]{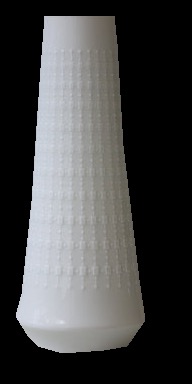}
\includegraphics[width=\linewidth]{Figures/Vase/im_2_99_0}
\includegraphics[width=\linewidth]{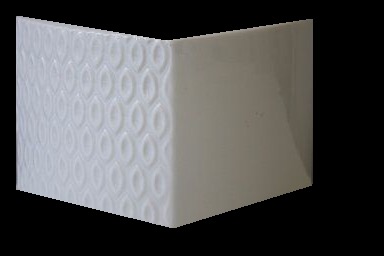}
\caption{Input Image}
\end{subfigure}
\begin{subfigure}[b]{.14\linewidth}
\includegraphics[width=\linewidth]{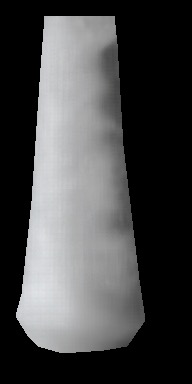}
\includegraphics[width=\linewidth]{Figures/Vase/shading_2_99_0}
\includegraphics[width=\linewidth]{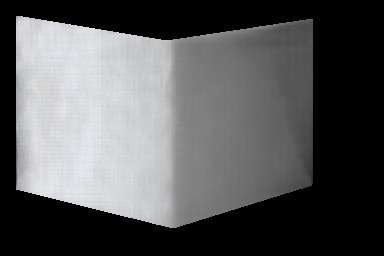}
\caption{Shading}
\end{subfigure}
\begin{subfigure}[b]{.14\linewidth}
\includegraphics[width=\linewidth]{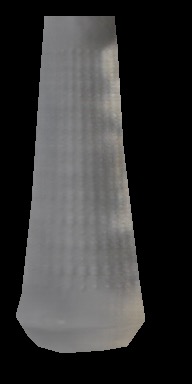}
\includegraphics[width=\linewidth]{Figures/Vase/material_2_99_0}
\includegraphics[width=\linewidth]{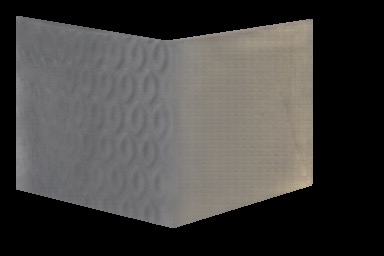}
\caption{Shading Detail}
\end{subfigure}
\begin{subfigure}[b]{.14\linewidth}
\includegraphics[width=\linewidth]{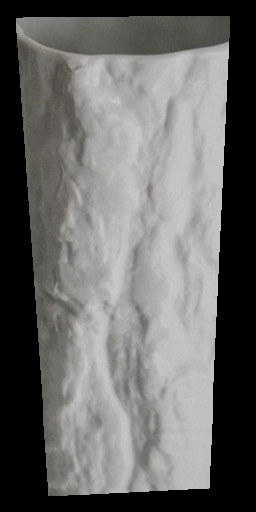}
\includegraphics[width=\linewidth]{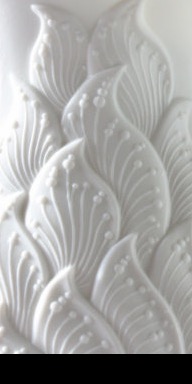}
\includegraphics[width=\linewidth]{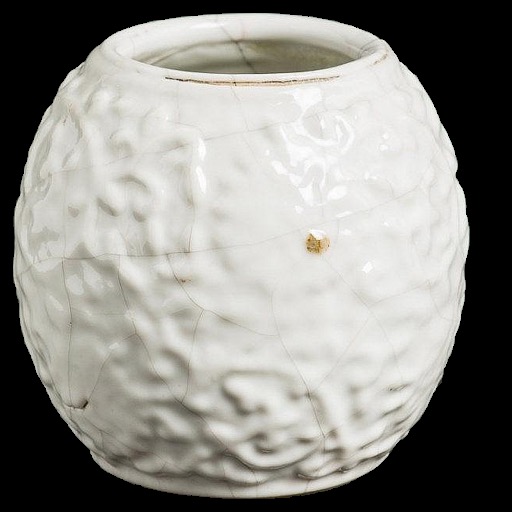}
\caption{Input Image}
\end{subfigure}
\begin{subfigure}[b]{.14\linewidth}
\includegraphics[width=\linewidth]{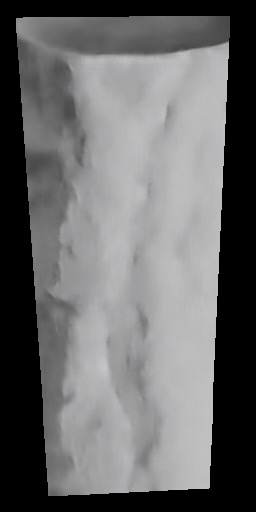}
\includegraphics[width=\linewidth]{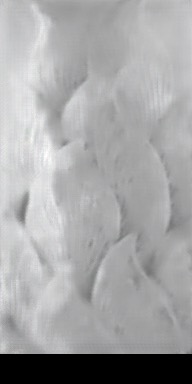}
\includegraphics[width=\linewidth]{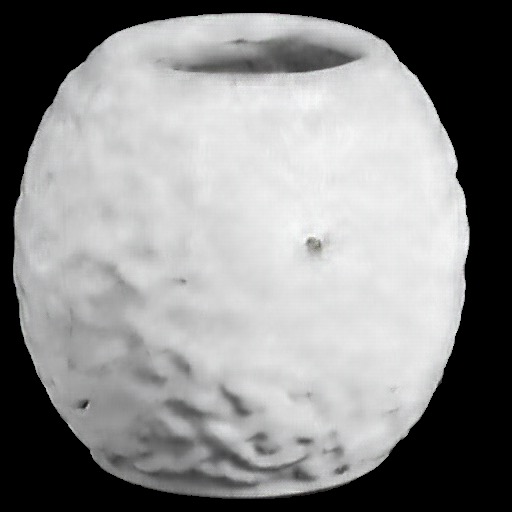}
\caption{Shading}
\end{subfigure}
\begin{subfigure}[b]{.14\linewidth}
\includegraphics[width=\linewidth]{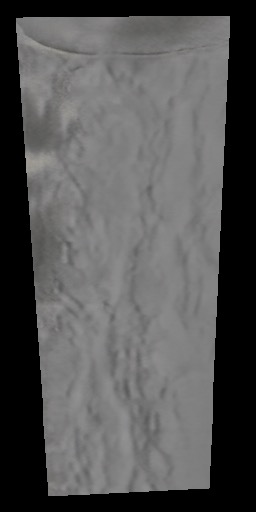}
\includegraphics[width=\linewidth]{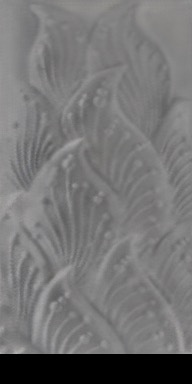}
\includegraphics[width=\linewidth]{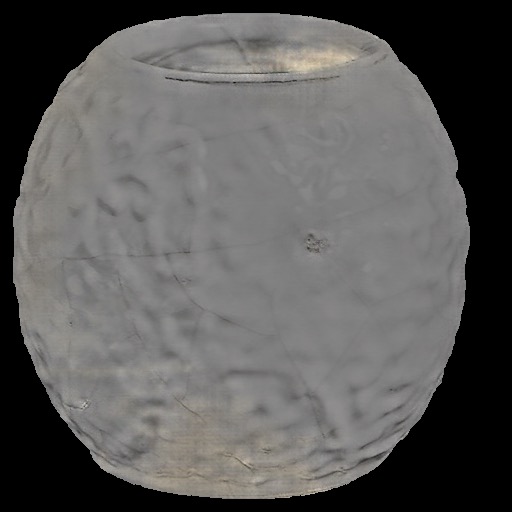}
\caption{Shading Detail}
\end{subfigure}
\caption{Shading and Shading Detail decompositions on vases for our method.}
\label{fig:app_vase}
\end{figure*}

Decompositions of vases into shading and shading detail are shown in
figure~\ref{fig:app_vase}

\end{appendices}
\end{document}